\documentclass[letterpaper, 10 pt, conference]{ieeeconf}

\IEEEoverridecommandlockouts                              % This command is only needed if
\usepackage{mathptmx} % assumes new font selection scheme installed
\usepackage{times} % assumes new font selection scheme installed
\usepackage{amsmath} % assumes amsmath package installed
\usepackage{amssymb}  % assumes amsmath package installed
\usepackage{tabularx}
\usepackage{url}
\usepackage[figurename=Fig.\ ]{caption}
\usepackage{epsfig} % for postscript graphics files
\usepackage{cite}
\usepackage{color}
\usepackage{booktabs}  %
\usepackage{amsmath}
\usepackage{amssymb}
\usepackage{indentfirst}
\usepackage{algorithm}
\usepackage{algorithmic}
\usepackage{caption}
\usepackage{comment}
\usepackage{color,soul}
\usepackage{ifthen}
\newcommand{\configuration}{\bm{q}}
\newcommand{\taskfunction}{\bm{e}} %% e

\newcommand{\pcont}{\bm{p}_{\mathrm{control}}}
\newcommand{\ptgt}{\bm{p}_{\mathrm{tgt}}}

\newcommand{\opg}{OPG}
\newcommand{\hfpg}{HFPG}
\newcommand{\ti}{Tablis Interface}
\newcommand{\tpg}{Target Pose Generator}
\newcommand{\wbc}{WBC}
\newcommand{\fsc}{Foot Step Commander}
\newcommand{\bms}{\bm{s}}
\newcommand{\bmu}{\bm{u}}
\newcommand{\bmp}{\bm{p}}
\newcommand{\bmh}{\bm{h}}
\usepackage{balance}
\usepackage{mathtools}%
\usepackage{bmpsize}
\usepackage{bm}
\newcommand{\secref}[1]{Sec. \ref{#1}}
\newcommand{\figref}[1]{{Fig.\ref{#1}}}
\newcommand{\tabref}[1]{{Table.\ref{#1}}}
\newcommand{\equref}[1]{{Eq.\ref{#1}}}
\newcommand{\subsecref}[1]{SubSec. \ref{#1}}

\newif\ifdraft
\drafttrue % turn on highlights and comments
\ifdraft

\else

\fi

\newcommand{\w}{\bm{w}}

\newcommand{\f}{\bm{f}}
\newcommand{\n}{\bm{n}}
\newcommand{\g}{\bm{g}}
\newcommand{\p}{\bm{p}}
\newcommand{\R}{\bm{R}}

\newcommand{\btau}{\bm{\tau}}
\newcommand{\jangle}{\bm{\theta}}

\newcommand{\bthetahfpg}{\bm{\theta}_\mathrm{hfpg}}

\newcommand{\q}{\bm{q}}

\newcommand{\A}{\bm{A}}

\newcommand{\function}{F}
\newcommand{\kin}{Kin}
\newcommand{\eom}{Eom}
\newcommand{\trq}{Trq}
\newcommand{\com}{Com}

\usepackage{subfigure}

\newcommand{\chapref}[1]{Sec.~\ref{#1}}

\newboolean{Draft}
\setboolean{Draft}{false}
\newboolean{with-hand}
\setboolean{with-hand}{false}
\ifthenelse{\boolean{Draft}}{ %%%%%%% Japanese start %%%%%%%
  \title{\LARGE \bf
    高耐久性ヒューマノイドにおける全身バイラテラル操縦デバイスと\\オンライン全身姿勢最適化を用いた全身作業模倣学習システム
  }
}{
  \title{\LARGE \bf
    Development of a Whole-body Work Imitation Learning System \\by a Biped and Bi-armed Humanoid
  }
} %%%%%%% English end %%%%%%%

\author{
Yutaro Matsuura$^{1}$,
Kento Kawaharazuka$^{1}$,
Naoki Hiraoka$^{1}$,
Kunio Kojima$^{1}$,
Kei Okada$^{1}$,
Masayuki Inaba$^{1}$   % <-this % stops a space
\thanks{$^{1}$Y.Matsuura, K.Kawaharazuka, N.HiraokaK. K.Kojima, K.Okada, M.Inaba, are with Department of Mechano-Infomatics, The University of Tokyo, 7-3-1 Hongo, Bunkyo-ku, Tokyo 113-8656, Japan
  {\tt\small matsuura@jsk.imi.i.u-tokyo.ac.jp}}%
  }

\begin{document}
\maketitle
\thispagestyle{empty}
\pagestyle{empty}
%%%%%%%%%%%%%%%%%%%%%%%%%%%%%%%%%%%%%%%%%%%%%%%%%%%%%%%%%%%%%%%%%%%%%%%%%%%%%%%%
\begin{abstract}
  \ifthenelse{\boolean{with-hand}}{
  \ifthenelse{\boolean{Draft}}{ %%%%%%% Japanese start %%%%%%%
  近年模倣学習の研究が盛んに行われており、多様な作業模倣が可能になりつつある。
  特に身体が固定されており、ロボットのルートリンクの位置姿勢やカメラ画角が変化しないロボットによるスキル獲得が多く実現されている。
  一方でヒューマノイドロボットのような浮遊リンク系のロボットの行動模倣は未だに難しい課題である。
  そこで本研究では, 浮遊リンク系の二足歩行ロボットによる模倣学習システムを開発する.
  二足歩行型のヒューマノイドにおいて模倣学習が可能なシステムを組む際の問題点は主に3つである.
  1つ目はヒューマノイドの操縦が可能な操縦デバイス, 2つ目は高負荷作業や長期的な作業継続のためのハードウェア、3つ目は浮遊リンク系で長期的なデータ取得に耐え得る制御である.
  1点目については、本研究ではTABLIS\cite{tablis}を用いる.
  腕だけでなく脚も制御可能であり, ロボットとの間でバイラテラル制御が可能である。このTABLISとヒューマノイドロボットJAXON\cite{JAXON}を繋ぎ, 模倣学習のための操縦システムを構築する。
  2点目については、環境との接点に高耐久の多指ハンド%% ヒューマノイドが全身の出力を活かした作業を行い、長時間に渡りデータを収集するためには、ハンドの耐久性能が重要となる。
  であるMSL HAND\cite{mslhand}を利用することで解決する。
  3点目については, 最適化に基づく姿勢制御を行うことで長期的なデータ収集が可能で、なおかつ、ロボットの手、頭や腰や足も同時に動かせるシステムを構築する.
  全身関節トルク最小化・接触力最適化を含む姿勢生成法と、追従性の求められるタスクのための高周期の姿勢生成を組み合わせる。
  上記の3つの特徴をもつヒューマノイドロボットに向けたシステムを用いて模倣学習を行い多様な動作を実現する.
  最後に、手だけでなく頭や腰も同時に動くような柔軟布の操作、ヒューマノイドに特徴的な脚を使った物体操作、そして手や腰,膝を曲げると同時に大きな力を必要とする重量物体の操作を行いこのシステムの有効性を示す。
}{
  } %%%%%%% English end %%%%%%%
  }{
    \ifthenelse{\boolean{Draft}}{ %%%%%%% Japanese start %%%%%%%
      近年模倣学習の研究が盛んに行われており、多様な作業模倣が可能になりつつある。
      特に身体が固定されており、ロボットのルートリンクの位置姿勢やカメラ画角が変化しないロボットによるスキル獲得が多く実現されている。
      一方でヒューマノイドロボットのような浮遊リンク系のロボットの行動模倣は未だに難しい課題である。
      そこで本研究では, 浮遊リンク系の二足歩行ロボットによる模倣学習システムを開発する.
      二足歩行型のヒューマノイドにおいて模倣学習が可能なシステムを組む際の問題点は主に2つである.
      1つ目はヒューマノイドの操縦が可能な操縦デバイス, 2つ目は浮遊リンク系で高負荷作業や長期的なデータ取得に耐え得る制御である.
      1点目については、本研究ではTABLIS\cite{tablis}を用いる.
      腕だけでなく脚も制御可能であり, ロボットとの間でバイラテラル制御が可能である。このTABLISとヒューマノイドロボットJAXON\cite{JAXON}を繋ぎ, 模倣学習のための操縦システムを構築する。
      2点目については, 最適化に基づく姿勢制御を行うことで長期的なデータ収集が可能で、なおかつ、ロボットの手、頭や腰や足も同時に動かせるシステムを構築する.
      全身関節トルク最小化・接触力最適化を含む姿勢生成法と、追従性の求められるタスクのための高周期の姿勢生成を組み合わせる。
      上記の2つの特徴をもつヒューマノイドロボットに向けたシステムを用いて模倣学習を行い多様な動作を実現する.
      最後に、手だけでなく頭や腰も同時に動くような柔軟布の操作、ヒューマノイドに特徴的な脚を使った物体操作、そして手や腰,膝を曲げると同時に大きな力を必要とする重量物体の操作を行いこのシステムの有効性を示す。
    }{
      Imitation learning has been actively studied in recent years. In particular, skill acquisition by a robot with a fixed body, whose root link position and posture and camera angle of view do not change, has been realized in many cases. On the other hand, imitation of the behavior of robots with floating links, such as humanoid robots, is still a difficult task. In this study, we develop an imitation learning system using a biped robot with a floating link. There are two main problems in developing such a system. The first is a teleoperation device for humanoids, and the second is a control system that can withstand heavy workloads and long-term data collection. For the first point, we use the whole body control device TABLIS. It can control not only the arms but also the legs and can perform bilateral control with the robot. By connecting this TABLIS with the high-power humanoid robot JAXON, we construct a control system for imitation learning. For the second point, we will build a system that can collect long-term data based on posture optimization, and can simultaneously move the robot's limbs. We combine high-cycle posture generation with posture optimization methods, including whole-body joint torque minimization and contact force optimization. We designed an integrated system with the above two features to achieve various tasks through imitation learning. Finally, we demonstrate the effectiveness of this system by experiments of manipulating flexible fabrics such that not only the hands but also the head and waist move simultaneously, manipulating objects using legs characteristic of humanoids, and lifting heavy objects that require large forces.
    }
    }
\end{abstract}
%% これまで模倣学習が発展してきて, 多様な動作が可能になってきた.
%% それらは主に, HOGE, FUGA, BARのような形である.
%% 一方で, これらは基本的に身体が固定されたロボットであり, 浮遊リンク系を扱っていない.
%% FOOのような研究もあるが, これこれこういう理由で微妙である.
%% そこで本研究では, 浮遊リンク系の二足歩行ロボットによる模倣学習システムを開発する.
%% 二足歩行型のヒューマノイドにおいて模倣学習可能なシステムを組む際の主な問題点は2つある.
%% それはヒューマノイド型を扱うことが可能な操縦デバイスと, 浮遊リンク系において長期的なデータ取得が可能な制御である.
%% 前者について, 通常はVRデバイスや3Dマウス, GUI等によりロボットを制御するが, 本研究では石黒らの開発したtablisを用いる.
%% これは, 腕だけでなく脚を制御可能なデバイスであり, タスクを行うロボットとの間でバイラテラル制御が可能である.
%% このtablisとヒューマノイドjaxonを繋ぎ, データ取得可能なシステムを構築する(ここでのニュアンスとして, tablisをただ用いたという感じより, tablisとjaxon, そしてデータ取得するためのボタンやデータの流れ等のシステムを詳細に述べて, 模倣学習のためのシステム感をしっかり出すことが大切)
%% 後者について, トルク最適化に基づく姿勢制御を行い, 長期的な動作が可能なシステムを構築する.
%% これは HOGE-であり, FUGAである.
%% 最後に, この2つの特徴をもつヒ-ューマノイドロボットに向けたシステムを用いて, データを収集, 模倣学習により多様な動作を行う.
%% これを, 手だけでなく頭や腰も同時に動くようなヒューマノイドにおける風呂敷のマニピュレーション実験, 手や腰, 膝を曲げると同時に, 大きな力を必要とする重量物体操作実験, ヒューマノイドに特徴的な脚を使い物体を操作するゴミ箱操作実験により有効性を示す.

%% あとはこの流れに沿って,
%% 手法: 全体システム概要, tablis-jaxonシステムとデータの流れやボタンの利用, トルク最適化姿勢制御, 模倣学習
%% 実験: 前に述べた3つ
%% 考察: 問題だった点
%%%%%%%%%%%%%%%%%%%%%%%%%%%%%%%%%%%%%%%%%%%%%%%%%%%%%%%%%%%%%%%%%%%%%%%%%%%%%%%%
\section{INTRODUCTION} \label{sec:intro}
%%\cite{Kanoulas2018CenterofMassBasedGP, RSJ-nozawa, murooka:groping, yamazaki:hierarchical-relationship}
\ifthenelse{\boolean{with-hand}}{
\ifthenelse{\boolean{Draft}}{ %%%%%%% Japanese start %%%%%%%
  近年模倣学習の研究が盛んに行われており、多様な動作の獲得が可能になりつつある。
  %% 双腕台車ロボットによるテーブル上での物体の移動\cite{vr-imitation}や,紐結びタスク\cite{himo}、布をたたむタスク\cite{humanoid-folding}などが実現されている。
  特に双腕ロボットによる様々なスキル獲得が実現されている\cite{vr-imitation, himo}。
  これらの研究では基本的に身体が固定されており、作業中にロボットのルートリンクの位置姿勢やカメラ画角が変化することは無い。
  %% これはカメラ画角が一定である点で有利である。
  一方でヒューマノイドロボットのような浮遊リンク系のロボットに関する研究は少なく未だに難しい課題である。
  そこで本研究では, 浮遊リンク系の二足歩行ロボットによる模倣学習システムを開発する.
  二足歩行型のヒューマノイドにおいて模倣学習が可能なシステムを組む際の問題点は主に2つである.
  1つ目はヒューマノイドの操縦が可能な操縦デバイス, 2つ目は浮遊リンク系においても長期的なデータ取得に耐え得る制御である.

  1点目については、通常はVRデバイスや3Dマウス, GUI等によりロボットを制御する場合が多い。
  動的ニューラルネットワークモデルを用いて、小型ヒューマノイドロボットの物体操作行動の学習を行い環境に合わせた動作を再現した研究\cite{small-humanoid-ball-roll-direct-teaching}や
  小型ヒューマノイドで直接教示を行い、複数のモダリティからの情報を自動的に考慮したモデルを作成する研究\cite{nao-multimodal}なども行われている。
  %% これらの教示では、人自身が作業を行うときの直感的な体の動かし方ではなく、ロボットが作業を完遂できる姿勢を外側から考えて作ってやることとなる。この時の人の動きは、自身が同じ作業を行うときの体の動かし方からは乖離しており直感的ではない。
  このような直接動作教示には、非直感的であり、かつバックドライバビリティが高いロボットでのみしか実現できないという難しさが存在する。
  モーションキャプチャを用いた遠隔操縦による簡単な動作の再現\cite{hello-motion}も実現されているが、操縦者への反力提示ができないため、重量物操作のような環境との接触力操作が重要な作業には適応が難しい。
  布の折りたたみタスクを実現するアプローチ\cite{similar}も提案されているが、ヒューマノイド特有の足の利用やカメラ画角の変化は含まれていない。
  そこで本研究では石黒らの開発したTABLIS\cite{tablis}を用いる.
  このデバイスは, 搭乗型の外骨格コックピッド操縦デバイスであり、腕だけでなく脚も制御可能なデバイスであり, タスクを行うロボットとの間でバイラテラル制御が可能である.
  このTABLISとヒューマノイドロボットJAXON\cite{JAXON}を繋ぎ, 模倣学習のためのデータの取得可能な操縦システムを構築する。%% (ここでのニュアンスとして, tablisをただ用いたという感じより, tablisとjaxon, そしてデータ取得するためのボタンやデータの流れ等のシステムを詳細に述べて, 模倣学習のためのシステム感をしっかり出すことが大切)

  2点目については、ヒューマノイドが全身の出力を活かした作業を行い、長時間に渡りデータを収集するためには、マニピュレーションにおいて環境と一番に接触するハンドの耐久性能が重要となる。
  ワイヤ駆動の高出力ハンドを用いてヒューマノイドによる自重保持動作や重量物の保持を実現した研究\cite{makino-hand}のように、作業の種類を増やすためにハンドの耐久性向上が重要となる。
  本研究では、MSL HAND\cite{mslhand}を利用することでこれを解決する。

  3点目については, トルクや接触力の最適化に基づく姿勢制御を行うことで重量物の操作と長期的なデータ収集が可能で、なおかつ、ロボットの手だけでなく頭や腰や足も同時に動かせるシステムを構築する.
  ロボットの前で操縦者が片足立ちなどの動作を行い、バランス制約を含む模倣を行う研究\cite{nao-oneleg}や、人の双腕作業の様子を観察し重要な動きの学習と干渉回避などの動作の最適化を経て動作を模倣する研究\cite{asimo-imitation}などが行われている。
  しかし学習上の課題が存在し、環境に対するマニピュレーションを行っていなかったり、短時間の双腕のみのタスク模倣にとどまっている。
  操縦の分野でも重量物を操作している研究は少なく、例えば石黒らの研究\cite{tablis}においても短時間の操作にとどまっている。重量物操作のための学習を行っている例はより少ない。
  そこで、本研究では高負荷作業や長時間のデータ収集に適した全身関節トルク最小化を含む低周期の姿勢生成法と、マニピュレーションタスクのような追従性の求められるタスクのための制約を絞った高周期の姿勢生成を組み合わせる。
  操作対象物体も含めたキーポーズの列を生成する最適化手法\cite{shigematsu}が存在するが、模倣学習のためのデータ収集に焦点を当て、操縦のために制約を絞った最適化計算を実装する。

  %% 目の前で人が両手を使った作業を行う様子をヒューマノイドロボットが再現する研究\cite{asimo-imitation}やHMDを使った操縦による教示学習の研究\cite{similar}は存在するが、短時間の双腕のみのタスク模倣にとどまっている。

  上記の3つの特徴をもつヒューマノイドロボットに向けたシステムを用いて、データを収集し模倣学習\cite{kawaharazuka-il}により多様な動作を実現する.
  手だけでなく頭や腰も同時に動くようなヒューマノイドによる風呂敷のマニピュレーション実験、ヒューマノイドに特徴的な脚を使い物体を操作するゴミ箱操作実験、そして手や腰,膝を曲げると同時に大きな力を必要とする重量物体の操作実験を行いこのシステムの有効性を示す。
}{ %%%%%% Japanese end: English start %%%%%
} %%%%%%% English end %%%%%%%

}{
  \ifthenelse{\boolean{Draft}}{ %%%%%%% Japanese start %%%%%%%
  近年模倣学習の研究が盛んに行われており、多様な動作の獲得が可能になりつつある。
  %% 双腕台車ロボットによるテーブル上での物体の移動\cite{vr-imitation}や,紐結びタスク\cite{himo}、布をたたむタスク\cite{humanoid-folding}などが実現されている。
  特に双腕ロボットによる様々なスキル獲得が実現されている\cite{vr-imitation, himo}。
  これらの研究では基本的に身体が固定されており、作業中にロボットのルートリンクの位置姿勢やカメラ画角が変化することは無い。
  %% これはカメラ画角が一定である点で有利である。
  一方でヒューマノイドロボットのような浮遊リンク系のロボットに関する研究は少なく未だに難しい課題である。
  そこで本研究では, 浮遊リンク系の二足歩行ロボットによる模倣学習システムを開発する.
  二足歩行型のヒューマノイドにおいて模倣学習が可能なシステムを組む際の問題点は主に2つである.
  1つ目はヒューマノイドの操縦が可能な操縦デバイス, 2つ目は浮遊リンク系においても長期的なデータ取得に耐え得る制御である.

  1点目については、通常はVRデバイスや3Dマウス, GUI等によりロボットを制御する場合が多い。
  動的ニューラルネットワークモデルを用いて、小型ヒューマノイドロボットの物体操作行動の学習を行い環境に合わせた動作を再現した研究\cite{small-humanoid-ball-roll-direct-teaching}や
  小型ヒューマノイドで直接教示を行い、複数のモダリティからの情報を自動的に考慮したモデルを作成する研究\cite{nao-multimodal}なども行われている。
  %% これらの教示では、人自身が作業を行うときの直感的な体の動かし方ではなく、ロボットが作業を完遂できる姿勢を外側から考えて作ってやることとなる。この時の人の動きは、自身が同じ作業を行うときの体の動かし方からは乖離しており直感的ではない。
  このような直接動作教示には、非直感的であり、かつバックドライバビリティが高いロボットでのみしか実現できないという難しさが存在する。
  モーションキャプチャを用いた遠隔操縦による簡単な動作の再現\cite{hello-motion}も実現されているが、操縦者への反力提示ができないため、重量物操作のような環境との接触力操作が重要な作業には適応が難しい。
  布の折りたたみタスクを実現するアプローチ\cite{similar}も提案されているが、ヒューマノイド特有の足の利用やカメラ画角の変化は含まれていない。
  そこで本研究では石黒らの開発したTABLIS\cite{tablis}を用いる.
  このデバイスは, 搭乗型の外骨格コックピッド操縦デバイスであり、腕だけでなく脚も制御可能なデバイスであり, タスクを行うロボットとの間でバイラテラル制御が可能である.
  このTABLISとヒューマノイドロボットJAXON\cite{JAXON}を繋ぎ, 模倣学習のためのデータの取得可能な操縦システムを構築する。%% (ここでのニュアンスとして, tablisをただ用いたという感じより, tablisとjaxon, そしてデータ取得するためのボタンやデータの流れ等のシステムを詳細に述べて, 模倣学習のためのシステム感をしっかり出すことが大切)

  %% 2点目については、ヒューマノイドが全身の出力を活かした作業を行い、長時間に渡りデータを収集するためには、マニピュレーションにおいて環境と一番に接触するハンドの耐久性能が重要となる。
  %% ワイヤ駆動の高出力ハンドを用いてヒューマノイドによる自重保持動作や重量物の保持を実現した研究\cite{makino-hand}のように、作業の種類を増やすためにハンドの耐久性向上が重要となる。
  %% 本研究では、MSL HAND\cite{mslhand}を利用することでこれを解決する。
  \begin{figure}[tb]
    \centering
    \includegraphics[width=0.9\columnwidth]{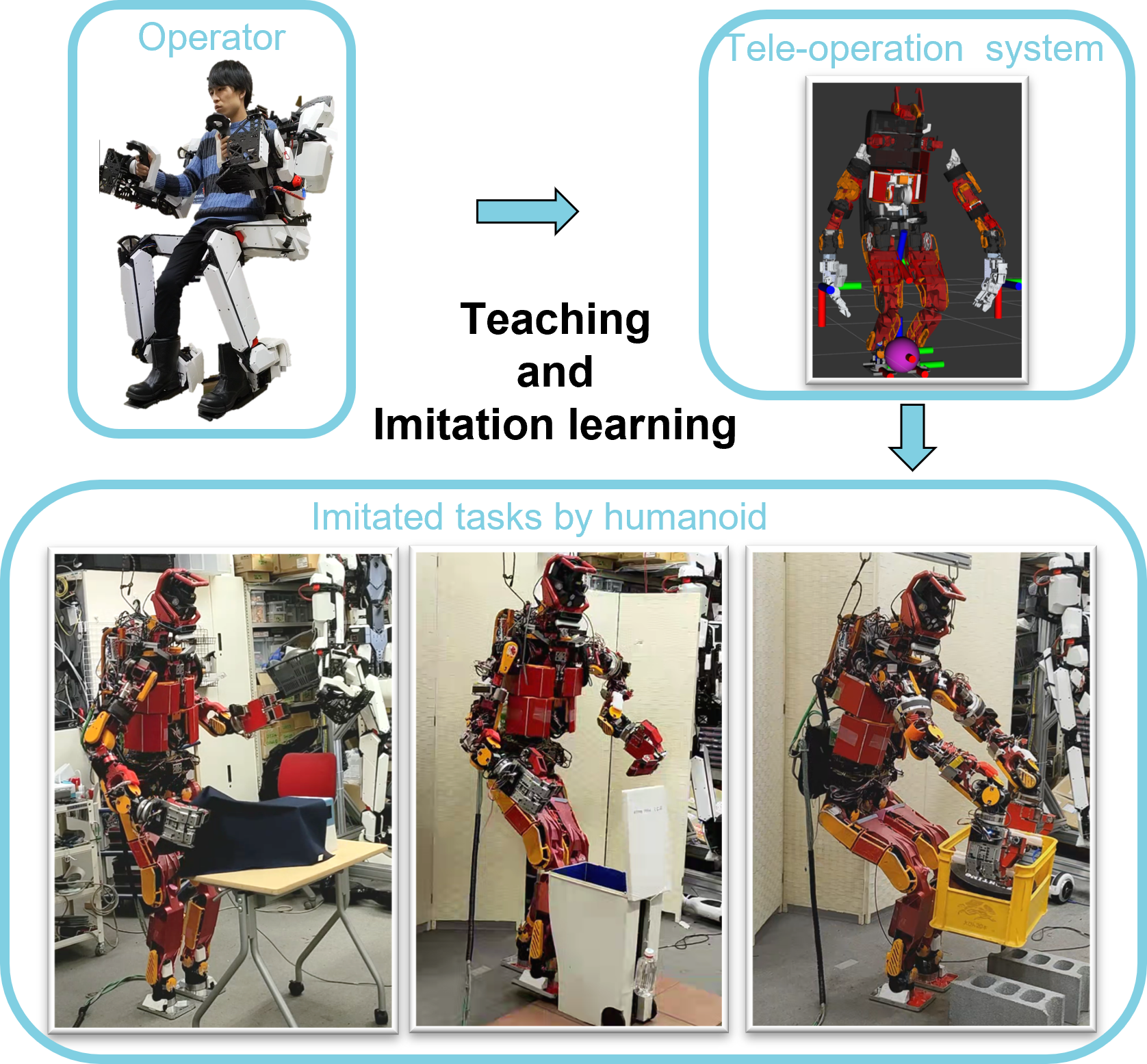}
    \caption{Imitation learning of whole-body tasks by humanoid}
    \label{fig:title}
  \end{figure}

  2点目については, トルクや接触力の最適化に基づく姿勢制御を行うことで重量物の操作と長期的なデータ収集が可能で、なおかつ、ロボットの手だけでなく頭や腰や足も同時に動かせるシステムを構築する.
  ロボットの前で操縦者が片足立ちなどの動作を行い、バランス制約を含む模倣を行う研究\cite{nao-oneleg}や、人の双腕作業の様子を観察し重要な動きの学習と干渉回避などの動作の最適化を経て動作を模倣する研究\cite{asimo-imitation}などが行われている。
  しかし学習上の課題が存在し、環境に対するマニピュレーションを行っていなかったり、短時間の双腕のみのタスク模倣にとどまっている。
  操縦の分野でも重量物を操作している研究は少なく、例えば石黒らの研究\cite{tablis}においても短時間の操作にとどまっている。重量物操作のための学習を行っている例はより少ない。
  そこで、本研究では高負荷作業や長時間のデータ収集に適した全身関節トルク最小化を含む低周期の姿勢生成法と、マニピュレーションタスクのような追従性の求められるタスクのための制約を絞った高周期の姿勢生成を組み合わせる。
  操作対象物体も含めたキーポーズの列を生成する最適化手法\cite{shigematsu}が存在するが、模倣学習のためのデータ収集に焦点を当て、操縦のために制約を絞った最適化計算を実装する。

  上記の2つの特徴をもつヒューマノイドロボットに向けたシステムを用いて、データを収集し模倣学習\cite{kawaharazuka-il}により多様な動作を実現する.
  手だけでなく頭や腰も同時に動くようなヒューマノイドによる風呂敷のマニピュレーション実験、ヒューマノイドに特徴的な脚を使い物体を操作するゴミ箱操作実験、そして手や腰,膝を曲げると同時に大きな力を必要とする重量物体の操作実験を行いこのシステムの有効性を示す。
  }{
  \begin{figure}[thb]
    \centering
    \includegraphics[width=0.95\columnwidth]{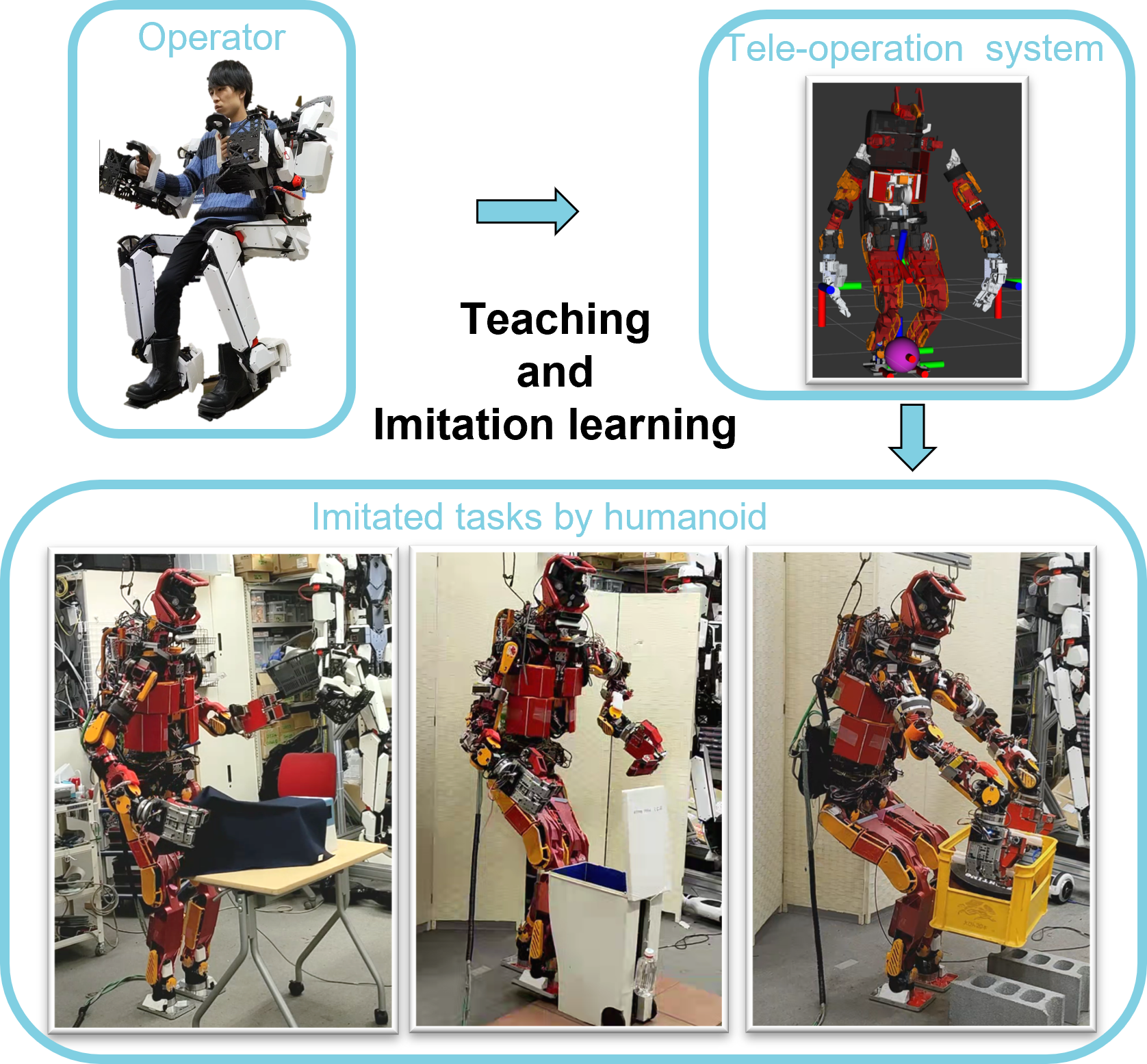}
    \caption{Imitation learning of whole-body tasks by humanoid}
    \label{fig:title}
  \end{figure}

  In the field of teleoperation, there is also limited research on the manipulation of heavy objects. For instance, even in the research by Ishiguro \cite{tablis}, manipulation was limited to short periods. There are even fewer examples of learning for heavy object manipulation. Therefore, in this study, we combine low-cycle posture generation, which includes whole-body joint torque minimization for high-load work and long-term data collection, with high-cycle posture generation, which focuses on the constraints for tasks that require tracking the operator's command posture.
  While there exists an optimization method \cite{shigematsu} that generates a sequence of key poses considering the manipulated objects, this study focuses on data collection for imitation learning and implements an optimization calculation with a narrowed number of constraints for remote piloting.

  Using a system for a humanoid robot with the above two features, we collect data and realize various motions by imitation learning\cite{kawaharazuka-il}.
  We will demonstrate the effectiveness of this system through experiments of manipulation of a flexible cloth, which requires simultaneous movement of the head and waist as well as the hands, manipulation of a trash can using the legs characteristic of humanoids, and manipulation of a heavy object that requires a large force to move the hands, waist, and knees at the same time.

  In recent years, imitation learning has been extensively studied and the acquisition of diverse movements became possible. In particular, various skill acquisitions have been accomplished by dual-armed robots \cite{vr-imitation, himo}. In these studies, the body is fixed, and there is no change in the pose of the robot's root links or the camera angle during operation. Conversely, research on floating-link robots such as humanoid robots remains a challenging task with few studies\cite{Osa_2018}. Therefore, this study aims to develop an imitation learning system for floating-link bipedal robots.
  There are two main problems in developing such a system for a biped humanoid robot with bipedal locomotion.

  The First is the development of a control device capable of operating humanoid robots. The second is a robust control system that can withstand long-term data collection in floating-link systems.
  For the first issue, typically, robots are controlled using VR devices, 3D mice, GUI, and so on. Some studies have utilized dynamic neural network models to learn the object manipulation behavior of small humanoid robots and reproduce appropriate movements according to the environment \cite{small-humanoid-ball-roll-direct-teaching}. Other studies have automatically created models that consider multiple modalities by directly instructing small humanoid robots \cite{nao-multimodal}. However, this direct teaching of actions can be non-intuitive and only achievable in robots with high back drivability. Although simple movements can be replicated using motion-capture-based remote control \cite{hello-motion}, this approach is not adaptable for tasks that require contact force control with the environment due to the lack of haptic feedback to the operator. Approaches to achieving tasks such as folding cloth have been proposed \cite{similar}, but these do not consider using humanoid-specific legs or camera angle variations. Therefore, in this study, we utilize the TABLIS device developed by Ishiguro \cite{tablis}. This device is a cockpit-like exoskeleton control device that can control both arms and legs and achieve bilateral control with a task-executing robot. We connect this TABLIS with the humanoid robot JAXON \cite{JAXON} to construct a control system capable of collecting data for imitation learning.

  For the second issue, we will build a system that can operate heavy objects and collect long-term data based on optimization of torque and contact force, and still be able to move not only the robot's limbs but also the head and waist. simultaneously. Research has been conducted on mimicking human movements that include balance constraints such as standing on one leg \cite{nao-oneleg}, as well as
  cloning human dual-arm work based on the learning of essential movements and optimization of movements such as avoiding collision through observation of human works \cite{asimo-imitation}.
  However, there are some learning issues, and the robot does not manipulate the environment, or only imitates dual-arm tasks for a short period.
}
}

%% \section{RELATED WORKS} \label{sec:related-works}

%% \ifthenelse{\boolean{Draft}}{ %%%%%%% Japanese start %%%%%%%

%%   ロボットの作業模倣は成功しつつある
%%   双腕台車ロボットによるテーブル上での物体の移動\cite{vr-imitation}や,紐結びタスク\cite{himo}%% 、布をたたむタスク\cite{humanoid-folding}
%%   などが実現されている。
%%   これらの研究では基本的に身体が固定されており、作業中にロボットのルートリンクの位置姿勢が変化することは無い。
%%   %% 
%%   一方で、ルートリンクが自由に動き、カメラ画角も変化するヒューマノイドロボットの作業模倣は難しい課題である。
%%   %%
%%   人の作業の様子を映像などで収集し学習するアプローチがある。
%%   ロボットの前で操縦者が片足立ちなどの動作を行い、バランス制約を含む模倣を行う研究\cite{nao-oneleg}や、人の双腕によるマニピュレーションタスクの様子を観察し、重要な動きの学習と干渉回避などの動作の最適化を経て動作を模倣する研究\cite{asimo-imitation}などが行われている。
%%   %% 学習上の課題がある
%%   %%
%%   %% 直接教示
%%   動的ニューラルネットワークモデルであるRNNPB(Recurrent Neural Network with Parametric Bias)を用いて、小型ヒューマノイドロボットの物体操作行動の学習を行い環境に合わせた動作再現を実現した研究\cite{small-humanoid-ball-roll-direct-teaching}や
%%   小型ヒューマノイドで直接教示を行い、複数の行動学習と複数のモダリティからの情報を自動的に考慮したモデルを作成する研究\cite{nao-multimodal}。
%%   これらは動作教示が非直感的であり、なおかつバックドライバビリティが高いロボットでのみしか実現できない。
%%   %%
%%   %% モーキャプによる直感的な操縦による教示
%%   モーションキャプチャを用いた遠隔操縦によるヒューマノイドロボットによる簡単な動作の再現\cite{hello-motion}も実現されているが、環境との接触力フィードバックが不足している。
%%   %% 環境との接触情報の不足
%%   %%
%%   HMDを用いた操縦により1人称視点でのタスクのデータ収集を行い、ヒューマノイドロボットによる布の折りたたみタスクを実現した研究\cite{similar}が存在する。しかしここでは足の利用やカメラ画角の変化は含まれていない。

%% }{ %%%%%% Japanese end: English start %%%%%

%% } %%%%%%% English end %%%%%%%

\ifthenelse{\boolean{Draft}}{ %%%%%%% Japanese start %%%%%%%
  \section{提案するシステム}

}{ %%%%%% Japanese end: English start %%%%%
  \section{SYSTEM IN THIS STUDY} \label{chap:update-posture} \label{chap:update-posture}
}

\ifthenelse{\boolean{Draft}}{ %%%%%%% Japanese start %%%%%%%
  操縦者と操縦デバイスTABLISから実ロボットJAXONまでの全体の構成図を\figref{fig:system}に示す。
  全体システムは大きく2つの構成要素からなる。
  
  1つ目はTABLISを含む模倣学習の作業教示のための遠隔操縦システムである。
  TABLISから取得した操縦者の手先・足先位置姿勢とハンドコントローラのトリガー入力を元にOPG (Opt Pose Gen)とHFPG (High Freq Pose Gen)の2つの姿勢生成器を通してJAXONへ送る全身関節角度列を生成する。
  TABLISの仕組みと全体のデータの流れ、全身制御器での姿勢のフィードバック修正、手元のコントローラのトリガー、ボタン入力の利用に関しては\secref{chap:jaxon-tablis}にて詳細を説明する。
  長期的な動作に貢献するトルク最小化を含んだ姿勢生成法の詳細に関しては\secref{chap:opt-pose-gen}にて説明する。

  2つ目は模倣学習のシステムである。
  操縦システムの\hfpg{}からの出力角度列と、実機からのセンサ値を用いて学習を行い、モデル獲得後の実行時にはロボット状態とセンサ値から次のステップの目標関節角度列を実機へ送信し作業を行う。
  ヒューマノイドへの具体的な適用方法について\secref{chap:imitation-learning}にて述べる。

  本研究では等身大ヒューマノイドJAXONに高耐久ハンドMSL HAND\cite{mslhand}を装着して実験を行った。
  このハンドは、3指5自由度で2本の指の角度を幾何的に固定することが可能である。これにより片手で1000N以上の荷重に耐久し、高負荷作業においても関節負荷上限や温度上昇の心配なく作業を継続することができる。重量箱の持ち上げ実験においてもこのロック機能を使用した。
  大出力の全身関節に加えて環境との接点であるハンドの耐久性も高めることで長時間のデータ収集が可能なハードウェア構成を実現した。

\begin{figure}[thb]
  \centering
  \includegraphics[width=0.99\columnwidth]{figs/all_system_modified_nonitalic.png}
  \caption{全体のシステム構成}
  \label{fig:system}
\end{figure}

}{ %%%%%% Japanese end: English start %%%%%
%%   操縦者と操縦デバイスTABLISから実ロボットJAXONまでの全体の構成図を\figref{fig:system}に示す。
%%   全体システムは大きく2つの構成要素からなる。

%%   1つ目はTABLISを含む作業教示のための遠隔操縦システムである。
%%   これはTABLISから取得した操縦者の手先・足先位置と姿勢とハンドコントローラのトリガー値を入力とする。
%% OPG (Opt Pose Gen)とHFPG (High Freq Pose Gen)の2つの姿勢生成器を通してJAXONの全身関節角度列を生成する。
%%   私達はTABLISの仕組みと全体のデータの流れ、全身制御器での姿勢のフィードバック修正、ハンドコントローラのトリガー、ボタン入力の利用に関しては\secref{chap:jaxon-tablis}にて詳細を説明する。
%%   長期的な動作に貢献するトルク最小化を含んだ姿勢生成法の詳細に関しては\secref{chap:opt-pose-gen}にて説明する。

%%   2つ目は模倣学習のシステムである。
%%   遠隔操縦システムの中の\hfpg{}からの出力角度列と、実機ロボットからのセンサ値を用いて学習を行う。モデル獲得後の作業実行時にはロボット状態とセンサ値から次のステップの目標関節角度列を実機へ送信する。
%%   ヒューマノイドへの具体的な適用方法については\secref{chap:imitation-learning}にて述べる。

%%   本研究では等身大ヒューマノイドJAXONに高耐久ハンドMSL HAND\cite{mslhand}を装着して実験を行った。
%%   このハンドは、3指5自由度で2本の指の角度を幾何的に固定することが可能である。これにより片手で1000N以上の荷重に耐久し、高負荷作業においても関節負荷上限や温度上昇の心配なく作業を継続することができる。重量箱の持ち上げ実験においてもこのロック機能を使用した。
%%   大出力の全身関節に加えて環境との接点であるハンドの耐久性も高めることで長時間のデータ収集が可能なハードウェア構成を実現した。
A diagram of the entire system from the operator and the control device TABLIS to the real robot JAXON is shown in \figref{fig:system}.
  The overall system consists of two major components.

  The first is a teleoperation system for work teaching including TABLIS.
  It takes as input the operator's hand and foot positions and posture obtained from TABLIS and the trigger values of the hand controller.
It generates JAXON's whole body joint angle sequences through two posture generators, OPG (Opt Pose Gen) and HFPG (High Freq Pose Gen).
  We will explain in detail how TABLIS works, the overall data flow, the posture feedback correction in the Whole Body Controller (\wbc{}), the hand controller triggers, and the use of button inputs in \secref{chap:jaxon-tablis}.
  The details of the posture generation method, including torque minimization, which contributes to long-term behavior, are described in \secref{chap:opt-pose-gen}.

  The second is a system of imitation learning.
  The learning is performed using the output angle sequence from the \hfpg{} in the teleoperation system and the sensor values from the actual robot. After acquiring a model, the target joint angle sequence for the next step is sent to the actual robot based on the robot status and sensor values during work execution.
  The specific application of this method to humanoids will be described in \secref{chap:imitation-learning}.

  In this study, we conducted experiments with a life-size humanoid JAXON equipped with a highly durable hand, MSL HAND \cite{mslhand}.
  This hand has three fingers and five degrees of freedom, and the angles of two fingers can be geometrically fixed. This allows the hand to withstand loads of 1000 N or more with one hand%% , and to continue working without concern for joint load limits or temperature rise even in high-load work
  . This locking function was also used in an experiment to lift a heavy box.
  In addition to the high output whole body joints, the durability of the hand, which is the point of contact with the environment, was also increased to achieve a hardware configuration that enables data collection over long periods.

  \begin{figure}[tb]
    \centering
    \includegraphics[width=0.99\columnwidth]{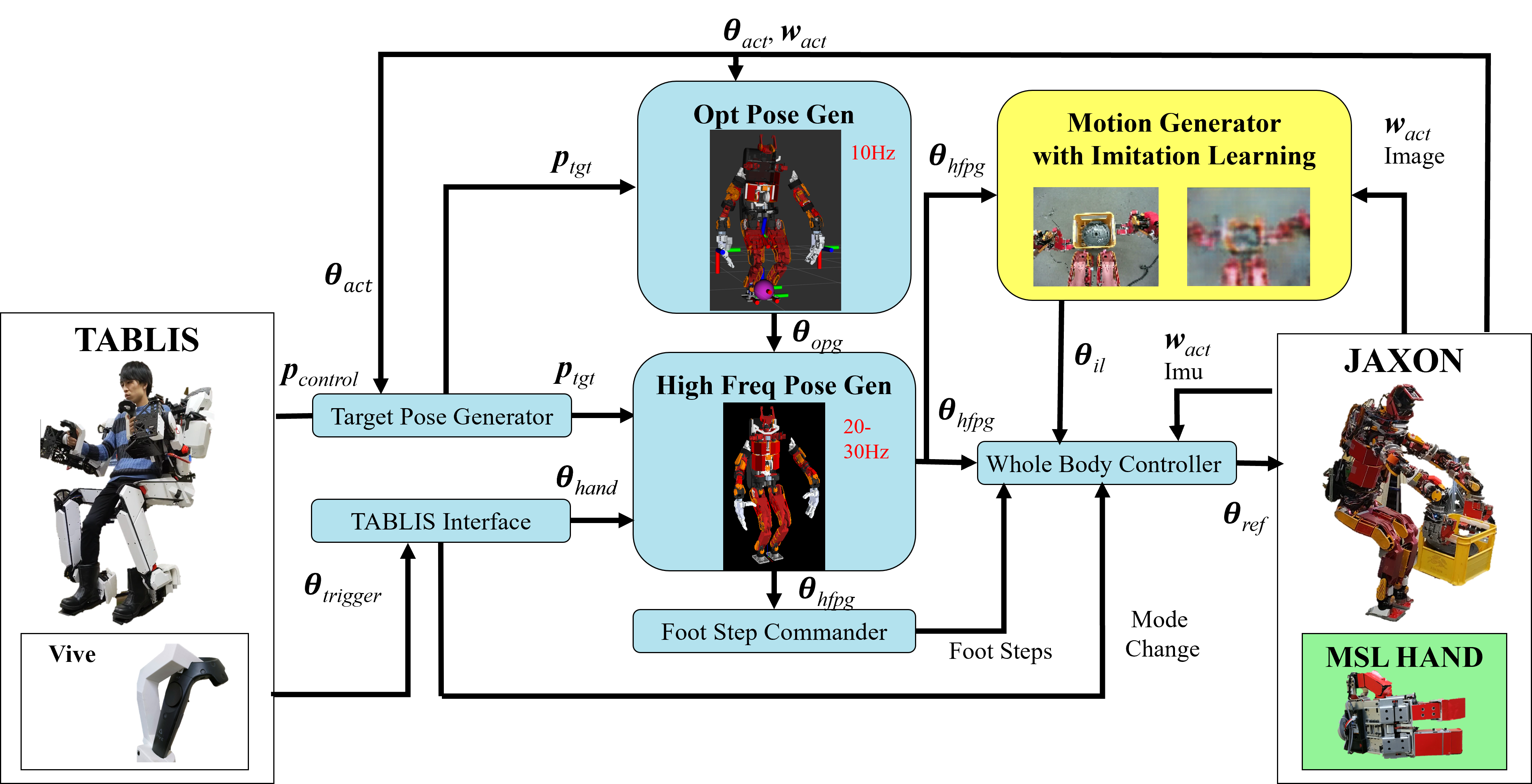}
    \caption{System overview}
    \label{fig:system}
  \end{figure}
} %%%%%%% English end %%%%%%%

\ifthenelse{\boolean{Draft}}{ %%%%%%% Japanese start %%%%%%%
\section{作業教示のためのヒューマノイドの全身操縦システム} \label{chap:jaxon-tablis}
}{ %%%%%% Japanese end: English start %%%%%
  \section{Humanoid Teleoperation System for Whole-body Work Teaching} \label{chap:jaxon-tablis}
}

\ifthenelse{\boolean{Draft}}{ %%%%%%% Japanese start %%%%%%%

  \subsection{操縦システムの構成とデータの流れ}
  教示のための操縦システムの構成を述べる。
  TABLISから取得した手先・足先の位置姿勢情報$\pcont$は、Target Pose Generator (\subsecref{tpg})へ入力され、変換の後に下位へ目標位置姿勢$\ptgt$として出力される。
  \chapref{chap:opt-pose-gen}にて説明する、2つの全身姿勢生成器がこの$\ptgt$を受け取り全身角度列を生成する。
  最終的な$\bthetahfpg$はWhole Body Controller (\subsecref{wbc})へ送られ、実機のレンチ、関節情報をもとに主にバランスを考慮した修正を行い実機への司令とする。
  足の操縦を行う際には、$\bthetahfpg$はFoot Step Commander (\subsecref{fsc})へも送られ、ここで足の高さの差分にもとづく接触状態の切り替え制御が行われる。
  指の操縦司令と教示の際のボタン入力についてはTABLIS Interface (\subsecref{ti})にて制御される。

  \subsection{操縦デバイスTABLIS}
  TABLIS はヒューマノイドのための搭乗型全身外骨格コックピットであり、各腕7自由度,体幹1自由度,各脚6自由度の合計27自由度を持つ。
  本研究では操縦者の手先・足先の4つのエンドエフェクタの位置姿勢6自由度の取得と、実ロボットの力センサ値のフィードバックを利用した。
  TABLISの手先には5個のボタンと1つのトリガーをもつコントローラを搭載した。
  トリガーによりハンドの操縦を行い、ボタンを使ってモードの切り替えなどを制御している。

  \subsection{目標リンク位置姿勢生成器(Target Pose Generator)}\label{tpg}
  TABLISより取得される、手先足先の位置姿勢6自由度$\pcont$を入力とし、実際に全身姿勢生成の計算のための目標リンクの目標位置姿勢$\ptgt$を出力する。
  本システムでは操縦時には開始時からの相対変位を使って操縦を行う。
  操縦開始時に$\pcont$と、実ロボットJAXONの手先足先の位置姿勢を取得し、初期値として保存する。
  操縦開始後は操縦者の初期位置姿勢からの相対的な差分を、TABLISとJAXONの間のプロポーション等をもとに指定する数値で定数倍して目標位置姿勢$\ptgt$を計算する。
  定数値を変更してロボットの動きの大きさを調整可能である。
  また、Tablis Interfaceからの司令で特定のリンクの目標値を固定することで特定リンクの変位を無視したり、操縦が容易なように目標姿勢を回転させたりとタスクに応じた目標リンク位置姿勢の出力ができる。

  \subsection{TABLISとの間のインターフェース(TABLIS Interface)}\label{ti}
  TABLISの手元のコントローラから取得されるトリガーやボタン入力をもとに、ハンドの操縦や作業教示のためのモード切替などを制御する。
  コントローラからの入力と機能を\tabref{table:vive}にまとめた。Lb-1は左手のボタン1、Rt-1は右手のトリガー入力を示す。なお, L/Rは左手と右手, b/tはボタンとトリガーを表している。
  操縦司令の停止時には、\hfpg{} に対して司令を送り実機への司令を停止する。再開時には、その時のTABLISとJAXONの姿勢を初期値として操縦を再開する司令を\opg{},\hfpg{}へ送る。
  指の操縦は、トリガーの1自由度の値を人さし指・中指・親指の関節角度に変換し\hfpg{}にて指以外の関節角度列と統合する。
  作業データ収集のための作業の開始・終了司令もこのノードで制御される。
  \begin{table}[htb]
    \begin{center}
      \caption{button control}
      \footnotesize
      \begin{tabularx}{0.7\columnwidth}{c|c}
        \hline
        button or trigger name & function \\
        \hline
        Lb-1 & pause control \\
        Lb-2 + Lb-3 & reset robot pose\\
        Lt-1 & control left finger angle \\
        Rb-1 & restart control \\
        Rb-2 + Rb-3 & send record trigger\\
        Rt-1 & control right finger angle \\
        \hline
      \end{tabularx}\label{table:vive}
    \end{center}
    \normalsize
  \end{table}

  \subsection{運脚の制御(Foot Step Commander)}\label{fsc}
  実ロボットの足の接触状態変更のためのシステムを説明する。
  %% ヒューマノイドロボットは全てのタスクにおいて、常に全身のバランス制御という大きな制約を考慮する必要がある。
  Foot Step Commanderは、操縦者の足先位置による意図をモード切替に反映するためのコンポーネントである。
  \hfpg{}からの角度列$\bthetahfpg$を取得し足の姿勢を計算する。
  両足の高さの差分に対して閾値を設定し、閾値以上の差分があれば遊脚モードに切り替える。ここでは以下を行う。
  \begin{itemize}
  \item \opg{}の接触力制約の対象リンクを片足or両足に変更
  \item \opg{}の力の釣り合いタスクの対象リンクを変更
  \item 生成姿勢の足の位置情報から足平位置を指定し\wbc{}へ送信
  \end{itemize}

  はじめの２つはrosのtopicで\opg{}に対しトリガーを送る。
  3つ目の足平位置の指定は、生成姿勢の両足の位置を\wbc{}へ送り、\wbc{}の中で片足立ちと両足立ちの際の足の遷移をバランスを考慮して制御している。
  %% モード切替に伴う姿勢の不連続が問題になり得るが、実機へ送る関節角度司令の完了時間を1sと長めに設定することで対応している。
  %% 2足接地、1足接地それぞれの状態での定常的なバランス制御と相互の切り替えフェーズでのバランス制御に大別される。
  %% それぞれの接地状態でのバランス制御部分に関しては、Opt Pose Genにおける接触力制約や重心制約の考慮によりバランスの取れる重心位置を実現する姿勢生成に加えて、Balance ControllerによるFF、FBの姿勢修正により実現している。

  \subsection{全身制御器(Whole Body Controller)における姿勢修正}\label{wbc}
  上位からの全身関節角度列$\bthetahfpg$とロボットからのimu値、力センサ値、関節エンコーダ値を入力とし以下を行う。
  \begin{itemize}
    \item センサ値を元に手先の発揮力の更新
    \item バランス制御器で発揮力と現在の足の位置から重心位置の上書き
    \item 環境接触リンク列の目標位置固定
    \item 足平での発揮力を計算しヤコビアンを使って目標関節トルクを出力
  \end{itemize}
  上記で取得した関節角度から求めた目標関節トルクと、発揮力から得る目標関節トルク、重力補償トルクを足し合わせて電流値に変換しモータへの司令を生成している。
  \fsc{}からの足の切り替え司令に対しては、足の現在位置姿勢と受け取った目標位置姿勢を使って重心位置の遷移を考慮した軌道を生成し足を動かす制御を行う。

  %% \subsection{高耐久ハンドの操縦システム}
  %% ロボットのハンドを除いた全身の姿勢についてはTABLISから取得した手先足先位置を利用して生成する。
  %% マニピュレーション時などに繊細な操縦が求められるハンドに関しての操縦手法については別途検討が必要であった。

  %% viveコントローラは合計5つのボタンを持つデバイスである。現状のTABLISの左右の手先にはハンド操縦用にこのデバイスがそれぞれ搭載されている。
  %% コントローラのトリガーの引き具合に比例してハンドの親指以外の2指の角度を同時に変えることで、ハンドの2指をトリガーの1自由度で操縦する。
  %% トリガーの値はrosのトピックとして出力されており、このトピックを受取りハンド司令に変換するノードを実装した。

  %% \subsection{教示のためのボタン入力}
  %% 他のボタンのON/OFFに応じて、指のロックや、ロボットの姿勢の初期化、操縦のON/OFFなどの機能の切り替えも可能とした。

}{ %%%%%% Japanese end: English start %%%%%

  \subsection{Configuration and data flow of teleoperation system}
  %% 私達は教示のための遠隔操縦システムの構成を述べる。
  %% まず、TABLISから取得した手・足の位置と姿勢$\pcont$は、Target Pose Generator (\subsecref{tpg})へ入力され、目標位置姿勢$\ptgt$として出力される。
  %% \chapref{chap:opt-pose-gen}にて説明する、2つの全身姿勢生成器がこの$\ptgt$を受け取り全身角度列を生成する。
  %% 最終的な$\bthetahfpg$はWhole Body Controller (\subsecref{wbc})へ送られ、実機ロボットのレンチ、関節情報をもとに主にバランスを考慮した修正を行い実機への司令とする。
  %% 足の操縦を行う際には、$\bthetahfpg$はFoot Step Commander (\subsecref{fsc})へも送られ、ここで足の高さの差分にもとづく接触状態の切り替え制御が行われる。
  %% 指の角度司令と教示の際のボタン入力についてはTABLIS Interface (\subsecref{ti})にて制御される。
  We describe the configuration of a teleoperation system for teaching.
  First, the hand and foot positions and posture $\pcont$ obtained from TABLIS are input to the Target Pose Generator (\subsecref{tpg}) and output as the target position posture $\ptgt$.
  The two whole-body pose generators, described in \chapref{chap:opt-pose-gen}, receive this $\ptgt$ and generate a sequence of whole-body angles.
  The final $\bthetahfpg$ is sent to \wbc{} (\subsecref{wbc}), which modifies it based on the wrench and joint information of the actual robot, mainly considering balance, and gives it to the actual robot.
  When performing foot manipulation, $\bthetahfpg$ is also sent to Foot Step Commander (\subsecref{fsc}), where switching control of the contact state based on the difference in foot height is performed.
  Finger angle command and button input for teaching are controlled by TABLIS Interface (\subsecref{ti}).

  \subsection{Whole body control device TABLIS}
  %% TABLIS はヒューマノイドのための搭乗型全身外骨格コックピットであり、各腕7自由度,体幹1自由度,各脚6自由度の合計27自由度を持つ。
  %% 本研究では私達は操縦者の手先・足先の4つのエンドエフェクタの位置姿勢6自由度の取得と、実ロボットの力センサ値のフィードバックを利用した。
  %% 私達はTABLISの手先には5個のボタンと1つのトリガーをもつコントローラを搭載した。
  %% トリガーによりハンドの操縦を行い、ボタンを使ってモードの切り替えなどを行っている。
  TABLIS is a boarding-type full-body exoskeleton cockpit for humanoids and has 7-DOFs for each arm, 1-DOF for the trunk, and 6-DOFs for each leg, totaling 27-DOFs.
  In this research, we used the acquisition of 6-DOFs of position and posture of the four end-effectors in the operator's hands and feet, and the feedback of force sensor values of the real robot.
  We mounted a controller with five buttons and one trigger on the TABLIS hand tip.
  The trigger is used to maneuver the hand, and the buttons serve to switch modes, etc.

  \subsection{Target Pose Generator}\label{tpg} %% Target Link Position and Attitude Generator
  %% \tpg()はTABLISより取得される、手足の位置姿勢6次元$\pcont$を入力とし、実際に全身姿勢生成のための目標位置姿勢$\ptgt$を出力する。
  %% 本システムでは操縦時には開始時からの相対変位を使って操縦を行う。
  %% 操縦開始時に$\pcont$と、JAXONの四肢の位置姿勢を取得し、初期値として保存する。
  %% 開始後は操縦者の初期位置姿勢からの相対的な差分を、TABLISとJAXONの間のプロポーションの違いを反映した数値で定数倍して目標位置姿勢$\ptgt$を計算する。
  %% この定数値によってロボットの動きの大きさを調整可能である。
  %% また、Tablis Interfaceからの司令で特定のリンクの目標値を固定したり、操縦が容易なように目標姿勢を回転させたりとタスクに応じた目標位置姿勢の出力ができる。
  \tpg{} takes as input the 6-dimensional positional posture $\pcont$ of the limbs obtained from TABLIS and outputs the target positional posture $\ptgt$ for the actual whole body posture generation.
  This system uses the relative displacement from the start of the maneuver to perform the maneuver.
  At the start of maneuvering, $\pcont$ and the position posture of JAXON's limbs are acquired and saved as initial values.
  After the start, the target positional posture $\ptgt$ is calculated by multiplying the relative difference from the operator's initial positional posture by a constant value that reflects the difference in proportions between TABLIS and JAXON.
The scale of the robot's movement can be tuned by this constant value.
  In addition, the target position and posture can be output according to the task, such as fixing the target value of a specific link by command from the Tablis Interface or rotating the target posture for easy maneuvering.

  \subsection{TABLIS Interface}\label{ti}
  %% ハンドコントローラから取得されるトリガーやボタン入力をもとに、\ti{}はハンドや作業教示のためのモード切替などを制御する。
  %% コントローラからの入力と機能を\tabref{table:vive}に示した。Lb-1は左手のボタン1、Rt-1は右手のトリガー入力を示す。なお, L/Rは左手と右手, b/tはボタンとトリガーを表している。
  %% \hfpg{} に対して司令を送りロボットへの司令送信を停止する。作業再開のときは私達はその時のTABLISとJAXONの姿勢を初期値として司令送信を再開する司令を\opg{},\hfpg{}へ送る。
  %% 指の操縦は、トリガーの1次元の値を人さし指・中指・親指の3指の関節角度に変換し\hfpg{}にて指以外の関節角度列と統合する。
  %% データ収集の開始・終了司令もこのノードで制御される。
  Based on the trigger and button inputs obtained from the hand controllers, \ti{} controls the hand and mode switching for work teaching.
  We show the inputs from the controller and functions in \tabref{table:vive}, where Lb-1 indicates left-hand button 1 and Rt-1 indicates right-hand trigger input. Note that L/R denotes left and right hand, and b/t denotes button and trigger.
  We send a command to \hfpg{} and stop sending commands to the robot. When resuming work we send commands to \hfpg{},\opg{},\hfpg{} to resume command sending with the initial values of TABLIS and JAXON's posture at that time.
  For finger maneuvering, the one-dimensional value of the trigger is converted to the joint angles of the three fingers (index, middle, and thumb) and integrated with the sequence of joint angles other than the fingers at \hfpg{}.
  The start and end commands for data collection are also controlled by this node.
  \begin{table}[htb]
    \begin{center}
      \caption{Button control}
      \footnotesize
      \begin{tabularx}{0.72\columnwidth}{c|c}
        \hline
        button or trigger name & function \\
        \hline
        Lb-1 & pause control \\
        Lb-2 + Lb-3 & reset robot pose\\
        Lt-1 & control left finger angle \\
        Rb-1 & restart control \\
        Rb-2 + Rb-3 & send record trigger\\
        Rt-1 & control right finger angle \\
        \hline
      \end{tabularx}\label{table:vive}
    \end{center}
    \normalsize
  \end{table}

  \subsection{Foot Step Commander}\label{fsc}
  %% \fsc{}は、足の接触状態変更のためのコンポーネントである。
  %% %% Foot Step Commanderは、操縦者の足先位置による意図をモード切替に反映するためのコンポーネントである。
  %% \hfpg{}からの角度列$\bthetahfpg$を取得し足の姿勢を計算する。
  %% 両足の高さの差分に対して閾値を設定し、閾値以上の差分があれば遊脚モードに切り替える。ここでは以下を行う。
  %% \begin{itemize}
  %% \item \opg{}の接触力制約の対象リンクを片足or両足に変更
  %% \item \opg{}の力の釣り合いタスクの対象リンクを変更
  %% \item 生成姿勢の足の位置情報から足平位置を指定し\wbc{}へ送信
  %% \end{itemize}
  %% %% はじめの２つはrosのtopicで\opg{}に対しトリガーを送る。
  %% 3つ目の機能は、はじめに生成姿勢の両足の位置を\wbc{}へ送り、\wbc{}の中で片足立ちと両足立ちの際の足の遷移をバランスを考慮したうえで制御している。

  \fsc{} is a component for changing the foot contact state.
  It obtains the angle sequence $\bthetahfpg$ from \hfpg{} and calculates the foot posture.
  It sets a threshold value for the difference between the heights of the two feet and switches to the free foot mode if the difference is greater than the threshold value. The following is performed here.
  \begin{itemize}
  \item It changes the target link for the contact force constraint of \opg{} to one foot or both feet.
  \item It changes the target link for the force balancing task of \opg{}.
  \item It Specifies the target foot position from that of the generated posture and send it to \wbc{}.
  \end{itemize}

  \begin{figure}[thb]
    \centering
    \subfigure[OPG]{\includegraphics[height=3.2cm]{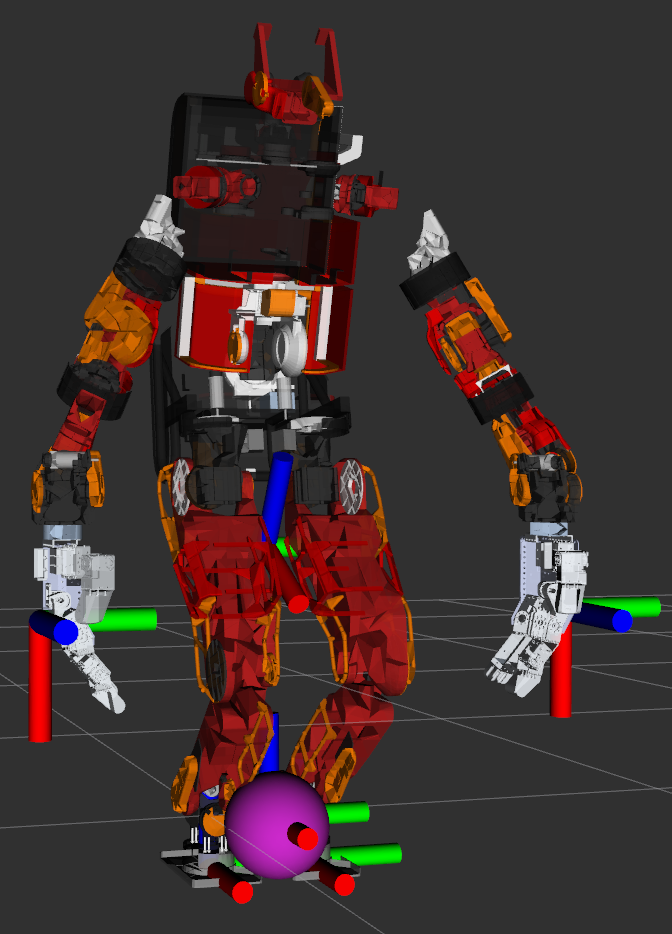} \label{fig:opg}}
    \subfigure[HFPG]{\includegraphics[height=3.2cm]{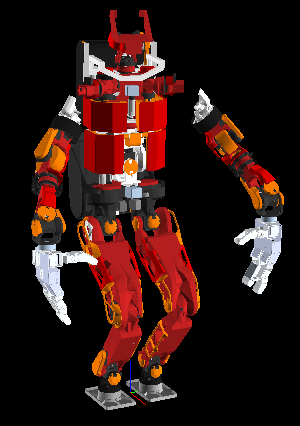}\label{fig:hfpg}}\\
    \caption{Example of result pose of Opt Pose Gen and High Freq Pose Gen}
  \end{figure}

  %% The first two are triggered to \opg{} in the ros topic.
  The third function sends the positions of both feet in the generated posture to \wbc{} at first, and then controls the foot transition between one-leg standing and both-leg standing in \wbc{}, taking balance into consideration.

  \subsection{Whole Body Controller}\label{wbc}
  %% $\bthetahfpg$とロボットからのimu値、力センサ値、関節エンコーダ値を入力とし以下を行う。
  %% \begin{itemize}
  %% \item センサ値を元に手先の目標発揮力の更新
  %% \item 発揮力と現在の足の位置から重心位置の上書き
  %% \item 環境との接触リンクの目標位置固定
  %% \item 足の目標発揮力を計算し目標関節トルクを出力
  %% \end{itemize}
  %% \wbc{}は上記で取得した関節角度から求めた目標関節トルクと、目標発揮力から得る目標関節トルク、重力補償トルクを足し合わせて電流値に変換しモータへの司令を生成している。
  %% \fsc{}からの足の切り替え司令に対しては、足の現在位置姿勢と受け取った目標位置姿勢を使って重心の遷移を考慮した軌道を生成し足を動かす制御を行う。

  The following is performed with $\bthetahfpg$, imu values, force sensor values, and joint encoder values from the robot as inputs.
  \begin{itemize}
  \item It updates the target exerted force of the hand tip based on the sensor values.
  \item It overwrites the center of gravity position based on the exerted force and current foot position.
  \item It fixes the target position of the link in contact with the environment.
  \item It calculates the target exerted force of the foot and output the target joint torque.
  \end{itemize}
  \wbc{} adds together the target joint torque obtained from the joint angle obtained above, the target joint torque obtained from the target exerted force, and the gravity compensation torque, and converts them into a current value to generate a command to the motor.
  For the foot switching command from \fsc{}, the current position posture of the foot and the received target position posture are used to generate a trajectory that takes the transition of the center of gravity into account and controls the foot movement.
}

\ifthenelse{\boolean{Draft}}{ %%%%%%% Japanese start %%%%%%%
  \section{長時間のデータ収集のためのトルク最小化を含む全身姿勢生成} \label{chap:opt-pose-gen}

}{ %%%%%% Japanese end: English start %%%%%
  \section{Whole-body pose generation with torque minimization for long time data collection} \label{chap:opt-pose-gen}
  }

\ifthenelse{\boolean{Draft}}{ %%%%%%% Japanese start %%%%%%%
  \subsection{制御周期の異なる2つの姿勢生成の統合}
  低周期の最適化姿勢生成と制約を絞った高周期の姿勢生成の2段階の姿勢計算を統合することで、全身トルク最小化と操縦性の両立を目指した。
  両者とも\tpg{}の手先足先の目標位置姿勢を受取り姿勢を生成している。
  はじめに\opg{}において複数タスク・制約を考慮した姿勢を計算する(\figref{fig:opg})。およそ5から10hzとなる。
  この姿勢を元に、\hfpg{}により約30Hz程度で手先の目標位置姿勢を用いた上半身の順運動学計算を行い、姿勢を僅かに修正する(\figref{fig:hfpg})。
  これによりマニピュレーションのような手先の追従性が求められる作業に対応する。
  以下ではこの2つの姿勢生成器\opg{}と\hfpg{}の詳細を説明する。
  %% この姿勢生成の際にトルクを考慮した姿勢生成による生成姿勢をもとに計算を開始することで安定性を両立することが可能になる。
\begin{figure}[hb]
  \centering
  \subfigure[Opt Pose Gen]{\includegraphics[height=3.99 cm]{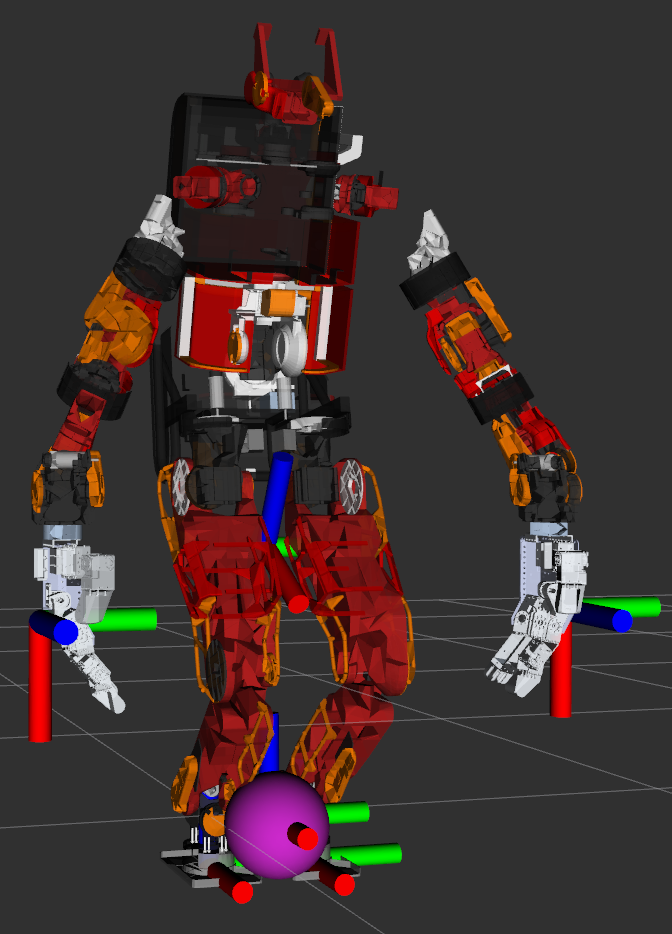} \label{fig:opg}}
  \subfigure[High Freq Pose Gen]{\includegraphics[height=3.99 cm]{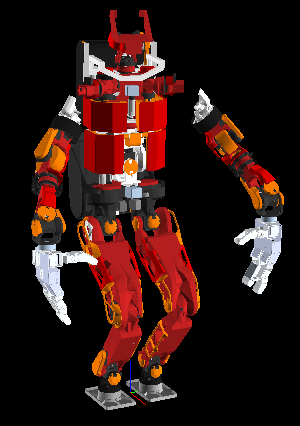}\label{fig:hfpg}}\\
  \caption{Example of result pose of Opt Pose Gen and High Freq Pose Gen}
  %% \label{fig:example-poses}
\end{figure}
  \subsection{安定化のためのトルク最適化を含む姿勢生成法 \\ (Opt Pose Gen)}
最適化計算による、トルク、力の釣り合い、干渉回避を始めとした緒制約を考慮した姿勢の生成方法について詳細を説明する。
生成される姿勢の一例を\figref{fig:opg}に示した。四肢の座標系はそれぞれがエンドエフェクタの目標位置姿勢である。
足平付近のピンク色の球は計算されたロボットの重心位置である。また、ルートリンク付近の座標系はルートリンクの目標姿勢を表している。このリンクでは位置は考慮していない。%% ルートリンクに関しては、姿勢のみに小さな重みをつけてタスクとして追加している。
%% \begin{figure}[hb]
%%   \centering
%%   \includegraphics[width=0.25\columnwidth]{figs/jaxon_tablis_system/opt_pose_gen_box.png}\\
%%   \caption{Example of result pose of Opt Pose Gen}
%%   \label{fig:opt_pose_gen}
%% \end{figure}

\subsubsection{静力学を考慮した逆運動学の最適化問題としての定式化}
まず最適化の方法について説明する。設計するコンフィグレーションを$\q \in \mathbb{R}^{N_q}$ とする。
$N_q$,$N_e$はそれぞれコンフィグレーションとタスクの次元である。
ロボット動作生成問題は、%% 幾何拘束、力・モーメント、関節トルクの釣り合いを含む
タスク関数 $\taskfunction(\q): \mathbb{R}^{N_q} \to \mathbb{R}^{N_e}$ を満たす $\q$ を獲得することと定義される。
\begin{eqnarray}
  \taskfunction(\q) = \bm{0} \label{eq:ik-eq}
  \vspace*{-2mm}
\end{eqnarray}
\equref{eq:ik-eq}は非線形方程式であるため、解析的に解を求めることが困難である。
そこで、一般的には\equref{eq:ik-opt-1}のような最適化問題を考え、数値的な手法を用いて反復計算を行うことで\equref{eq:ik-eq}の最適なコンフィギュレーションを求める。
\begin{subequations}\label{eq:ik-opt-1}
\begin{eqnarray}
  &&\min_{\configuration} \ \function(\configuration) \label{eq:ik-opt-1a} \\
  &&\text{where} \ \ \function(\configuration) \coloneqq  \frac{1}{2} \| \taskfunction(\configuration) \|^2 \label{eq:ik-opt-1b}
  \vspace*{-2mm}
\end{eqnarray}
\end{subequations}
%% さらに、最小の$\configuration_{min}$と最大の$\configuration_{max}$の間に含まれる必要がある場合、逆運動学計算は以下の制約付き非線形最適化問題として表現される。
%% \begin{eqnarray}
%%   &&\min_{\configuration} \  \function(\configuration) \hspace{4mm} \text{ s.t.} \ \  \configuration_{\mathit{min}} \leq \configuration \leq \configuration_{\mathit{max}}\label{eq:ik-opt-2}
%% \end{eqnarray}
%% \equref{eq:ik-opt-2}の制約
これををより一般的な線形等式制約、線形不等式制約として以下のように表現し、制約付き非線形最適化問題として逐次二次計画法により解析的に解くことによりロボットの姿勢を生成する。
\begin{eqnarray}
  \min_{\configuration} \  \function(\configuration) \label{eq:ik-opt-3a}
  &&\text{s.t.} \ \  \bm{A} \configuration = \bm{\bar{b}} \label{eq:ik-opt-3b}\\
  &&\phantom{ \text{s.t.}} \ \  \bm{C} \configuration \geq \bm{\bar{d}}\label{eq:ik-opt-3c}
\end{eqnarray}

\subsubsection{設計変数}
探索変数としては%% ロボットの関節角度とロボットにかかるレンチを採用した。
$\q$は関節角$\jangle$、接触レンチ$\w$から構成される。
ここで、$\jangle$は全身の関節角度に加えてルートリンクの位置姿勢の6次元を加えたものとした。
また$\w$は環境との接触部位ごとの力及びモーメントを並べたベクトルである。
%% トルクを探索変数に入れる最適化手法も提案されている%% \cite{shigematsu}
%% が、本研究では接触レンチからトルクヤコビアンを用いて全身トルクを算出する手法を取った。
$N_{joint}$はロボットの全身関節数、$N_{\eom}$は対象とする接触力目標の数である。
\begin{eqnarray}
  \q \coloneqq \left( \jangle^T \quad \w^T \right)^T \label{eq:q}
 %% \jangle
\end{eqnarray}

\begin{description}[]%% [labelindent=3 cm,labelwidth=50mm]
  \setlength{\itemsep}{-2pt}
\item[$\jangle \in \mathbb{R}^{6+N_\rm{joint}}$] \quad \quad \quad \quad \quad \quad Joint angles [rad] [m]
\item[$\w \in \mathbb{R}^{6N_\rm{\eom}}$] \quad \quad \quad \quad \quad \quad Contact wrench [N] [Nm]
\end{description}

\subsubsection{目的関数}
最適化計算の目的関数を以下のように定義する。
ここでは満たすべき目的関数として、幾何到達目標、接触力に関する力・モーメントの釣り合い、全身のトルクの釣り合い、重心位置目標を考慮した。
$N_{\kin}$は幾何拘束の数を示す。
\begin{eqnarray}
    \taskfunction(\q) \coloneqq \left( \taskfunction^{\rm{\kin}}(\jangle) \quad \taskfunction^{\rm{\eom}}(\jangle,\w) \quad \taskfunction^{\rm{\trq}}(\jangle,\w) \quad \taskfunction^{\rm{\com}}(\jangle)  \right)^T \label{taskdefinition}
\end{eqnarray}

\begin{description}[]%% [labelindent=30mm, labelwidth=50mm]
  \setlength{\itemsep}{-2pt}
\item[$\taskfunction^{\rm{\kin}}(\jangle) \in \mathbb{R}^{6N_{\rm{\kin}}}$] \quad \quad \quad \quad \quad \quad Kinematic task [rad] [m]
\item[$\taskfunction^{\rm{\eom}}(\jangle,\w) \in \mathbb{R}^{6N_\rm{\eom}}$] \quad \quad \quad \quad \quad \quad Wrench task [N] [Nm]
\item[$\taskfunction^{\rm{\trq}}(\jangle,\w) \in \mathbb{R}^{N_{\rm{joint}}}$] \quad \quad \quad \quad \quad \quad Torque task [Nm]
\item[$\taskfunction^{\rm{\com}}(\jangle) \in \mathbb{R}^{3}$] \quad \quad \quad \quad \quad \quad COM position task [m]
\end{description}

\subsubsection{幾何到達目標}
幾何到達目的関数は\equref{eq:kin1},\equref{eq:kin2}のようになる。
\begin{eqnarray}
  \taskfunction^{\kin} (\jangle) &\coloneqq& \begin{pmatrix} \taskfunction^{\rm{\kin}}_1\\ \taskfunction^{\rm{\kin}}_2 \\ .\\ .\\ .\\ \taskfunction^{\kin}_{N_{\rm{tgt}}} \end{pmatrix}\label{eq:kin1}\\
  \taskfunction_m^{\rm{\kin}} &\coloneqq& K_{\rm{weight}} * \begin{pmatrix} \p_m^{\rm{tgt}} - \p_m^{\rm{tgt\_cur}} \\ a(\R_m^{\rm{tgt}} * {\R_m^{\rm{tgt\_cur}}}^\mathrm{T}) \end{pmatrix} \label{eq:kin2}
\end{eqnarray}
ここで、$\taskfunction^{\kin}_{i}$は目的リンクごとの目標位置姿勢と目的リンクの現在の位置姿勢の誤差を示している。
$\p_m^{tgt}$ はm番目の目標位置、$\p_m^{tgt\_cur}$はm番目の目的リンクの現在位置、$\R_m^{tgt}$はm番目の目標姿勢行列、${\R_m^{tgt\_cur}}$はm番目の目的リンクの現在姿勢行列である。
$a(\R)$ は回転行列$\R$の角軸ベクトルを取り出す関数である
$K_{weight}$はm個の目標リンクそれぞれの1次元の重みであり、ここでの$*$はアダマール積である。リンクごとに違う重み値を設定することで、その相対的な大小関係によってm個のリンクの内でどのリンクの位置姿勢の誤差を小さくし、どのリンクの位置姿勢の誤差を許容するかを決めることができる。
%% 例えば、マニピュレーションタスクなどにおいては、両手の重みを大きくし、肘などの重みを小さめに設定することで、手先は目標位置姿勢に対しより正確に追従し、肘の位置姿勢誤差はある程度許容するといった運用が可能となる。
本研究での幾何目標は、四肢の位置姿勢に加えてタスクに応じてルートリンクと頭リンクの姿勢を与えた。
ルートリンクについては、姿勢のみ、小さな重みをつけて追加している。これはロボット姿勢の対称性に寄与する。
頭リンクに関しても姿勢のみに対して目標を追加した。%% これはカメラの方向を正面に固定するために使用される。

\subsubsection{力・モーメントの釣り合い}
$\taskfunction^{\rm{\eom}}(\jangle,\w)$をわかりやすさのために\equref{eq:force1}の力の誤差関数$\taskfunction^{\rm{\eom force}}(\jangle)$と\equref{eq:moment2}のモーメント$\taskfunction^{\rm{\eom moment}}(\jangle,\w)$の項に分けて示した。
\begin{eqnarray}
  \taskfunction^{\rm{\eom force}} (\w) &\coloneqq& \sum_{m=1}^{N_{\rm{tgt}}} \f_{m} - M\g \label{eq:force1}\\
  \taskfunction^{\rm{\eom moment}} (\jangle, \w)%%  &=& \sum_{m=1}^{N_{tgt}} \{(\p_{m}^{tgt} - \p^{root}) \times \f_{m} + \n_m \} \nonumber \\
  %% & & +\quad   (\p^{com} - \p^{root}) \times (-M\g) \label{eq:moment1}\\
  %% &=& \sum_{m=1}^{N_{tgt}} \{ [\f_{m} \times](\p_{m}^{tgt} - \p^{root}) + \n_m \} \\
  %% & & +\quad [(-M\g) \times] (\p^{com} - \p^{root})
  &\coloneqq& \sum_{m=1}^{N_{tgt}} \{ [(\p_{m}^{tgt} - \p^{root}) \times] \f_{m}+ \n_m \}\nonumber\\
  & & +\quad [(-M\g) \times] (\p^{com} - \p^{root})\label{eq:moment2}
\end{eqnarray}
ここで、$\f_m$はm番目の目的リンクにおける発揮力、$n_m$はm番目の目的リンクにおける発揮モーメント、$M$はロボットの重量[kg]、$g$は重力加速度[m/s/s]、$\p_m^{tgt}$ はm番目の目標位置、$\p^{root}$はロボットのルートリンクの現在位置、$\p^{root}$はロボットの重心の現在位置である。
%% 力・モーメントの釣り合いに関しては、レンチ6次元の各次元ごとに重みをかけることで例えばz軸方向の力に対する最適化を重視するといった運用が可能である。

\subsubsection{トルクの釣り合い}
駆動トルクと必要トルクの釣り合いの式を\equref{eq:torque}に示した。
\begin{eqnarray}
  \taskfunction^{\trq} &\coloneqq& \btau^{ref} - \btau(\jangle, \w)\nonumber \label{eq:torque} \\
  &=& \btau^{ref} - \btau^{grav}(\jangle) + \btau^{cnt}(\jangle, \w)
\end{eqnarray}

ここで、$\btau^{ref}$は目標関節トルク、$\btau^{grav}$は重力による自重トルク、$\btau^{cnt}$は接触力によるトルクを示す。
$\btau^{ref}$としては0ベクトルを採用することで、可能な限り全身の発揮トルクを0に近づける最適化を行うことができる。

%% ここで$\tau_{grav}$と$\tau_{cnt}$は以下の式で示される。
%% \begin{eqnarray}
%%   \tau^{grav} &=& ( \sum_{k=1}^{N_{joint}} m_k {\J_{cog,k}}^\mathrm{T}) \g\\
%%   %% \J_{cog,k} &=& (\J_{cog,k}^1 \quad \J_{cog,k}^2 \quad.\quad.\quad. \J_{cog,k}^{N_{joint}})\\
%%   \J_{cog,k} &=& (\J_{cog,k}^1 \quad \J_{cog,k}^2 \quad.\quad.\quad. \J_{cog,k}^{N_{joint}})\\
%%   \J_{cog,k}^i &=& \begin{Bmatrix} \bm{a}_{\theta_i} \times (\p_{cog,k} - \p_{\theta_i}) \quad (link_k depends on link_i)\\ 0 \end{Bmatrix}\\
%%   \tau^{cnt} &=& \sum_{m=1}^{N_{trq}} {\J_{i,m}^{cnt-tgt}}^\mathrm{T} \w_m
%% \end{eqnarray}
%% $\J_{cog,k}$は関節角度に対するリンク$k$の重心位置へのヤコビアンを、$\J_{i,m}^{cnt-tgt}$は関節iの変化に対するm番目の接触レンチ$\w_m$の発生する接触リンク位置姿勢の変化を表すヤコビアンを示す。
%% $\J_{cog,k}^i$はリンクiの原点からリンクkの重心位置へのリンクi座標系でのヤコビアンを示す。リンクkがリンクiの子リンクで無い場合は0とする。

\subsubsection{重心位置目標}
重心位置目標の式を\equref{eq:com-tgt}に示した。
\begin{eqnarray}
  \taskfunction^{\com} (\jangle) &\coloneqq& \p_{\com}^{tgt} - \p_{\com}^{cur}\label{eq:com-tgt}
\end{eqnarray}
ここで$\p_{\com}^{tgt}$は重心の目標位置、$\p_{\com}^{cur}$は現在の重心位置に対応する。
この目標タスクに関しても他と同様に重みをつけることができ、重みを小さすることで姿勢や発揮力の対称性を向上させる役割を持つ。
%% 逆に重みを大きくすると所望の位置に重心を調整することが可能である。

\subsection{考慮した制約条件}
%% \section{contact\_objective}
制約条件としては、以下を考慮した。
それぞれに対して重みを設定することができ、相対的な重みの大きさを調整することで、各制約の重要度を調整し姿勢を調整することができる。

\subsubsection{関節角度上下限}
関節上下限の制約式を\equref{eq:minmax-angle}に示す。
\begin{eqnarray}
  \jangle_{\rm{min}} \leqq \jangle \leqq \jangle_{\rm{max}} \label{eq:minmax-angle}
\end{eqnarray}
ロボットのハードウェア的な角度上下限とバランス制御のための姿勢修正のための角度余裕を考慮して上下限値($\jangle_{\rm{min}},\jangle_{\rm{max}}$)を設定した。
タスクに応じて体幹3軸や肘、膝の関節角度を絞り込むことで無理のない姿勢生成が可能となる。
%% 例えば、片腕を使った間にピュレーション時には体幹のyaw軸の可動範囲を狭く設定することで体幹の動きを抑えた。使用するヒューマノイドロボットは自由度が多く、同じように右腕を体の前に出す動作でも、体幹のyaw軸、腕の付け根のカラーyaw軸のように重複した回転軸があり、どちらを動かしても良い場面がある。他にも同様のことは腕の方のyaw軸と上腕のyaw軸においても起こる。このような場合は、タスクの性質によって関節角度に制限を設けることでどちらの軸を動かすかをある程度設定できる。例えば対象物を常にカメラの画角内に収めたいといった要求がある場合には体幹の動きを抑制し腕の動きを促進するような方法が考えられる。
%% また膝の伸び具合の調整にも有効である。トルク最小化タスクの効果として2足直立時には膝を伸ばし切る姿勢が生成されやすい。%% しかし、膝が伸び切る姿勢は特異点に近くIK的には不利な姿勢であり、歩行などの足を動かす場合には向いていない。
%% そこで膝関節の角度制限により膝の伸び切りを防ぐといった方法を取ることができる。

\subsubsection{関節トルク上下限}
関節トルク上下限の制約式を\equref{eq:minmax-torque}に示す。
\begin{eqnarray}
  \btau_{\rm{min}} \leqq \btau \leqq \btau_{\rm{max}}\label{eq:minmax-torque}
\end{eqnarray}
ハードウェア的制限であるモータのラチェッティングトルクにマージンをかけたもの($\btau_{\rm{min}},\btau_{\rm{max}}$)を設定した。ハーモニックギヤのギヤ飛びを抑制することが可能となる。

\subsubsection{干渉回避制約}
干渉回避制約式を\equref{eq:collision}に示す。
\begin{eqnarray}
  \bm{d}_{margin} \leqq || \p_1 - \p_2 || \label{eq:collision}
\end{eqnarray}
指定した２つのリンク間の距離をマージン以上にする制約を設定した。
$\bm{d}_{margin}$は設定する余裕、$\p_1$,$\p_2$は2つのリンクの最近傍点の位置である。%% $\bm{d}_{12}$は$\p_1$から$\p_2$へ向かう単位ベクトルであり、$\J_{\theta, 1}$と$\J_{\theta, 2}$は$\theta$の変化に対する$\p_1$と$\p_2$の変化率を表すヤコビアンである。$\bm{k}_{collision}$は定数でありこの値を調整することで更新時に干渉が発生する可能性を調整できる。

\subsubsection{ルートリンクの高さ制約}
ルートリンクの高さに対しての制約式を\equref{eq:root-link}に示す。
\begin{eqnarray}
   \p_z^{root} + p_{offset} \leqq \min \,(\p_z^{r-hand},\p_z^{l-hand}) \label{eq:root-link}
\end{eqnarray}
$\p_z^{root}$はルートリンク位置のzの値、$p_{offset}$は設定する高さオフセット、$\p_z^{r-hand}$,$\p_z^{l-hand}$は両手の位置のzの値である。
%% 膝を使ってかがみ込むような動作を生成したときに追加すると良い。
基本的にはトルク最適化タスクと対立するタスクとなるので、両者の重みの相対的な大きさの設定が重要である。%% トルクタスクの重みのほうが十分に大きい場合は、手を下げてもギリギリまで膝を伸ばした姿勢を生成するのに対して、ルートリンク高さに関するこのタスクの重みを大きくすると、膝を曲げてかがみ込む動作が出やすくなる。
今回は両手の位置姿勢を使った操縦手法を採用しているため両手の高さを重心を下げる基準とした。
%% 両手の高さの最小値に変わって、両手の高さの平均値などを用いることもできる。

\subsubsection{足平の接触力制約}
%% \begin{eqnarray}
%%   d_w \leqq A_w \w
%% \end{eqnarray}
%% \begin{eqnarray}
%%   0 &\leqq& -f_x + \mu_x f_z \leqq \infty\\
%%   0 &\leqq& -f_y + \mu_y f_z \leqq \infty\\
%%   0 &\leqq& (b_{+}-1) f_z \leqq \infty\\
%%   0 &\leqq& l_y f_z - n_x \leqq \infty\\ %% \quad (n_x<0)\\
%%   0 &\leqq& l_x f_z - n_y \leqq \infty\\ %% \quad (n_y<0)\\
%%   %% 0 &\leqq& \infty f_z - n_z \leqq \infty\\ %
%%   %% -\infty &\leqq& - n_z \leqq \infty\\ %
%%   0 &\leqq& \mu_z f_z - n_z \leqq \infty\\ %
%%   0 &\leqq& f_x + \mu_x f_z \leqq \infty\\
%%   0 &\leqq& f_y + \mu_y f_z \leqq \infty\\
%%   0 &\leqq& (b_{-}+1) f_z \leqq \infty\\
%%   0 &\leqq& l_y f_z + n_x \leqq \infty\\ %% \quad (n_x \geqq0)\\
%%   0 &\leqq& l_x f_z + n_y \leqq \infty\\  %% \quad (n_y\geqq0)\\
%%   %% 0 &\leqq& \infty f_z + n_z \leqq \infty\\
%%   %% -\infty &\leqq&  n_z \leqq \infty\\
%%   0 &\leqq& \mu_z f_z + n_z \leqq \infty \label{eq:contact-each}
%% \end{eqnarray}
制約式としては\equref{eq:contact-matrix}を考えた。
%% これを行列を用いて表現すると以下のようになる。

\begin{eqnarray}
  0 \leqq \A \R^\mathrm{T} \w %% \leqq \infty
  \label{eq:contact-matrix}
\end{eqnarray}
%% \begin{eqnarray}
%%   \A = \begin{pmatrix} -1 & 0 & \mu_x & 0 & 0 & 0 \\
%%     0 & -1 & \mu_y & 0 & 0 & 0 \\
%%     0 & 0 & b_{+}-1 & 0 & 0 & 0 \\
%%     0 & 0 & l_y & -1 & 0 & 0 \\
%%     0 & 0 & l_x & 0 & -1 & 0 \\
%%     %% 0 & 0 & \infty & 0 & 0 & -1 \\
%%     0 & 0 & \mu_z & 0 & 0 & -1 \\
%%     1 & 0 & \mu_x & 0 & 0 & 0 \\
%%     0 & 1 & \mu_y & 0 & 0 & 0 \\
%%     0 & 0 & b_{-}+1 & 0 & 0 & 0 \\
%%     0 & 0 & l_y & 1 & 0 & 0 \\
%%     0 & 0 & l_x & 0 & 1 & 0 \\
%%     0 & 0 & \mu_z & 0 & 0 & 1
%%     %% 0 & 0 & \infty & 0 & 0 & 1
%%   \end{pmatrix},
%%   \w = \begin{pmatrix}
%%     f_x\\
%%     f_y\\
%%     f_z\\
%%     n_x\\
%%     n_y\\
%%     n_z
%%   \end{pmatrix}
%% \end{eqnarray}
%% \begin{eqnarray}
%%   b_{-} = 0\\
%%   b_{+} = 1%% \infty
%% \end{eqnarray}

%% 参考\\
%% 変形元の式
%% \begin{eqnarray}
%%   |f_{x,y}| \leqq \mu_{x,y} f_z\\
%%   -l_{x,y} f_z \leqq |n_{y,x}| \leqq l_{x,y} f_z
%% \end{eqnarray}
足裏に適用する接触力に対する制約条件式を示す。
$\A$は12行の行列であり、足裏のx,y軸方向の力を静止摩擦力以内にする条件、z方向の力を正の値とする条件、重心位置を指定範囲に収める条件などが含まれる。$\R$はエンドエフェクタへの姿勢行列であり、転置して$/w$に左からかけることで対象とするレンチの座標系を変換している。
%% $f_x$,$f_y$,$f_z$,$n_x$,$n_y$,$n_z$は世界座標系におけるそれぞれ各方向の力と各軸回りのモーメントを示す。$l_x$,$l_y$は重心位置の足裏2次元平面内の存在可能範囲に対応し、$\mu_x$,$\mu_y$,$\mu_x$は摩擦係数を示す。ここで$l_x$,$l_y$は足裏サイズに余裕を取った値を使用し、$\mu_x$yaw軸回りのモーメントを抑制するために設定している。
%% $n_z$に関しては回転摩擦係数を仮定して制約をかした。\\
%% $b_{-},b_{+}$については$0<f_z$となるように調整した。

%% \subsubsection{手先の接触力制約}
%% 手先での環境接触のための制約式に相当する。式としては上述の足平の接触力制約と同等である。
%% ロボットのエンドエフェクタリンクの座標系の調整のために回転行列を追加して座標変換を行っている点と、接触範囲が異なる。
%% 具体的には、雲梯動作のようなハンドによるぶら下がりを想定した。
%% 実際に検証実験においてはこの制約を追加しぶら下がりを行った。

\subsubsection{エンドエフェクタ発揮力制約}
エンドエフェクタにおいて、所望の発揮力を満たす姿勢を生成するための制約を\equref{eq:tgt-wrench}示す。
\begin{eqnarray}
  %% \w_{ref} \leqq \w \leqq \w_{ref}\label{eq:tgt_wrench}
  \w_{ref} = \w_{ref}\label{eq:tgt-wrench}
\end{eqnarray}
$\w_{ref}$は手先と足先の目標発揮レンチを表す。
各リンクの6軸レンチの各値に対して重みを設定できる。
各目標リンクの重みを調整することで、考慮したいリンクを変更できる。
%% 実際には0か1の値を設定し、実用時は常に4肢のリンクを目標リンクとしておき、発揮力を指定する時は重みを1,指定しないときには0と設定することでこの制約を有効化、無効化する運用となっている。
%% 重みが1のときは\equref{eq:tgt_wrench}のとおりに与えられる目標レンチを上下限として設定する。
%% 重みが0のときには、下限を$-\infty$上限を$\infty$とすることで制約を無効化する。その際は、他のタスクもしくは制約に基づき発揮力が最適化されることとなる。
%% %% 例えば、マニピュレーションタスクにおいては両手の重みを1,両足の重みを0とすることで両手の発揮力は上位から制御可能となり、両足は両手の発揮力を考慮して、接触力制約や力の釣り合いタスクにより最適化された値となる。

\subsection{高周期の姿勢生成(High Freq Pose Gen)}
ここでは、上位から両手先の位置姿勢を、\opg{}から全身関節角度列を取得し、上半身の逆運動学をとき関節角度を修正する。
\hfpg{}により、マニピュレーションタスクのような手先の操縦に対する追従性が求められる場面での操縦性を保証する。
対象関節には体幹と両腕の関節を含み、足関節は対象としない。
そのため両足関節角度列とルートリンクの姿勢には\opg{}のものがそのまま採用され、上半身の関節角度のみが僅かに修正される。
タスクに応じて30Hz程度で計算が行われる。
%% 対象としない関節に関しては、定周期の姿勢生成の結果姿勢を使用する。
生成される姿勢の一例を\figref{fig:hfpg}に示す。
%% \begin{figure}[hb]
%%   \centering
%%   \includegraphics[width=0.25\columnwidth]{figs/jaxon_tablis_system/high_freq_pose_gen_box.png}\\
%%   \caption{Example of result pose of High Freq Pose Gen}
%%   \label{fig:high_freq_ik}
%% \end{figure}

}{ %%%%%% Japanese end: English start %%%%%

  \subsection{Integration of two posture generators with different control cycles}
  %% 低周期の最適化姿勢生成と制約を絞った高周期の姿勢生成の2段階の姿勢計算を統合することで、全身トルク最小化と操縦性の両立を目指した。
  %% 両者とも\tpg{}の四肢の目標位置姿勢を受取り全身関節角度を生成している。
  %% はじめに\opg{}において複数タスク・制約を考慮した関節角度を計算する(\figref{fig:opg})。周期はおよそ5から10hzとなる。
  %% この角度列を元に、\hfpg{}により約30Hz程度で手先の目標位置姿勢を用いた上半身の順運動学計算を行い、角度を僅かに修正する(\figref{fig:hfpg})。
  %% これによりマニピュレーションのような手先の細かい操作が求められる作業に対応する。
  %% 以下ではこの2つの姿勢生成器\opg{}と\hfpg{}の詳細を説明する。
  By integrating two stages of pose calculation, low-cycle optimized pose generation and constraint-focused high-cycle pose generation, we aimed to achieve both whole-body torque minimization and maneuverability.
  Both of them receive the target positional posture of the limbs in \tpg{} and generate the whole-body joint angles.
  First, the joint angles are computed in \opg{}, taking into account multiple tasks and constraints (\figref{fig:opg}). The cycles are approximately 5 to 10 hz.
  Based on this angle sequence, a forward kinematics calculation of the upper body using the target positional posture of the hand tip is performed at about 30 Hz by \figref{fig:hfpg}, and the angles are slightly modified (\figref{fig:hfpg}).
  This corresponds to tasks such as manipulation, which require fine manipulation of the hand tip.
  The details of these two pose generators are explained in the following sections.
  % \begin{figure}[thb]
  %   \centering
  %   \subfigure[OPG]{\includegraphics[height=3.2cm]{figs/jaxon_tablis_system/opt_pose_gen_box.png} \label{fig:opg}}
  %   \subfigure[HFPG]{\includegraphics[height=3.2cm]{figs/jaxon_tablis_system/high_freq_pose_gen_box.png}\label{fig:hfpg}}\\
  %   \caption{Example of result pose of Opt Pose Gen and High Freq Pose Gen}
  % \end{figure}

  \subsection{Whole body posture generation including torque optimization (\opg{})}
  %% ここでは最適化計算による、トルク、力の釣り合い、干渉回避を始めとした緒制約を考慮した姿勢の生成方法について詳細を説明する。
  %% 生成される姿勢の一例を\figref{fig:opg}に示した。四肢の座標はそれぞれがエンドエフェクタの目標位置姿勢である。
  %% 足付近のピンク色の球は計算されたロボットの重心位置である。また、ルートリンク付近の座標はルートリンクの目標姿勢を表している。このリンクでは位置は考慮していない。
  This section describes in detail how to generate postures by optimization calculations, taking into account the torque, force balance, collision avoidance, and other constraints.
  An example of the generated posture is shown in \figref{fig:opg}. The coordinates of each limb are the target position posture of the end-effector.
  The pink sphere near the feet is the calculated center of gravity of the robot. The coordinates near the root link represent the target posture of the root link. The position is not considered for this link.

  \subsubsection{Formulation as an inverse kinematics optimization problem considering statics}
  %% まず私達は最適化の方法について説明する。設計するコンフィグレーションを$\q \in \mathbb{R}^{N_q}$ とする。
  %% $N_q$,$N_e$はそれぞれコンフィグレーションとタスクの次元である。
  %% ロボット動作生成問題は、
  %% タスク関数 $\taskfunction(\q): \mathbb{R}^{N_q} \to \mathbb{R}^{N_e}$ を満たす $\q$ を獲得することと定義される。
  First, we describe the optimization method. Let $\q \in \mathbb{R}^{N_{\rm{q}}}$ be the configuration to be designed.
  $N_{\rm{q}}$ and $N_{\rm{e}}$ are the dimensions of the configuration and task, respectively.
  The robot motion generation problem is
  defined as obtaining a task function $\taskfunction(\q): \mathbb{R}^{N_{\rm{q}}} \to \mathbb{R}^{N_{\rm{e}}}$ satisfying $\q$.
  \begin{eqnarray}
    \taskfunction(\q) = \bm{0} \label{eq:ik-eq}
    \vspace*{-2mm}
  \end{eqnarray}
  %% \equref{eq:ik-eq}は非線形方程式であるため、解析的に解を求めることが困難である。
  %% そこで、一般的には\equref{eq:ik-opt-1}のような最適化問題を考え、数値的な手法を用いて反復計算を行うことで\equref{eq:ik-eq}の最適なコンフィギュレーションを求める。
  Since \equref{eq:ik-eq} is a nonlinear equation, it is difficult to find a solution analytically.
  Therefore, we generally consider an optimization problem such as \equref{eq:ik-opt-1} and find the optimal configuration of \equref{eq:ik-eq} by performing iterative calculations using numerical methods.
  \begin{subequations}\label{eq:ik-opt-1}
    \begin{eqnarray}
      &&\min_{\configuration} \ \function(\configuration) \label{eq:ik-opt-1a} \\
      &&\text{where} \ \ \function(\configuration) \coloneqq  \frac{1}{2} \| \taskfunction(\configuration) \|^2 \label{eq:ik-opt-1b}
      \vspace*{-2mm}
    \end{eqnarray}
  \end{subequations}
  %% 私達はこれをより一般的な線形等式制約、線形不等式制約として以下のように表現する。
  %% この式を制約付き非線形最適化問題として逐次二次計画法により解析的に解くことによりロボットの姿勢を生成する。
  We express this as more general linear equality and linear inequality constraints as follows.
  We generate the robot posture by analytically solving this equation as a constrained nonlinear optimization problem using sequential quadratic programming.
  \begin{eqnarray}
    \min_{\configuration} \  \function(\configuration) \label{eq:ik-opt-3a}
    &&\text{s.t.} \ \  \bm{A} \configuration = \bm{\bar{b}} \label{eq:ik-opt-3b}\\
    &&\phantom{ \text{s.t.}} \ \  \bm{C} \configuration \geq \bm{\bar{d}}\label{eq:ik-opt-3c}
  \end{eqnarray}

  \subsubsection{Design variables}
  %% 探索変数$\q$は関節角$\jangle$と接触レンチ$\w$から構成される。
  %% ここで、$\jangle$は全身の関節角度に加えてルートリンクの位置姿勢の6次元を加えたものとした。
  %% また$\w$は環境との接触リンクごとの力及びモーメントを並べたベクトルである。
  %% $N_{joint}$は全身関節数、$N_{\eom}$は接触力目標の数である。
  The search variable $\q$ consists of the joint angle $\jangle$ and the contact wrench $\w$.
  Here, $\jangle$ is the joint angle of the whole body plus 6 dimensions of the position and posture of the root link.
  Also, $\w$ is a vector of forces and moments for each link in contact with the environment.
  $N_{\rm{joint}}$ is the number of whole body joints and $N_{\rm{eom}}$ is the number of contact force targets.
  \begin{eqnarray}
    \q \coloneqq \left( \jangle^T \quad \w^T \right)^T \label{eq:q}
    %% \jangle
  \end{eqnarray}

  \begin{description}[]%% [labelindent=3 cm,labelwidth=50mm]
    \setlength{\itemsep}{-2pt}
  \item[$\jangle \in \mathbb{R}^{6+N_{\rm{joint}}}$] \quad \quad \quad \quad \quad \quad Joint angles [rad] [m]
  \item[$\w \in \mathbb{R}^{6N_{\rm{\eom}}}$] \quad \quad \quad \quad \quad \quad Contact wrench [N] [Nm]
  \end{description}

  \subsubsection{Objective function}
  %% 最適化計算の目的関数を以下のように定義する。
  %% ここでは私達は満たすべき目的関数として、幾何到達目標、力・モーメントの釣り合い、全身のトルクの釣り合い、重心位置目標を考慮した。
  %% $N_{\kin}$は幾何拘束の数を示す。
  The objective function of the optimization calculation is defined as follows.
  Here we considered the kinematic goal, the force-moment balance, the whole-body torque balance, and the center-of-gravity position goal as objective functions to be satisfied.
  The $N_{\rm{\kin}}$ denotes the number of kinematic constraints.
  \begin{eqnarray}
    \taskfunction(\q) \coloneqq \left( \taskfunction^{\rm{\kin}}(\jangle) \quad \taskfunction^{\rm{\eom}}(\jangle,\w) \quad \taskfunction^{\rm{\trq}}(\jangle,\w) \quad \taskfunction^{\rm{\com}}(\jangle)  \right)^T \label{taskdefinition}
  \end{eqnarray}

  \begin{description}[]%% [labelindent=30mm, labelwidth=50mm]
    \setlength{\itemsep}{-2pt}
  \item[$\taskfunction^{\rm{\kin}}(\jangle) \in \mathbb{R}^{6N_{\kin}}$] \quad \quad \quad \quad \quad \quad Kinematic task [rad] [m]
  \item[$\taskfunction^{\rm{\eom}}(\jangle,\w) \in \mathbb{R}^{6N_\eom}$] \quad \quad \quad \quad \quad \quad Wrench task [N] [Nm]
  \item[$\taskfunction^{\rm{\trq}}(\jangle,\w) \in \mathbb{R}^{N_{\rm{joint}}}$] \quad \quad \quad \quad \quad \quad Torque task [Nm]
  \item[$\taskfunction^{\rm{\com}}(\jangle) \in \mathbb{R}^{3}$] \quad \quad \quad \quad \quad \quad COM position task [m]
  \end{description}

  \subsubsection{Kinematic objective function}
  %% 幾何到達目的関数は\equref{eq:kin1},\equref{eq:kin2}のようになる。
  The kinematic objective functions are as in \equref{eq:kin1},\equref{eq:kin2}.
  \begin{eqnarray}
    %% \taskfunction^{\kin} (\jangle) &\coloneqq& \begin{pmatrix} \taskfunction^{\kin}_1\\ \taskfunction^{\kin}_2 \\ .\\ .\\ .\\ \taskfunction^{\kin}_{N_{tgt}} \end{pmatrix}\label{eq:kin1}\\
    \taskfunction^{\rm{\kin}} (\jangle) &\coloneqq& (\taskfunction^{\rm{\kin}}_1, \taskfunction^{\rm{\kin}}_2, . . . , \taskfunction^{\rm{\kin}}_{N_{\rm{tgt}}} \label{eq:kin1})^T\\
    \taskfunction_m^{\rm{\kin}} &\coloneqq& K_{\rm{weight}} * \begin{pmatrix} \p_m^{\rm{tgt}} - \p_m^{\rm{tgt\_cur}} \\ a(\R_m^{\rm{tgt}} * {\R_m^{\rm{tgt\_cur}}}^\mathrm{T}) \end{pmatrix} \label{eq:kin2}
  \end{eqnarray}
  %% $\taskfunction^{\kin}_{i}$は注目リンクの目標位置姿勢と現在の位置姿勢の誤差を示している。
  %% $\p_m^{tgt}$ はm番目の目標位置、$\p_m^{\rm{tgt\_cur}}$はm番目の目的リンクの現在位置、$\R_m^{tgt}$はm番目の目標姿勢行列、${\R_m^{\rm{tgt\_cur}}}$はm番目の目的リンクの現在姿勢行列である。
  %% $a(\R)$ は回転行列$\R$の角軸ベクトルを取り出す関数である

  %% $K_{weight}$は目標リンクそれぞれの1次元の重みであり、ここでの$*$はアダマール積である。リンクごとに違う重み値を設定することで、どのリンクの誤差を小さくするかを決めることができる。
  %% 私達は、本研究での幾何目標として四肢の位置姿勢を考え、タスクに応じてルートリンクと頭リンクの姿勢を与えた。
  %% ルートリンクについては、私達は姿勢のみに対して小さな重みをつけて追加している。これはロボット姿勢の対称性に寄与する。
  %% 頭リンクに関しても姿勢のみに対して重みを与えた。
  The $\taskfunction^{\rm{\kin}}_{i}$ denotes the error between the target position-posture and the current position-posture of the link of interest.
  $\p_m^{\rm{tgt}}$ is the mth target position, $\p_m^{\rm{tgt\_cur}}$ is the current position of the mth objective link, $\R_m^{\rm{tgt}}$ is the mth target posture matrix and ${\R_m^{\rm{tgt\_cur}}}$ is the current posture matrix of the mth objective link.
  $a(\R)$ is a function to extract the angular axis vector of the rotation matrix $\R$.

  $K_{\rm{weight}}$ is the one-dimensional weight of each target link, where $*$ is the Adamar product. By setting different weight values for each link, it is possible to determine which link has a smaller error.
  We considered the positional posture of the limbs as the kinematic goal in this study and gave the posture of the root link and the head link according to tasks.
  For the root link and head link, we add a small weight for the posture only. %% This contributes to the symmetry of the robot's posture.
  %% For the head link, we also gave weights to only the posture.

  \subsubsection{Balance of forces and moments}
  %% $\taskfunction^{\eom}(\jangle,\w)$をわかりやすさのために\equref{eq:force1}の力の誤差関数$\taskfunction^{eom force}(\jangle)$と\equref{eq:moment2}のモーメントの誤差関数$\taskfunction^{eom moment}(\jangle,\w)$の項に分けて示した。
  For simplicity $\taskfunction^{\rm{\eom}}(\jangle,\w)$ was replaced by the error function of the force $\taskfunction^{\rm{eom force}} (\jangle)$ in \equref{eq:force1} and the moment error function $\taskfunction^{\rm{eom moment}}(\jangle,\w)$  in \equref{eq:moment2}.
  \begin{eqnarray}
    \taskfunction^{\rm{\eom force}} (\w) &\coloneqq& \sum_{m=1}^{N_{\rm{tgt}}} \f_{m} - M\g \label{eq:force1}\\
    \taskfunction^{\rm{\eom moment}} (\jangle, \w)%%  &=& \sum_{m=1}^{N_{\rm{tgt}}} \{(\p_{m}^{\rm{tgt}} - \p^{root}) \times \f_{m} + \n_m \} \nonumber \\
    &\coloneqq& \sum_{m=1}^{N_{\rm{tgt}}} \{ [(\p_{m}^{\rm{tgt}} - \p^{\rm{root}}) \times] \f_{m}+ \n_m \}\nonumber\\
    & & +\quad [(-M\g) \times] (\p^{\rm{com}} - \p^{\rm{root}})\label{eq:moment2}
  \end{eqnarray}
  %% $\f_m$はm番目の目的リンクにおける目標発揮力、$n_m$は目標発揮モーメント、$M$はロボットの重量[kg]、$g$は重力加速度[m/s/s]、$\p_m^{\rm{tgt}}$ はm番目の目標位置、$\p^{\rm{root}}$はロボットのルートリンクの現在位置、$\p^{\rm{root}}$はロボットの重心の現在位置である。
  $\f_m$ is the target exerted force at the mth objective link, $n_m$ is the target exerted moment, $M$ is the robot weight [kg], $g$ is the gravitational acceleration [m/s/s], $\p_m^{\rm{tgt}}$ is the mth target position, $\p^{\rm{root}}$ is the current position of the robot root link, $\p^{\rm{root}}$ is the current position of the robot's center of gravity.

  \subsubsection{Balance of whole body joint torque}
  %% 駆動トルクと必要トルクの釣り合いの式を\equref{eq:torque}に示した。
  The equation for the balance between the drive torque and the required torque is shown in \equref{eq:torque}.
  \begin{eqnarray}
    \taskfunction^{\trq} &\coloneqq& %% \btau^{\rm{ref}} - \btau(\jangle, \w)\nonumber
    %% &=&
    \btau^{\rm{ref}} - \btau^{\rm{grav}}(\jangle) + \btau^{cnt}(\jangle, \w)    \label{eq:torque}
  \end{eqnarray}
  %% $\btau^{\rm{ref}}$は目標関節トルク、$\btau^{\rm{grav}}$は重力による自重トルク、$\btau^{cnt}$は接触力によるトルクを示す。
  %% $\btau^{\rm{ref}}$としては0ベクトルを採用することで、可能な限り全身の発揮トルクを0に近づける最適化を行うことができる。
  $\btau^{\rm{ref}}$ is the target joint torque, $\btau^{\rm{grav}}$ is the self-weight torque by gravity, and $\btau^{cnt}$ is the torque by contact force.
  By adopting a 0 vector as $\btau^{\rm{ref}}$, we can optimize the whole body exerted torque as close to 0 as possible.

  \subsubsection{Center of gravity objective function}
  %% 重心位置目標の式を\equref{eq:com-tgt}に示した。
  The equation for the center-of-gravity position target is shown in \equref{eq:com-tgt}.
  \begin{eqnarray}
    \taskfunction^{\rm{\com}} (\jangle) &\coloneqq& \p_{\rm{com}}^{\rm{tgt}} - \p_{\rm{com}}^{cur}\label{eq:com-tgt}
  \end{eqnarray}
  %% $\p_{\rm{\com}}^{\rm{tgt}}$は重心の目標位置、$\p_{\rm{\com}}^{cur}$は現在の重心位置に対応する。
  %% 私達はこの目標タスクに関しても他と同様に重みを設定することができ、重みを小さすることで姿勢や発揮力の対称性を向上させる役割を持つ。
  $\p_{\rm{com}}^{\rm{tgt}}$ corresponds to the target position of the center of gravity and $\p_{\rm{com}}^{cur}$ to the current center of gravity position.
  We can set weights for this target task as well as others, and smaller weights have the role of improving the symmetry of posture and exerted force.

  \subsection{Constraints}
  %% 制約条件としては、以下を考慮した。
  %% それぞれに対して重みを設定することができ、相対的な重みの大きさを調整することで、各制約の重要度を調整し姿勢を変えるすることができる。
  As constraints, we considered the following.
  We can set weights for each of them, and by adjusting the relative magnitude of the weights, we can adjust the importance of each constraint and change the posture.

  \subsubsection{Upper and lower limits of joint angle}
  %% 関節上下限の制約式を\equref{eq:minmax-angle}に示す。
  The constraint equation for the joint upper and lower limits is shown in \equref{eq:minmax-angle}.
  \begin{eqnarray}
    \jangle_{\rm{min}} \leqq \jangle \leqq \jangle_{\rm{max}} \label{eq:minmax-angle}
  \end{eqnarray}
  %% ハードウェア的な上下限とバランス制御のための角度余裕を考慮して上下限値($\jangle_{min},\jangle_{max}$)を設定した。
  %% タスクに応じて体幹3軸や肘、膝の関節角度を絞り込むことで無理のない姿勢生成が可能となる。
  The upper and lower limits ($\jangle_{\rm{min}},\jangle_{\rm{max}}$) are set considering the upper and lower limits of hardware and the angle margin for balance control.

  \subsubsection{Upper and lower limits of joint torque}
  %% 関節トルク上下限の制約式を\equref{eq:minmax-torque}に示す。
  The constraint equation for the joint torque upper and lower limits is shown in \equref{eq:minmax-torque}.
  \begin{eqnarray}
    \btau_{\rm{min}} \leqq \btau \leqq \btau_{\rm{max}} \label{eq:minmax-torque}
  \end{eqnarray}
  %% モータのラチェッティングトルクにマージンをかけたものを$\btau_{min},\btau_{max}$として設定した。ハーモニックギヤの歯飛びを抑制することが可能となる。
  The motor ratcheting torque with a margin is set as $\btau_{\rm{min}},\btau_{\rm{max}}$. It is possible to suppress tooth skipping of harmonic gears.

  \subsubsection{Collision avoidance constraints}
  %% 干渉回避制約式を\equref{eq:collision}に示す。
  The collision avoidance constraint equation is shown in \equref{eq:collision}.
  \begin{eqnarray}
    \bm{d}_{\rm{margin}} \leqq || \p_1 - \p_2 || \label{eq:collision}
  \end{eqnarray}
  %% 私達は指定した２つのリンク間の距離をマージン以上にする制約を設定した。
  %% $\bm{d}_{\rm{margin}}$は余裕、$\p_1$,$\p_2$は2つのリンクの最近傍点の位置である。
  We set the constraint to make the distance between the two specified links greater than the margin.
  $\bm{d}_{\rm{margin}}$ is the margin and $\p_{\rm{1}}$,$\p_{\rm{2}}$ are the positions of the nearest neighbor points of the two links.

  \subsubsection{Height of root link constraints}
  %% ルートリンクの高さに対しての制約式を\equref{eq:root-link}に示す。
  The constraint equation for the height of the root link is shown in \equref{eq:root-link}.
  \begin{eqnarray}
    \p_z^{\rm{root}} + p_{\rm{offset}} \leqq \min \,(\p_z^{\rm{r-hand}},\p_z^{\rm{l-hand}}) \label{eq:root-link}
  \end{eqnarray}
  %% $\p_z^{\rm{root}}$はルートリンク位置のzの値、$p_{\rm{offset}}$は高さオフセットである。$\p_z^{\rm{r-hand}}$,$\p_z^{\rm{l-hand}}$は両手の位置のzの値である。
  %% 基本的にはトルク最適化タスクと対立するタスクとなるので、両者の重みの値が重要である。
  %% 今回は両手の位置姿勢を使った操縦をしているため両手の高さを重心を下げる基準とした。
  $\p_z^{\rm{root}}$ is the z value of the root link position and $p_{\rm{offset}}$ is the height offset. $\p_z^{\rm{r-hand}}$ and $\p_z^{\rm{l-hand}}$ are the z values of both hand positions.
  The values of both weights are important because they are basically in conflict with the torque optimization task.
  In this case, the height of both hands was used as the criterion for lowering the center of gravity since the maneuver was performed using the positional posture of both hands.

  \subsubsection{Foot contact force constraints}
  %% 足の接触力に対する制約条件式を\equref{eq:contact-matrix}に示す。
  The constraint equation for the foot contact force is shown in \equref{eq:contact-matrix}.
  \begin{eqnarray}
    0 \leqq \A \R^\mathrm{T} \w %% \leqq \infty
    \label{eq:contact-matrix}
  \end{eqnarray}
  %% $\A$は12行の行列であり、足裏のx,y軸方向の力を静止摩擦力以内にする条件、z方向の力を正の値とする条件、重心位置を指定範囲に収める条件などが含まれる。$\R$はエンドエフェクタへの姿勢行列であり、転置して$/w$に左からかけることで対象とするレンチの座標を変換している。
  $\A$ is a 12-row matrix, which includes the conditions that the forces in the x- and y-axis of the sole should be within the static friction force, that the force in the z-direction should be positive, and that the center of gravity should be within the specified range.
  $\R$ is the orientation matrix to the end-effector, which is transposed and multiplied by $\w$ from the left to transform the coordinate of the target wrench.

  \begin{figure}[thb]
    \centering
    \includegraphics[width=0.99\columnwidth]{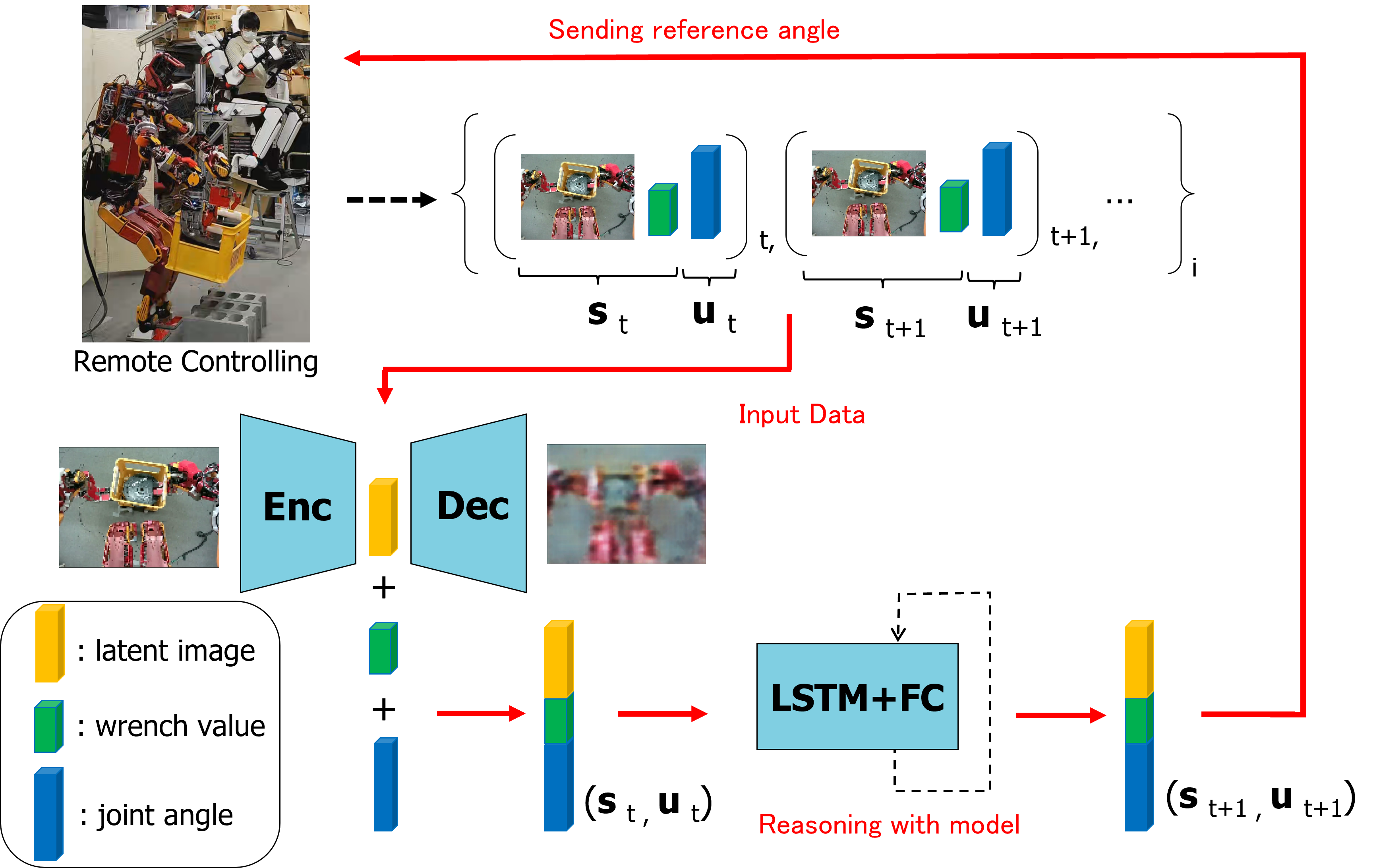}\\
    \caption{Network structure and whole system of imitation learning}
    \label{fig:network}
  \end{figure}

  \subsubsection{End-effector exertion force constraints}
  %% エンドエフェクタにおいて、所望の発揮力を満たす姿勢を生成するための制約を\equref{eq:tgt-wrench}示す。
  Constraints for generating a posture that satisfies the desired exerted force at end effectors are shown in \equref{eq:tgt-wrench}.
  \begin{eqnarray}
    %% \w_{\rm{ref}} \leqq \w \leqq \w_{\rm{ref}}\label{eq:tgt_wrench}
    \w = \w_{\rm{ref}}\label{eq:tgt-wrench}
  \end{eqnarray}
  %% $\w_{\rm{ref}}$は四肢の目標発揮レンチを表す。
  %% あなたは各リンクの6軸レンチの各値に対して重みを設定できる。
  %% 各重みを調整することで、考慮したいリンクを変更できる。
  $\w_{\rm{ref}}$ denotes the target exerted wrench of the limb.
  You can set weights for each value of the 6-axis wrench for each link.
  By adjusting each weight, the link to be considered can be changed.

  \subsection{High frequency pose generation (\hfpg{})}
  %% \hfpg{}は\tpg{}から取得した両手先の位置姿勢と、\opg{}から取得した全身関節角度列から、上半身の逆運動学をとき関節角度を修正する。
  %% \hfpg{}により、マニピュレーションタスクのような手先の細かい操縦が求められる場面での操縦性を保証する。
  %% 対象関節には体幹と両腕の関節を含み、足は対象としない。
  %% タスクに応じて30Hz程度の周期で計算が行われる。
  %% 生成される姿勢の一例を\figref{fig:hfpg}に示す。
  \hfpg{} modifies the joint angles based on the inverse kinematics of the upper body using the positional posture of both limbs obtained from \tpg{} and the whole body joint angle sequence from \opg{}.
  \hfpg{} ensures maneuverability in situations such as manipulation tasks, which require detailed maneuvering of hands.
  The target joints include the joints of the trunk and both arms, but not the legs.
Calculations are performed at a period of about 30 Hz, depending on each task.
  An example of the generated posture is shown in \figref{fig:hfpg}.
}

\ifthenelse{\boolean{Draft}}{ %%%%%%% Japanese start %%%%%%%
  \section{ヒューマノイドロボットのための模倣学習}\label{chap:imitation-learning}
  提案した操縦システムを用いて作業を教示し、模倣学習\cite{kawaharazuka-il}によりスキルを獲得する。

  \subsection{模倣学習手法の概要}
  \equref{eq:network}に学習の基本式を示す。また、\figref{fig:network}に全体のシステムを示す。
  ここでは$t$は現在のタイムステップ、$\bms$はロボットの感覚状態、$\bmu$はロボットへの制御入力、$\bmp$はパラメトリックバイアス、$\bmh$はネットワークの重み$W$を含む予測モデルを表している。
本研究では$\bms$はロボットのカメラ画像や力センサ値、$\bmu$は全身の関節角度列とした。
  全10層で4層は全結合(FC)層、2層のLSTM層、4層の全結合層からなる再帰型ネットワークである。
  活性化関数はHyperbolic Tangentで更新則はAdamを採用している。
  $\bms$と$\bmu$は正規化した値を使用し、2次元のパラメトリックバイアス$\bmp$を使用している。
  \begin{eqnarray}
    (\bms_{t+1}, \bmu_{t+1}) = \bmh(\bms_t, \bmu_t, \bmp) \label{eq:network}
  \end{eqnarray}

  \begin{figure}[htb]
    \centering
    \includegraphics[width=0.99\columnwidth]{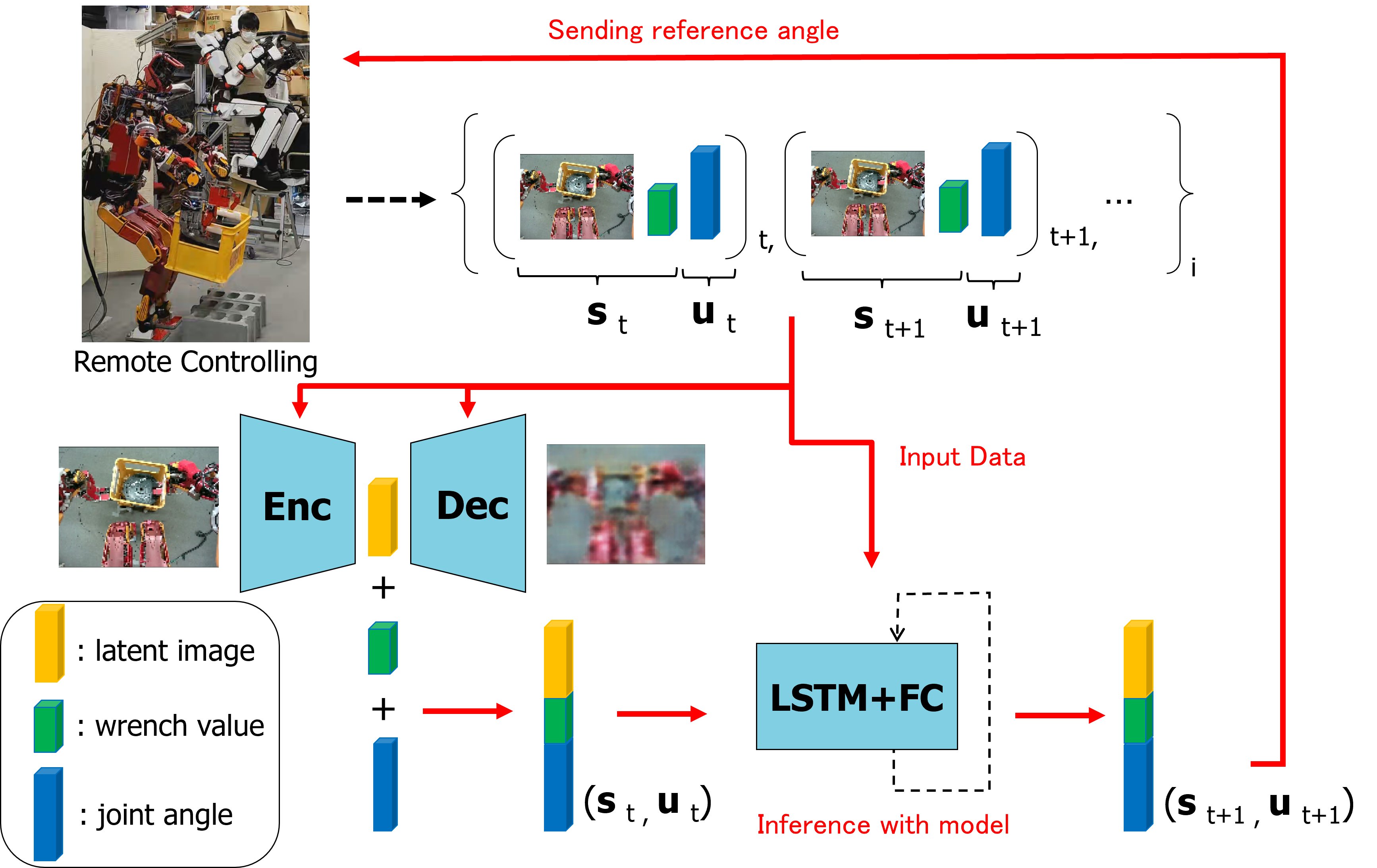}\\
    \caption{模倣学習のネットワーク構造と全体システム}
    \label{fig:network}
  \end{figure}

  \subsection{教示と学習}
  はじめに、$\bms$と$\bmu$の時系列データを収集する。本研究ではTABLIS\cite{tablis}と\chapref{chap:jaxon-tablis}で提案したシステムを利用した教示によるデータ収集を行う。
  試行のたびに対象環境を変化させたタスク実現を複数回行いデータを収集するが、それぞれの試行における時系列データの組を$D_k = \{(\bms_1,\bmu_1),(\bms_2,\bmu_2), ..., (\bms_{T_k},\bmu_{T_k})\}$($1 \leqq k \leqq K$、Kは全試行回数、$T_k$は各試行における動作ステップ数）とする。
画像についてはAuto Encoderを用いて事前に学習して次元削減・特徴抽出を行い、128x96のRGB画像を12次元の特徴量に圧縮して学習に用いた。
これにパラメトリックバイアスを加えた$D_{train} = \{ (D_1, \bmp_1), (D_2, \bmp_2), ...,  (D_K, \bmp_K)\}$を学習データとして用いる。ここで$\bmp_k$は各ダイナミクス（試行）毎の値であり、操縦者の動作スタイルが埋め込まれる。このデータを用いてモデルを学習させ、ネットワークの重み$W$とともに$\bmp_k$を更新する。
$\bmp_k$の初期値は０とし損失関数は平均二乗誤差を使用した。

}{ %%%%%% Japanese end: English start %%%%%

    \section{Imitation learning for humanoid robots}\label{chap:imitation-learning}
    %% 提案した操縦システムを用いて作業を教示し、模倣学習\cite{kawaharazuka-il}によりスキルを獲得する。
    The proposed system is used for work teaching, and the skills are acquired through imitation learning \cite{kawaharazuka-il}.

    \subsection{Overview of imitation learning methods}
    %%   \equref{eq:network}に学習の基本式を示す。また、\figref{fig:network}に全体のシステムを示す。
    %%   ここでは$t$は現在のタイムステップ、$\bms$はロボットの感覚状態、$\bmu$はロボットへの制御入力、$\bmp$はパラメトリックバイアス、$\bmh$はネットワークの重み$W$を含む予測モデルを表している。
    %% 本研究では$\bms$はロボットのカメラ画像や力センサ値、$\bmu$は全身の関節角度列とした。
    %%   全10層で4層は全結合(FC)層、2層のLSTM層、4層の全結合層からなる再帰型ネットワークである。
    %%   活性化関数はHyperbolic Tangentで更新則はAdamを採用している。
    %%   $\bms$と$\bmu$は正規化した値を使用し、2次元のパラメトリックバイアス$\bmp$を使用している。
    The basic formula for learning is shown in \equref{eq:network}. We also show the overall system in \figref{fig:network}.
    Here $t$ represents the current time step, $\bms$ the sensory state of the robot, $\bmu$ the control input to the robot, $\bmp$ the parametric bias, and $\bmh$ the prediction model including the network weight $W$.
    In this study, $\bms$ is the camera images and force sensor values of the robot, and $\bmu$ is the joint angle sequences of the whole body.
    It is a recursive network with 10 layers in total, 4 of which are the Fully Connected (FC) layer, 2 of which are the Long Short Term Memory (LSTM) layers\cite{hochreiter1997lstm}, and 4 of which are FC layers.
    The activation function is Hyperbolic Tangent and the optimizer is Adam \cite{kingma2015adam}.
    The $\bms$ and $\bmu$ are normalized values and a two-dimensional parametric bias $\bmp$ is used.
    \begin{eqnarray}
      (\bms_{t+1}, \bmu_{t+1}) = \bmh(\bms_t, \bmu_t, \bmp) \label{eq:network}
    \end{eqnarray}

    % \begin{figure}[thb]
    %   \centering
    %   \includegraphics[width=0.99\columnwidth]{figs/learning_system_0213.png}\\
    %   \caption{Network structure and whole system of imitation learning}
    %   \label{fig:network}
    % \end{figure}
    \subsection{Teaching and Learning}
  %%   まず私達は$\bms$と$\bmu$の時系列データを収集する。本研究では\chapref{chap:jaxon-tablis}で提案したシステムを利用した教示によるデータ収集を行う。
  %%   対象環境を変化させたタスク実現を複数回行いデータを収集する.
  %%   それぞれの試行における時系列データの組を$D_k = \{(\bms_1,\bmu_1),(\bms_2,\bmu_2), ..., (\bms_{T_k},\bmu_{T_k})\}$($1 \leqq k \leqq K$、Kは全試行回数、$T_k$は各試行における動作ステップ数）とする。
  %% 画像についてはAuto Encoderを用いて事前に学習して次元削減・特徴抽出を行う。128x96のRGB画像を12次元のベクトルに圧縮して学習に用いた。
  %% 私達はこれらのベクトルにパラメトリックバイアスを加えたベクトルを、学習データとして用いる。
  %% $\bmp_k$は各ダイナミクス毎の値であり、操縦者の動作スタイルが埋め込まれる。このデータを用いてモデルを学習させ、ネットワークの重み$W$とともに$\bmp_k$を更新する。
  %% $\bmp_k$の初期値は０とし損失関数は平均二乗誤差を使用した。
    First, we collect $\bms$ and $\bmu$ time sequence data. In this study, we collect data by teaching using the system proposed in \chapref{chap:jaxon-tablis}.
    Data are collected through multiple task realizations in which the target environment is changed.
    The pairs of time sequence data in each trial are defined as $D_k = \{(\bms_1,\bmu_1),(\bms_2,\bmu_2), ..., (\bms_{T_k},\bmu_{T_k})\}$($1 \leqq k \leqq K$, K is the number of total trials, $T_k$ is the number of motion steps in each trial).
    The images are pre-trained using Auto Encoder \cite{hinton2006reducing} for dimensionality reduction and feature extraction. 128x96 RGB images were compressed into a 12-dimensional vector and used for training.
    We use these vectors with parametric bias $D_{train} = \{ (D_1, \bmp_1), (D_2, \bmp_2), ...,  (D_K, \bmp_K)\}$ as training data.
    The $\bmp_k$ is a value for each dynamic, in which the operator's movement style is embedded. The model is trained using this data, and $\bmp_k$ is updated along with the network weight $W$.
    The initial value of $\bmp_k$ is set to 0 and the loss function is the mean squared error.
  }

%% \ifthenelse{\boolean{Draft}}{ %%%%%%% Japanese start %%%%%%%
%%   \section{高耐久のハードウェア構成} \label{chap:hardware}
%%   本研究では等身大ヒューマノイドJAXONに高耐久ハンドMSL HANDを装着して実験を行った。
%%   大出力の全身関節に加えて環境との接点であるハンドの耐久性も高めることで長時間のデータ収集が可能なハードウェア構成を実現した。
%%   \subsection{高耐久ハンドMSL HAND}
%%   長期的な作業継続、重量物の操作のために本研究ではヒューマノイドのための高耐久ハンド\cite{mslhand}を使用した。このハンドは、3指5自由度で2本の指の角度を幾何的に固定することが可能である。これにより片手で1000N以上の荷重に耐久し、高負荷作業においても関節負荷上限や温度上昇の心配なく作業を継続することができる。重量箱の持ち上げ実験においてもこのロック機能を使用した。
%% }{ %%%%%% Japanese end: English start %%%%%
%%   \section{} \label{chap:hardware}
%% }

  \ifthenelse{\boolean{Draft}}{ %%%%%%% Japanese start %%%%%%%
    \section{全身作業における操縦模倣実験} \label{chap:update-posture}
    提案した全身トルクを考慮した姿勢生成法を含むヒューマノイドロボットの操縦システムを用いた作業教示と、学習によるスキル獲得実験を示す。
    柔軟物としての風呂敷の操作、ヒューマノイドに特徴的な足の操縦、最後に重心移動を含む重量物の持ち上げ、の3種の作業を行った。
  }{ %%%%%% Japanese end: English start %%%%%
    \section{Experiments on imitation of whole-body tasks} \label{chap:update-posture}
    %% この章では提案した全身トルクを考慮したヒューマノイドロボットの操縦システムによる作業教示実験と、模倣学習によるスキル獲得実験を示す。
    %% 柔軟布の操作、ヒューマノイドに特徴的な足の操縦、重心移動が必要な重量物の持ち上げ、の3種の作業を行った。
    %% 私達はそれぞれの実験についての結果と考察を述べる。
    In this chapter, we describe work teaching experiments using the proposed humanoid robot maneuvering system considering whole-body torque and skill acquisition experiments using imitation learning.
    Three types of tasks were performed: manipulating a flexible cloth, maneuvering legs characteristic of humanoids, and lifting a heavy object requiring center-of-gravity movement.
    We present the results and discussion of each experiment.
  }
\ifthenelse{\boolean{Draft}}{ %%%%%%% Japanese start %%%%%%%
  \subsection{柔軟風呂敷の操作実験}
  TABLISによる操縦で風呂敷を剥がしている様子を\figref{fig:furoshiki-teleop}に示す。
  左が外部動画、右がロボット頭部のカメラからのRGB画像となっている。
  操縦者は風呂敷の様子を直接確認しながら風呂敷へ手を伸ばし,ハンドを操縦して把持し、風呂敷を剥がした後、捨てるという一連の動作を行う。
  この操作の間、RGB画像と、操縦システムを経て出力される\wbc{}に入力される直前の全身の関節角度司令を記録しておく。
  %% この時学習の成功のためには色による区別が重要であったため、ハンドの色を赤色にし、風呂敷の色を黒にして主に机、ハンド、風呂敷、床の色の区別がつきやすいようにした。
  %% ここで使用した操縦システムの最適化姿勢生成のための重みを\tabref{table:weight-furoshiki}に示した。
  %% \begin{figure}[htb]
  %% \centering
  %% \includegraphics[width=0.48\columnwidth]{figs/furoshiki/control/crop0.png}
  %% \includegraphics[width=0.48\columnwidth]{figs/furoshiki/control/crop8.png}\\
  %% \includegraphics[width=0.48\columnwidth]{figs/furoshiki/control/crop16.png}
  %% \includegraphics[width=0.48\columnwidth]{figs/furoshiki/control/crop20.png}
  %% \caption{Experiment of removing cloth on box with proposed teleoperation system}
  %% \label{fig:furoshiki-teleop}
  %% \end{figure}
  \begin{figure}[htb]
  \centering
  \includegraphics[width=0.99\columnwidth]{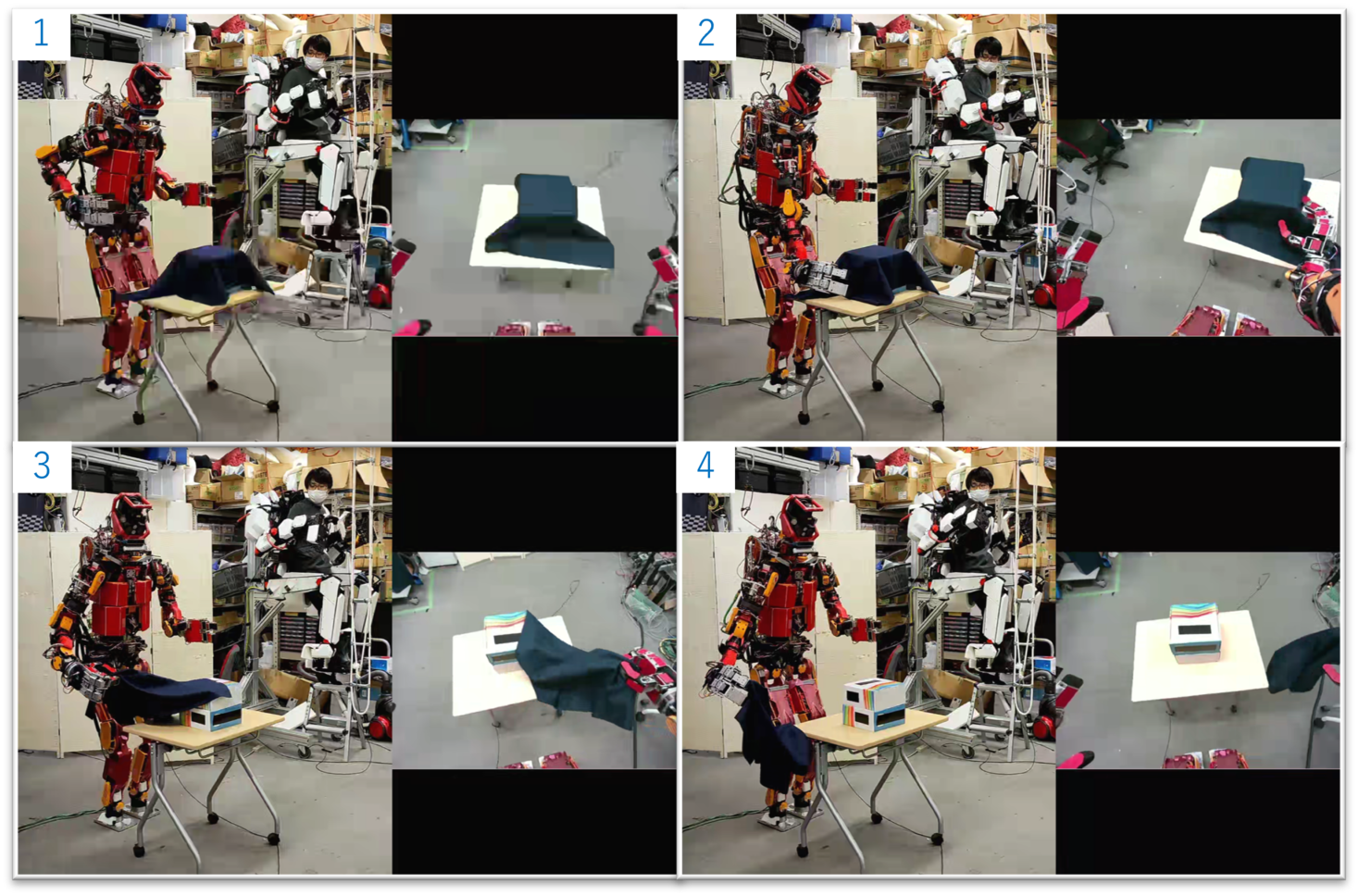}
  \caption{Experiment of removing cloth with proposed teleoperation system}
  \label{fig:furoshiki-teleop}
  \end{figure}

  %% \begin{table}[htb]
  %%   \begin{center}
  %%     \caption{Weights on experiment of removing a flexible cloth}
  %%     \footnotesize
  %%     \begin{tabularx}{0.7\columnwidth}{c|c}
  %%       \hline
  %%       task & weight\\
  %%       \hline
  %%       Kin arm & [1.0,1.0,1.0,1.0,1.0,1.0] \\
  %%       Kin leg & [1.0,1.0,1.0,1.0,1.0,1.0] \\
  %%       Trq & [0.1,0.1,0.1,0.1,0.1,0.1]\\
  %%       Eom & [1000,1000,1000,1000,1000,1000]\\
  %%       COM & [0.05, 0.05, 1.0]\\
  %%       \hline
  %%     \end{tabularx}\label{table:weight-furoshiki}
  %%   \end{center}
  %%   \normalsize
  %% \end{table}

  学習後の作業の様子を\figref{fig:furoshiki-learned-success}に示す。左側が実験時の動画、右上がその時のカメラ画像、右下がAuto Encoderを通して復元された後の画像となっている。
  風呂敷を剥がして捨て、人が落ちた風呂敷を掛け直し、同じ動作を繰り返すという一連の作業を3度連続で成功させている。ここには一度分のデータを示した。
  現在のRGB画像から次のステップの画像を予測して動作を行うため、一度剥がす作業が完了した後に人が再度風呂敷をかぶせた状態を作ってやると、再び剥がす動作が誘起される。ここにはRGB画像と関節角度を用いた模倣学習の特徴の一つがよく現れているといえる。
  模倣が進んでいることを確認するため、把持の様子を表す指の関節角度の変化を\figref{fig:furoshiki-learned-success-finger-angle}に示した。
  データ収集時の操縦時のデータと、学習後の自律的な作業時のデータを比較した。
  両者の親指の第２関節(Thumb1)の角度変化の開始時間を揃えてプロットした。そのため、風呂敷にアプローチ後の把持とリリースの間の角度変化が表されている。
  グラフから把持に利用されるThumb1,Index0,Index1,Middle0の関節角度の遷移のグラフ形状は立ち上がり、低下開始のタイミングが類似しており、角度の最大・最小値も類似したものとなっている。
  %% 今回の指の動きは把持とリリースという簡単な動作であったためこのようにわかりやすい類似が見られたと考えられる。
  適切な把持とリリースにおいては指の関節角度と、その変化のタイミングが重要であり、今回のグラフから操縦データを適切に模倣できていると判断できる。

    \begin{figure}[htb]
    \centering
    \includegraphics[width=0.97\columnwidth]{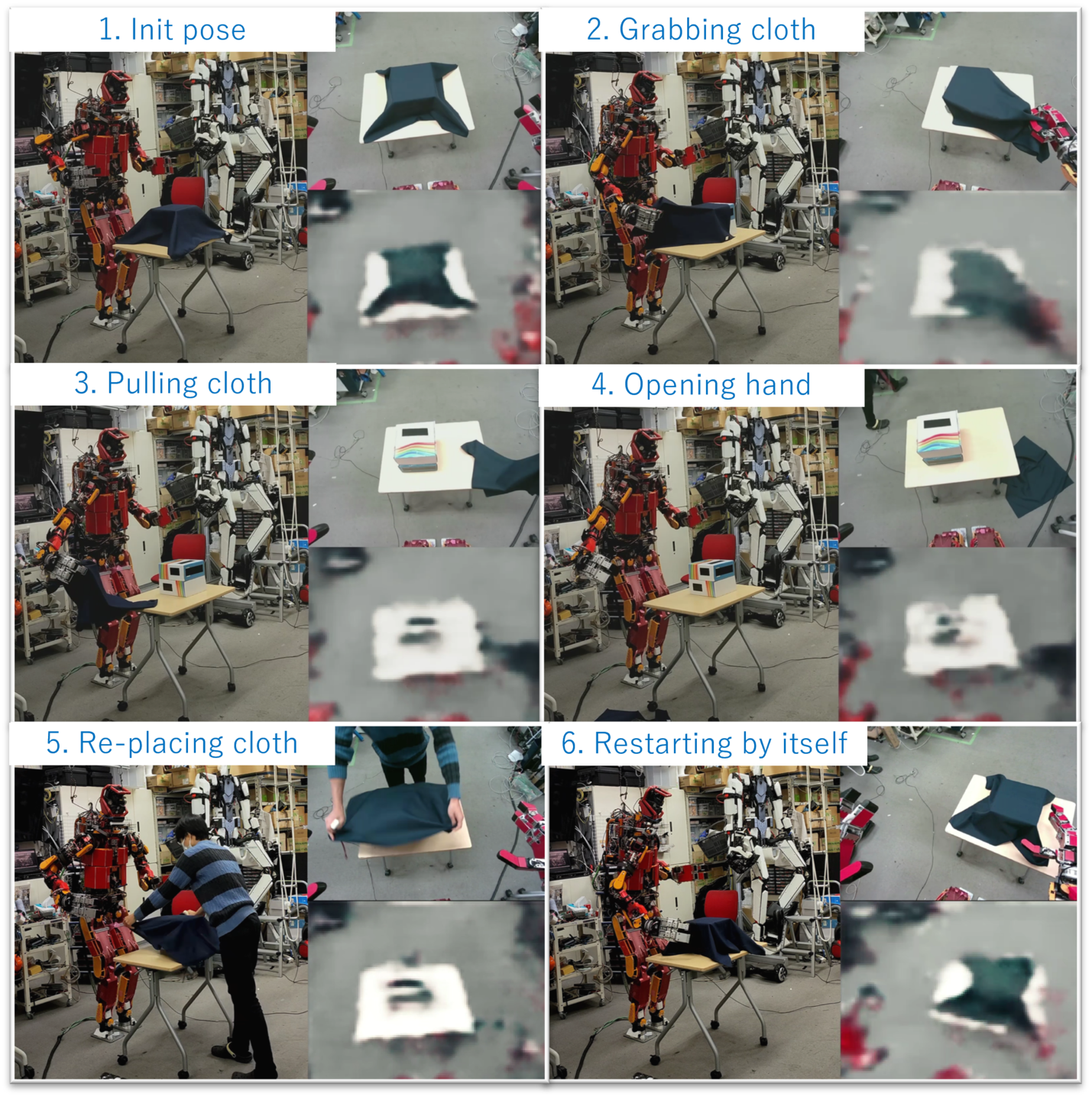}
    \caption{Experiment of succeeded removing cloth with imitation learning}
    \label{fig:furoshiki-learned-success}
  \end{figure}

  \begin{figure}[htb]
    \centering
    \includegraphics[width=0.48\columnwidth]{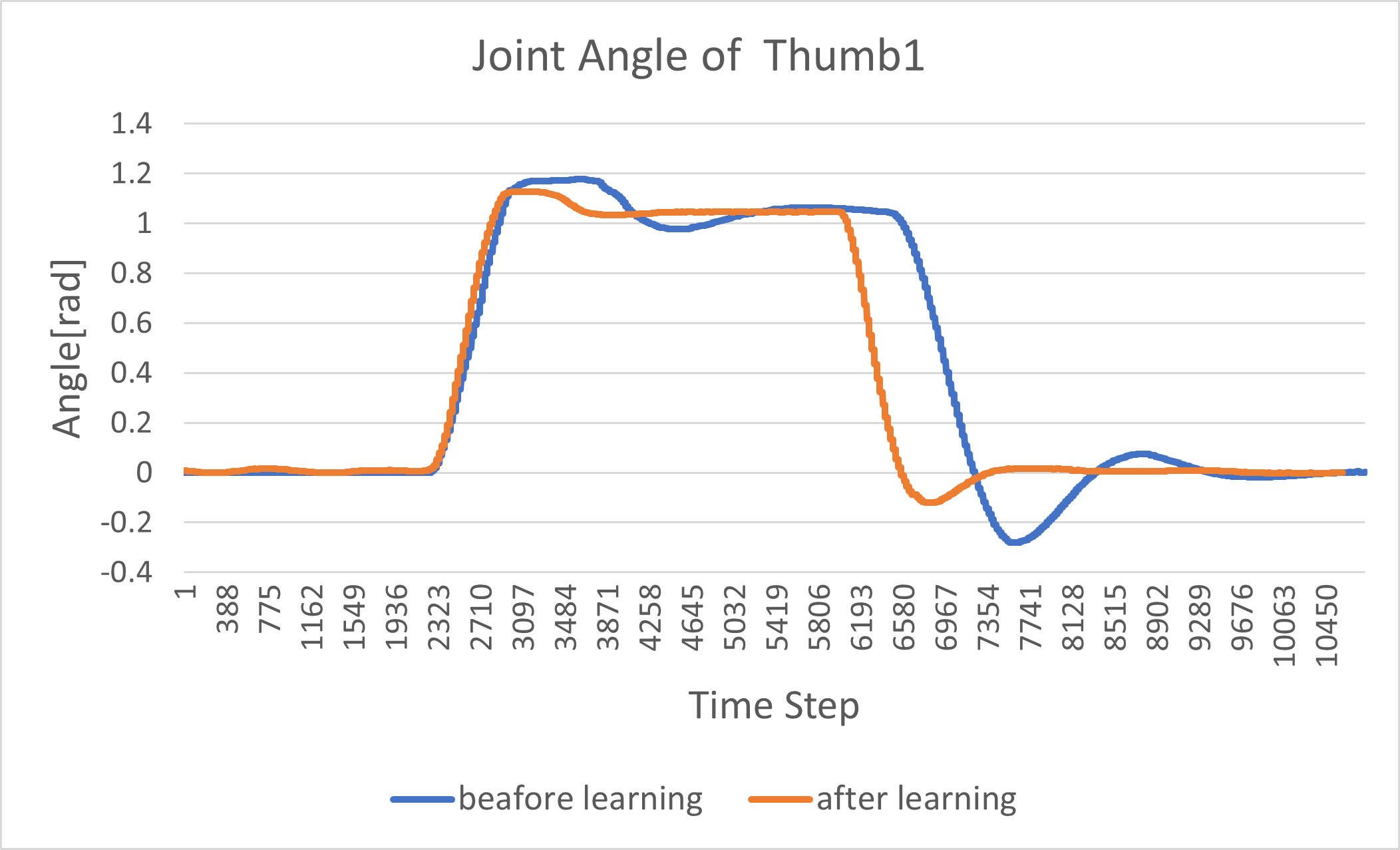}
    \includegraphics[width=0.48\columnwidth]{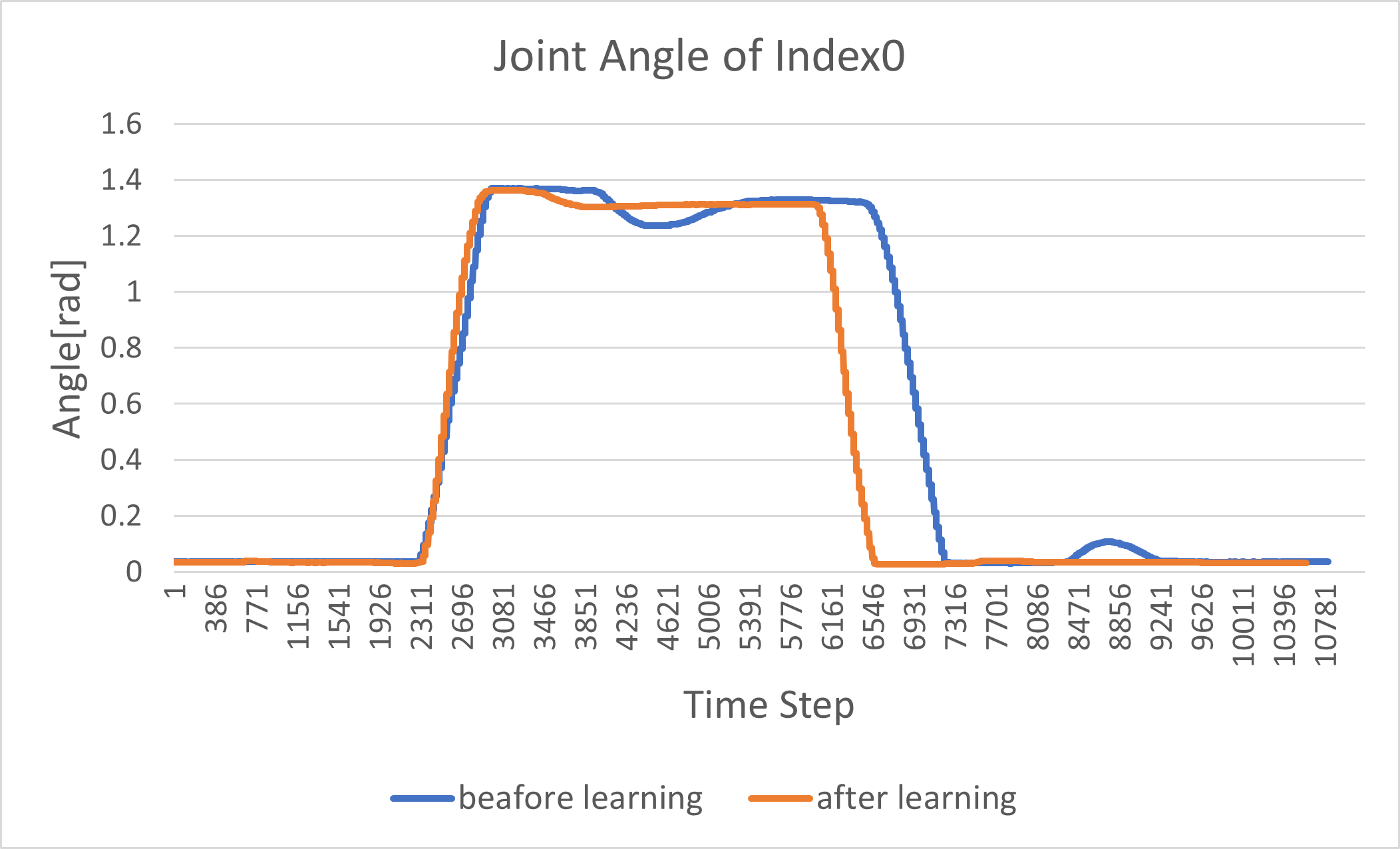}\\
    \includegraphics[width=0.48\columnwidth]{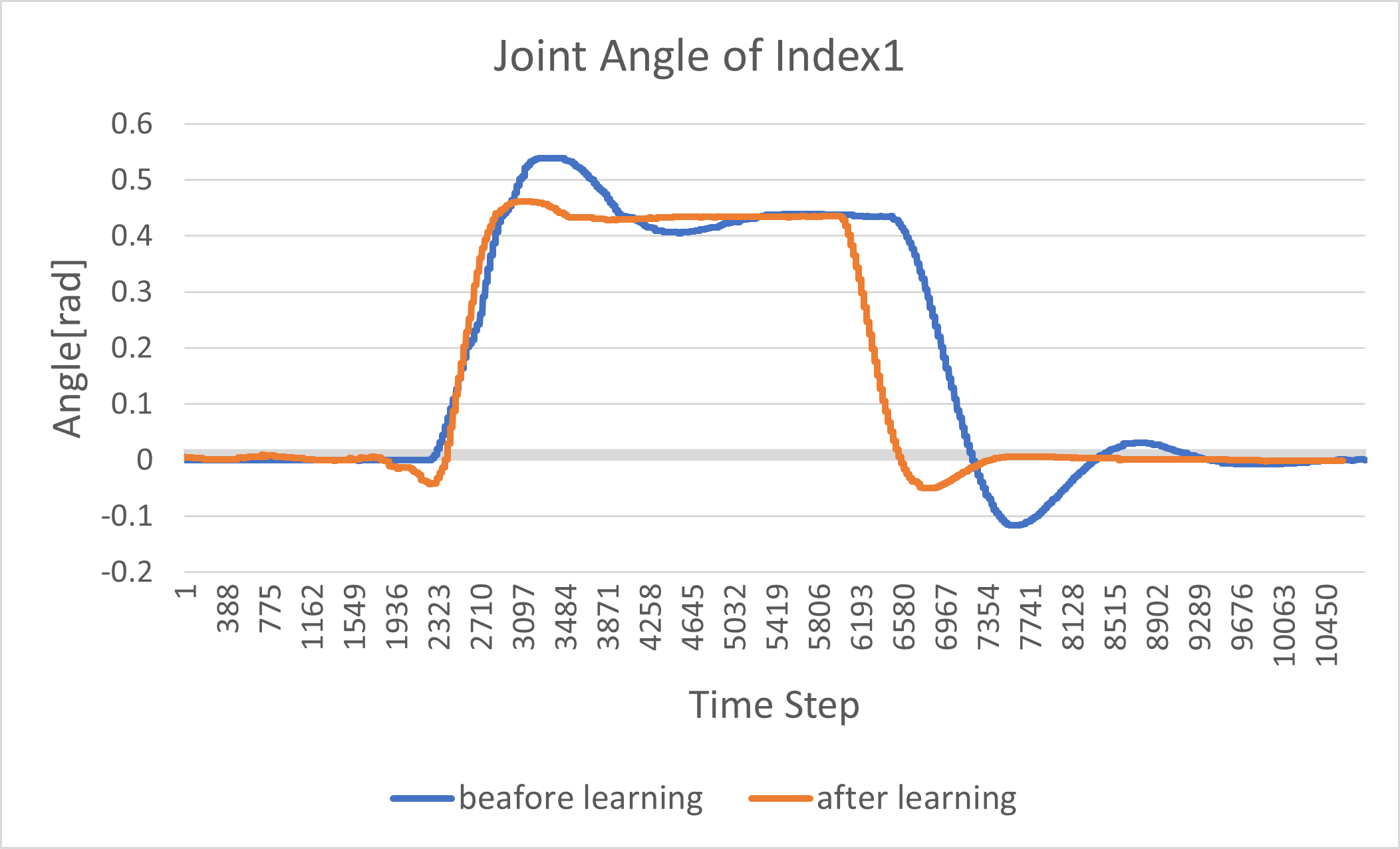}
    \includegraphics[width=0.48\columnwidth]{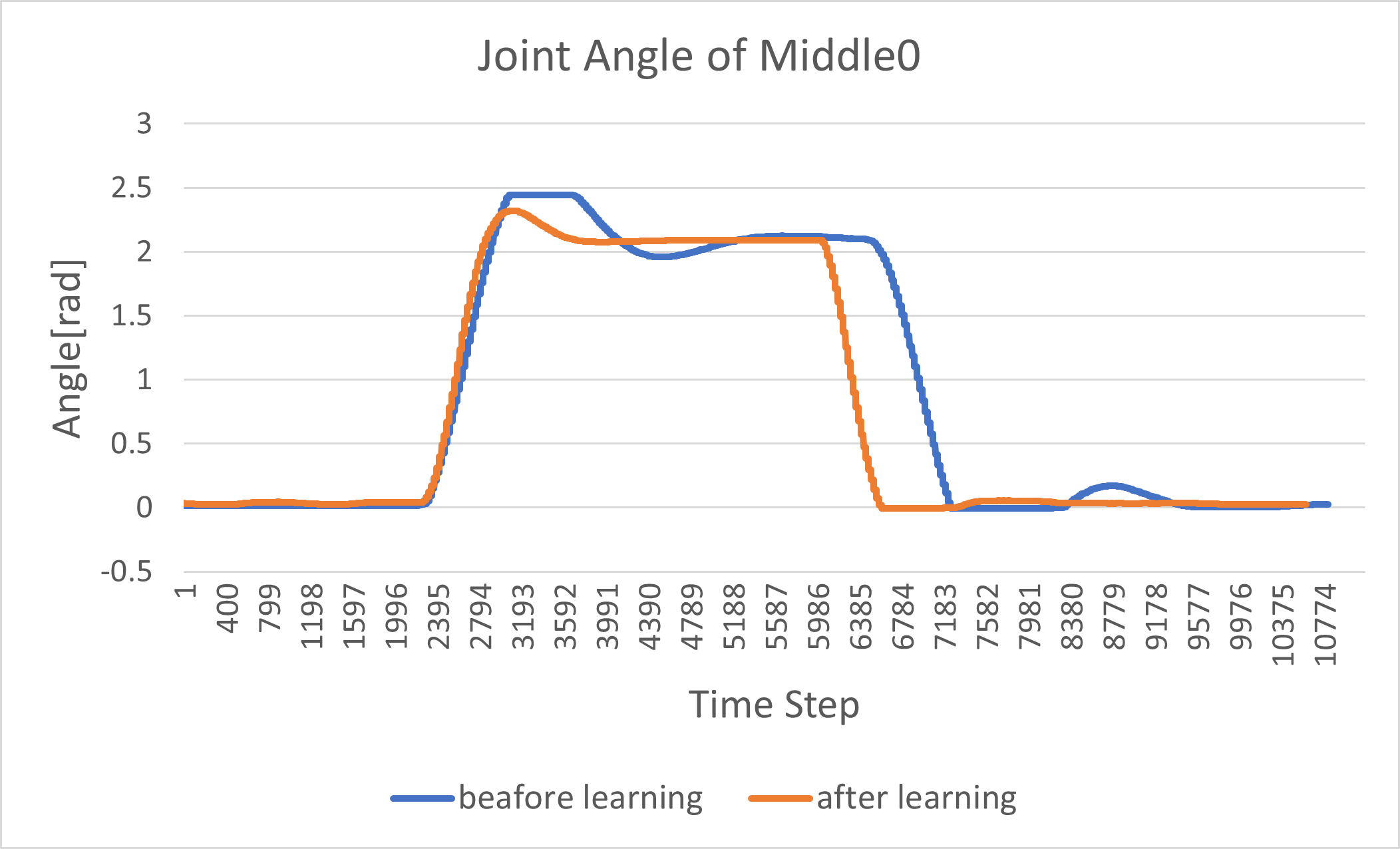}\\
    \caption{Comparison of finger angle during teleoperation and autonomous operation}
    \label{fig:furoshiki-learned-success-finger-angle}
  \end{figure}

  \subsection{足を使った物体操作実験}
  ヒューマノイドに特徴的な足を使った物体操作の実験を行った。
  JAXONの足を操縦しゴミ箱を開けている様子を\figref{fig:gomibako-teleop}に示す。左が外部動画と右がRGB画像である。
  手順としては、操縦者の足先の位置を取得し提案システムを通して全身の関節角度列を計算する。この計算された姿勢の両足の高さの差分が閾値以上(6 cm)となった際に、操縦司令を止めて\fsc{}からWhole Body Controllerに対して2足歩行モードから片足立ちモードへ遷移する司令を送る。
  その際には、重心軌道を計算し遷移するためバランスを考慮した片足立ちへの移行が可能となっている。遷移後は操縦者による足の操縦を再開する。
  片足立ちモードではロボットの足先力センサの値を低ゲインで操縦者へフィードバックしているため、ゴミ箱のペダルとの接触、操作時の反力を操縦者が認知することができる。
  ゴミ箱を開けた後は同様の手順を逆再生し、地面に足を戻す。
  この際も両足の高さの差分に対し同じ閾値を用いてモードを切り替えるが、ペダルを踏む時にモード切替が起こらないように、ロボットのx軸方向に新たな閾値を設け、両足のx方向の位置が十分近い状態でのみモードを切り替えるようにした。
  %% ここで使用した操縦システムの最適化姿勢生成のための重みを\tabref{table:weight-gomibako}に示した。
  学習後の作業の様子を\figref{fig:gomibako-success}に示した。
  左側が実験時の動画、右上がその時のカメラ画像、右下がAuto Encoderを通して復元された後の画像となっている。
  足の接触状態の切り替えを含む足の操縦に成功している。
  %% 左右の足のレンチを\figref{fig:gomibako-wrench}に示す。学習には右足のレンチも入力している。
  %% z軸力に注目すると両足に600N程度の荷重がかかった状態から、片足立ちの瞬間から左足の荷重が1200N程度まで増加している。
  %% ペダル操作時も反力を感じており特にy軸回りのモーメントの値が-60Nmから20Nmの間で変化している。最後には2足保持状態に戻っている様子がわかる。
  %% \begin{figure}[htb]
  %%   \centering
  %%   \includegraphics[width=0.48\columnwidth]{figs/gomibako/control/crop0.png}
  %%   %% \includegraphics[width=0.48\columnwidth]{figs/gomibako/control/crop1.png}\\
  %%   %% \includegraphics[width=0.48\columnwidth]{figs/gomibako/control/crop2.png}
  %%   %% \includegraphics[width=0.48\columnwidth]{figs/gomibako/control/crop3.png}\\
  %%   %% \includegraphics[width=0.48\columnwidth]{figs/gomibako/control/crop4.png}
  %%   \includegraphics[width=0.48\columnwidth]{figs/gomibako/control/crop5.png}\\
  %%   %%   \caption{Experiment of operating a right foot with proposed teleoperation system: first 6 step}\label{fig:gomibako-teleop-first}
  %%   %% \end{figure}
  %%   %% \begin{figure}[htb]
  %%   %%   \centering
  %%   %% \includegraphics[width=0.48\columnwidth]{figs/gomibako/control/crop6.png}
  %%   %% \includegraphics[width=0.48\columnwidth]{figs/gomibako/control/crop7.png}\\
  %nnn%   \includegraphics[width=0.48\columnwidth]{figs/gomibako/control/crop8.png}
  %%   %% \includegraphics[width=0.48\columnwidth]{figs/gomibako/control/crop9.png}\\
  %%  nnn %% \includegraphics[width=0.48\columnwidth]{figs/gomibako/control/crop10.png}
  %%   \includegraphics[width=0.48\columnwidth]{figs/gomibako/control/crop11.png}\\
  %%   \caption{Experiment of operating a right foot with proposed teleoperation system}\label{fig:gomibako-teleop}
  %% \endn{figure}
  \begin{figure}[htb]
    \centering
    \includegraphics[width=0.99\columnwidth]{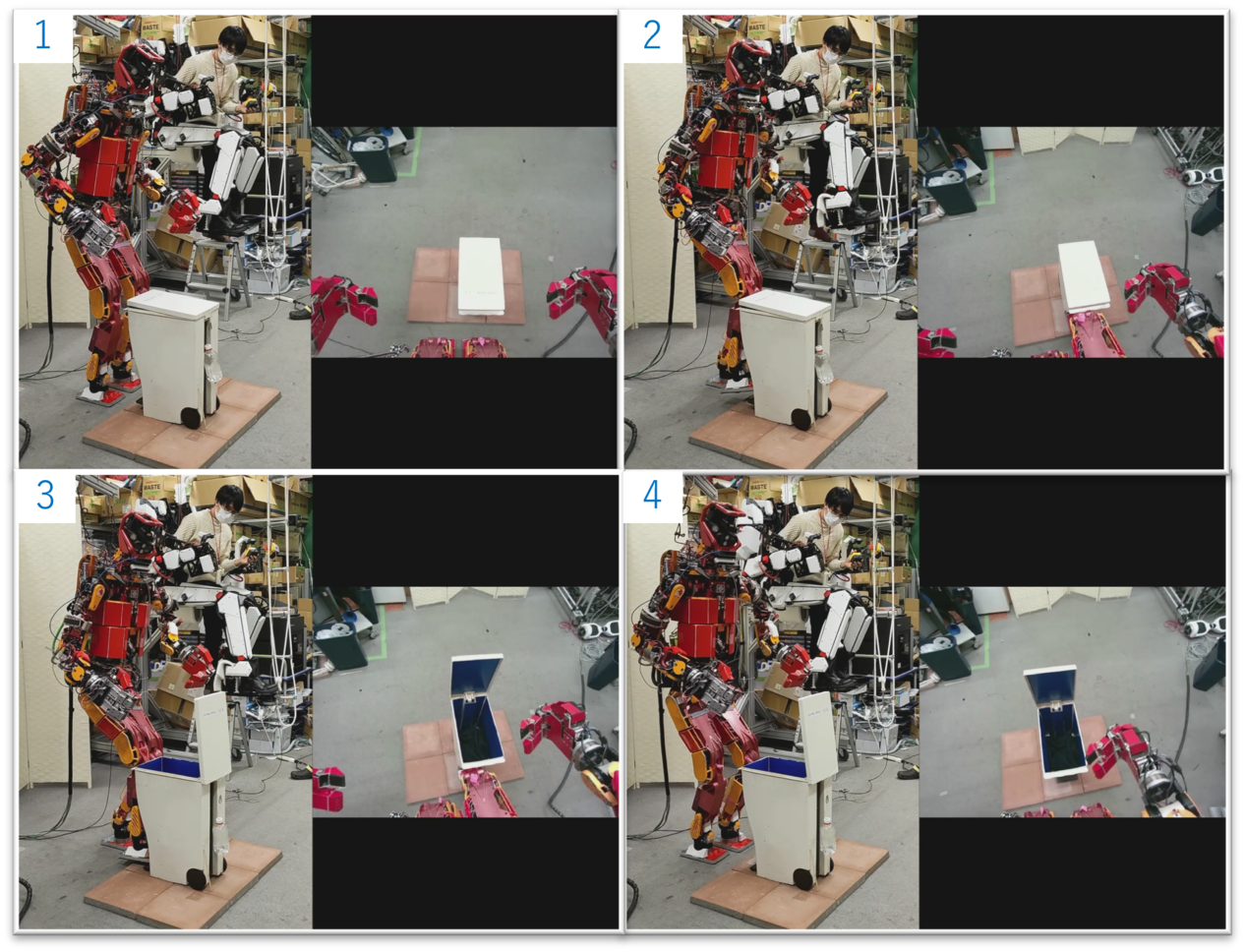}
    \caption{Experiment of operating a foot with proposed teleoperation system}\label{fig:gomibako-teleop}
  \end{figure}

  %% \begin{table}[htb]
  %%   \begin{center}
  %%     \nnnnnnnnnnnnnnnnnnnnnnnnnnnnnnnnnnnnnnnnnnnnnnnnnnnnnnnnncaption{Weights on experiment of remote control of a foot}
  %%     \footnotesize
  %%     \begin{tabularx}{0.7\columnwidth}{c|c}
  %%       \hline
  %%       task & weight\\
  %%       \hline
  %%       Kin arm & [1.0,1.0,1.0,1.0,1.0,1.0] \\
  %%       Kin leg & [1.0,1.0,1.0,1.0,1.0,1.0] \\
  %%       Kin com & [0.0,0.0,0.0,1.0,0.0,1.0] \\
  %%       Kin head & [0.0,0.0,0.0,0.0,0.0,0.5] \\
  %%       Trq & [0.1,0.1,0.1,0.1,0.1,0.1]\\
  %%       Eom & [1000,1000,1000,1000,1000,1000]\\
  %%       COM & [0.05, 0.05, 1.0]\\
  %%       COM height & 0.5\\
  %%       \hline
  %%     \end{tabularx}\label{table:weight-gomibako}
  %%   \end{center}
  %%   \normalsize
  %% \end{table}

  %% \begin{figure}[htb]
  %%   \centering
  %%   \includegraphics[width=0.475\columnwidth]{figs/gomibako/success/resize/crop0.png}
  %%   %% \includegraphics[width=0.475\columnwidth]{figs/gomibako/success/resize/crop1.png}\\
  %%   \includegraphics[width=0.475\columnwidth]{figs/gomibako/success/resize/crop2.png}\\
  %%   %% \includegraphics[width=0.475\columnwidth]{figs/gomibako/success/resize/crop3.png}\\
  %%   \includegraphics[width=0.475\columnwidth]{figs/gomibako/success/resize/crop4.png}
  %%   %% \includegraphics[width=0.475\columnwidth]{figs/gomibako/success/resize/crop5.png}\\
  %%   \includegraphics[width=0.475\columnwidth]{figs/gomibako/success/resize/crop6.png}\\
  %%   \caption{Experiment of succeeded operating a right foot with imitation learning}\label{fig:gomibako-success}
  %% \end{figure}
  \begin{figure}[htb]
    \centering
    \includegraphics[width=0.99\columnwidth]{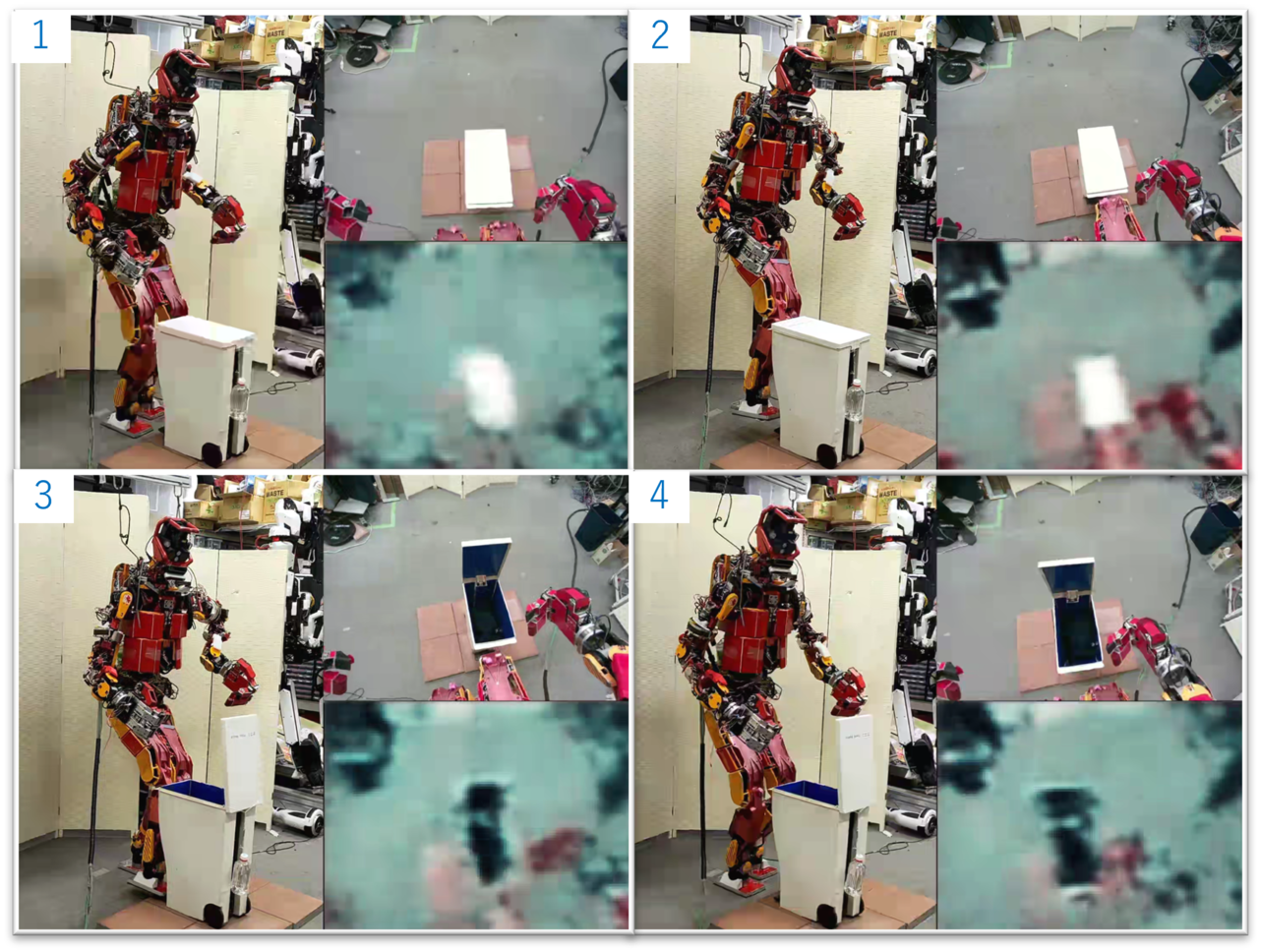}
    \caption{Experiment of succeeded operating a foot with imitation learning}\label{fig:gomibako-success}
  \end{figure}
  %% \begin{figure}[htb]
  %% \centering
  %% \includegraphics[width=0.48\columnwidth]{figs/gomibako/success/plot/gomibako_suc_force_x.png}
  %% \includegraphics[width=0.48\columnwidth]{figs/gomibako/success/plot/gomibako_suc_moment_x.png}\\
  %% \includegraphics[width=0.48\columnwidth]{figs/gomibako/success/plot/gomibako_suc_force_y.png}
  %% \includegraphics[width=0.48\columnwidth]{figs/gomibako/success/plot/gomibako_suc_moment_y.png}\\
  %% \includegraphics[width=0.48\columnwidth]{figs/gomibako/success/plot/gomibako_suc_force_z.png}
  %% \includegraphics[width=0.48\columnwidth]{figs/gomibako/success/plot/gomibako_suc_moment_z.png}\\
  %% \caption{Plot of foot wrench sensor value}
  %% \label{fig:gomibako-wrench}
  %% \end{figure}

  \subsection{屈み込みを伴う重量物操作}
  ここでは、重量物として合計約16 kgの箱の持ち上げに取り組んだ。
  膝を使った屈み込みと手先の反力を考慮したバランス維持が重要となる。
  操縦による教示の様子を\figref{fig:mochiage-teleop}に示す。
  人さし指と中指を90度でロックした初期姿勢から両手を箱の持ち手に伸ばし、反力を感じたら鉛直に持ち上げ始める動作を操縦で行った。
  %% ここで使用した操縦システムの最適化姿勢生成のための重みを\tabref{table:weight-mochiage}に示した。
  操縦システムの重心高さに対する制約のおかげで手先の高さが下がると膝を曲げてルートリンクをかたむけて重心を下げる姿勢が生成されている。
  今回は手先の位置のみを操縦しており、操縦デバイスの手先とロボットの座標系の関係を調整し、操縦者の手の動きとロボットの手先の動きを対応付けることで直感的な操作ができるようにした。
  ヒューマノイドロボットには2脚で立ち続けるという強い制約が存在するため、特に重量物の操作では、手先の外力に合わせて重心位置を変更するといったバランス制御が不可欠である。
  今回の操縦においては、操縦システム内の最適化姿勢生成において重心を足平中心に維持する制約を加え、更に手先のセンサ値に合わせて重心位置を移動させるフィードバック制御を用いてバランスを維持している。
  力センサ値に対するローパスフィルタに起因する時間遅れに対応するため、手先にかかる外力が大きく変化する持ち上げ始めでは、十分にゆっくりと手先を動かすようにする必要があった。
  学習後の作業の様子を\figref{fig:mochiage-teleop}に示す。
  持ち手に対して両手を差し込み、重心を適切に移動させることでバランスを維持しながら持ち上げに成功している。
  実験の中では一度目の差し込み動作に失敗して箱が初期位置からず足してしまった場合でも、ロボットが自律的に再度、適切な位置に対して差し込み直しをする動作が見られた。リカバリーの動きが自律的に発現するという模倣学習による作業のロバスト性も確認できた。
  操縦時と学習後の右手のセンサ値をプロットしたものを\figref{fig:mochiage-compare-wrench}に示す。両手のレンチも学習の入力として使用した。
  作業開始時を揃えてプロットしているため、環境の初期状態によって持ち上げはじめのタイミングはずれるのが自然であるが、特にz軸方向の力や、z軸回りのモーメントの変化の概形は青色の学習前とオレンジ色の学習後で類似していると言える。
  %% この持ち上げから箱を下ろすまでの間の手先の力センサ値の変化を\figref{fig:mochiage-control-wrench}に示す。

  %% ここでは右手と左手の値を比較しているが、両手にバランス良く荷重がかかっていることがわかる。
  %% y軸方向のちからがかかった後にz軸方向の力が変化し始めていることから、たしかに操縦者が箱を掴んだ際に反力を感じてから持ち上げを始めていることがわかる。
  %% 先にy軸方向の力を感じてから持ち上げを開始することで、学習モデルが箱の持ち上げを開始するタイミングを決めやすくなり、成功率が上がることが狙いである。

  %% \begin{figure}[htb]
  %%   \centering
  %%   \includegraphics[width=0.48\columnwidth]{figs/mochiage/control2/resize/crop0.png}
  %%   %% \includegraphics[width=0.48\columnwidth]{figs/mochiage/control2/resize/crop1.png}\\
  %%   %% \includegraphics[width=0.48\columnwidth]{figs/mochiage/control2/resize/crop2.png}
  %%   %% \includegraphics[width=0.48\columnwidth]{figs/mochiage/control2/resize/crop3.png}\\
  %%   \includegraphics[width=0.48\columnwidth]{figs/mochiage/control2/resize/crop4.png}\\
  %%   \includegraphics[width=0.48\columnwidth]{figs/mochiage/control2/resize/crop5.png}
  %%   %% \includegraphics[width=0.48\columnwidth]{figs/mochiage/control2/resize/crop6.png}
  %%   \includegraphics[width=0.48\columnwidth]{figs/mochiage/control2/resize/crop7.png}\\
  %%   \caption{Experiment of lifting heavy box up with proposed teleoperation system}
  %%   \label{fig:mochiage-teleop}
  %% \end{figure}
  \begin{figure}[htb]
    \centering
    \includegraphics[width=0.99\columnwidth]{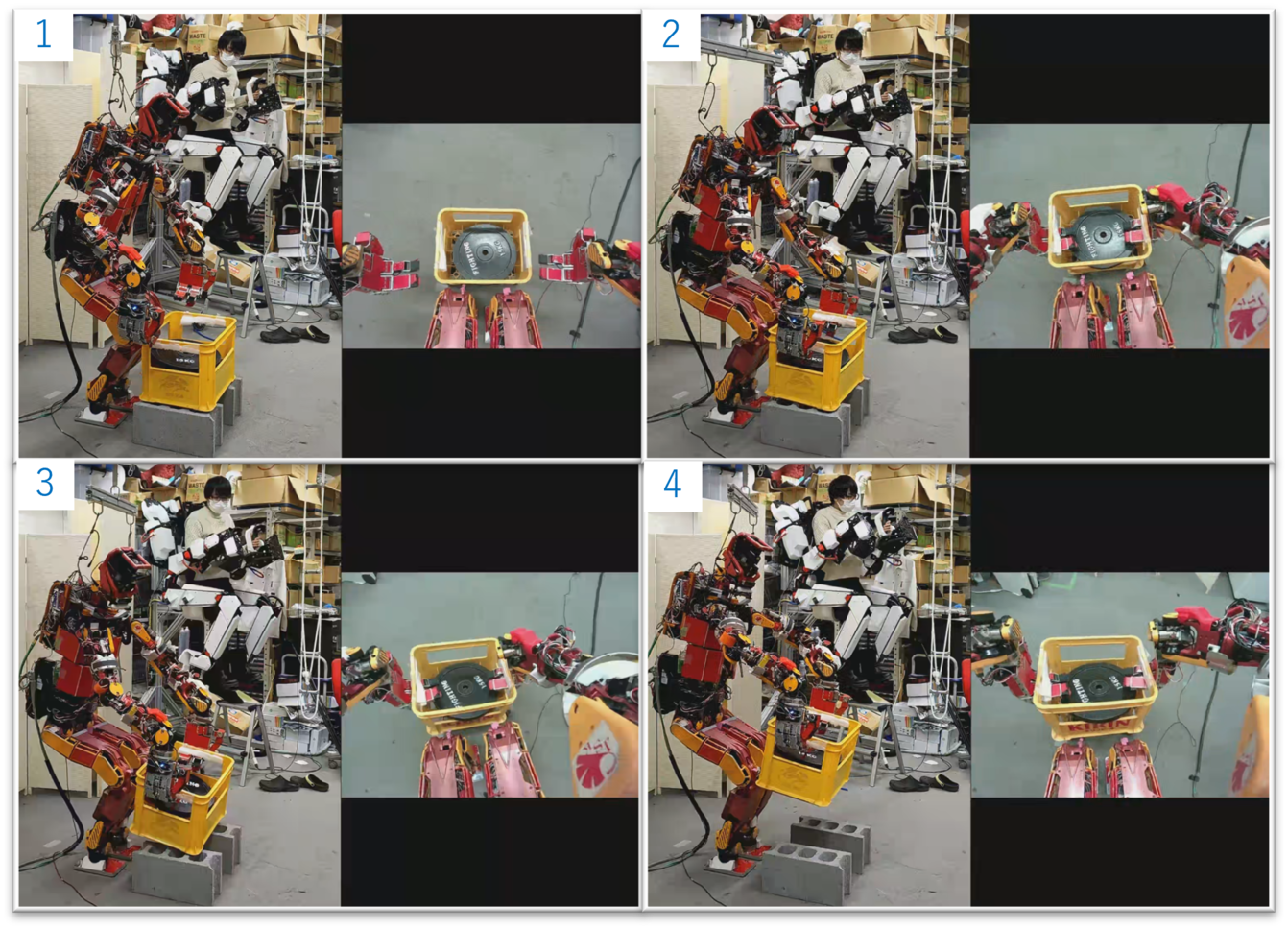}
    \caption{Experiment of lifting a heavy box with proposed teleoperation system}
    \label{fig:mochiage-teleop}
  \end{figure}
  %% \begin{figure}[htb]
  %%   \centering
  %%   \includegraphics[width=0.475\columnwidth]{figs/mochiage/success2/resize/crop0.png}
  %%   %% \includegraphics[width=0.475\columnwidth]{figs/mochiage/success2/resize/crop1.png}\\
  %%   %% \includegraphics[width=0.475\columnwidth]{figs/mochiage/success2/resize/crop2.png}\\
  %%   %% \includegraphics[width=0.475\columnwidth]{figs/mochiage/success2/resize/crop3.png}
  %%   \includegraphics[width=0.475\columnwidth]{figs/mochiage/success2/resize/crop4.png}\\
  %%   \includegraphics[width=0.475\columnwidth]{figs/mochiage/success2/resize/crop5.png}
  %%   %% \includegraphics[width=0.475\columnwidth]{figs/mochiage/success2/resize/crop6.png}
  %%   \includegraphics[width=0.475\columnwidth]{figs/mochiage/success2/resize/crop7.png}\\
  %%   \caption{Experiment of succeeded lifting heavy box up with imitation learning}
  %%   \label{fig:mochiage-success}
  %% \end{figure}

%%   \begin{table}[hbt]
%%   \begin{center}
%%     \caption{Weight on experiment of lifting up a heavy box}
%%     \footnotesize
%%     \begin{tabularx}{0.7\columnwidth}{c|c}
%%       \hline
%%       task & weight\\
%%       \hline
%%       Kin arm & [1.0,1.0,1.0,1.0,1.0,1.0] \\
%%       Kin leg & [1.0,1.0,1.0,1.0,1.0,1.0] \\
%%       Kin com & [0.0,0.0,0.0,1.0,0.0,1.0] \\
%%       Kin head & [0.0,0.0,0.0,0.0,0.0,0.5] \\
%%       Trq & [0.1,0.1,0.1,0.1,0.1,0.1]\\
%%       Eom & [1000,1000,1000,1000,1000,1000]\\
%%       COM & [1.0, 1.0, 1.0]\\
%%       COM height & 0.5\\
%%       \hline
%%     \end{tabularx}  \label{table:weight-mochiage}
%%   \end{center}
%%   \normalsize
%% \end{table}

  \begin{figure}[htb]
    \centering
    \includegraphics[width=0.99\columnwidth]{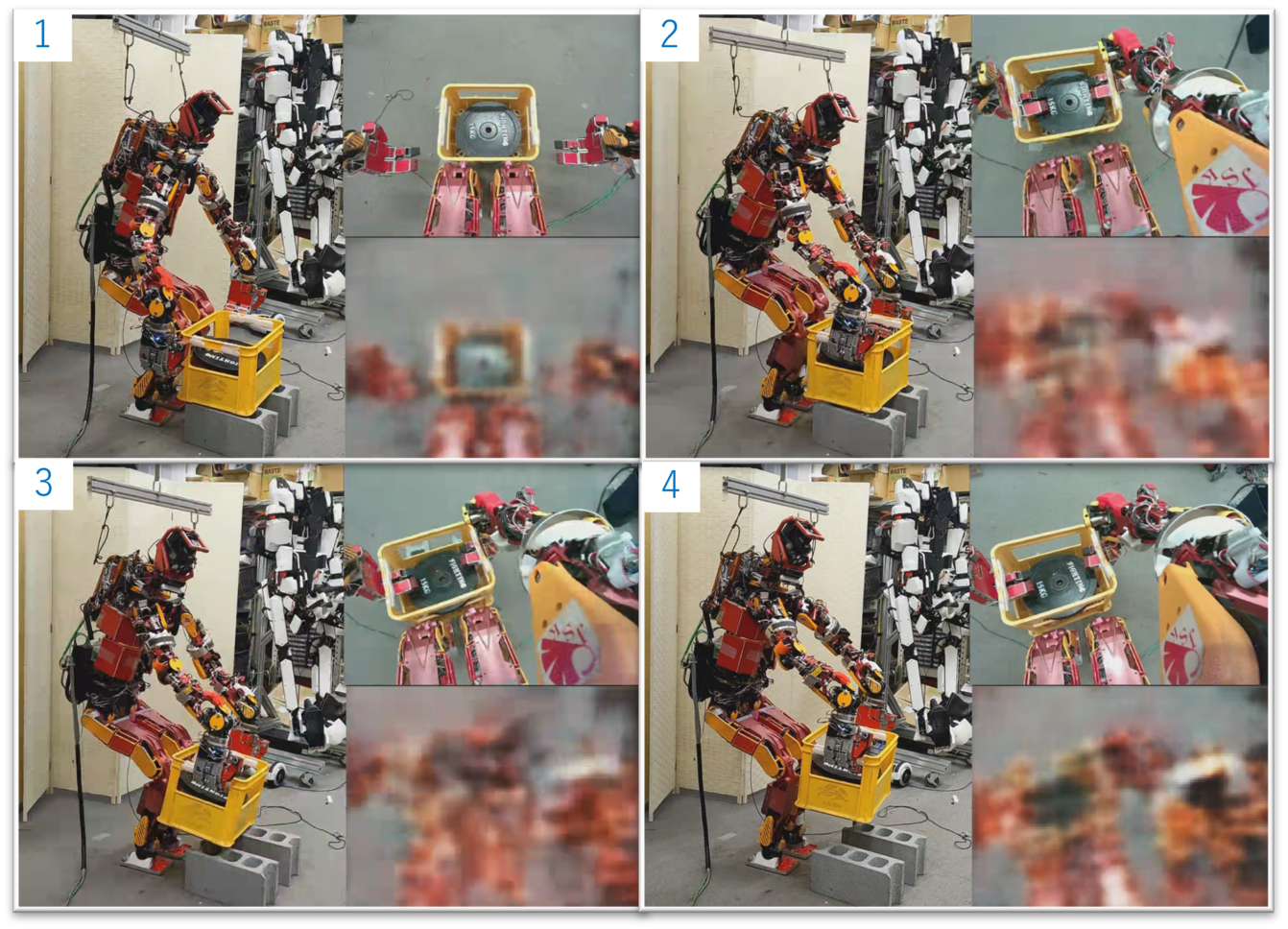}
    \caption{Experiment of succeeded lifting heavy box with imitation learning}
    \label{fig:mochiage-success}
  \end{figure}

  \begin{figure}[htb]
    \centering
    \includegraphics[width=0.48\columnwidth]{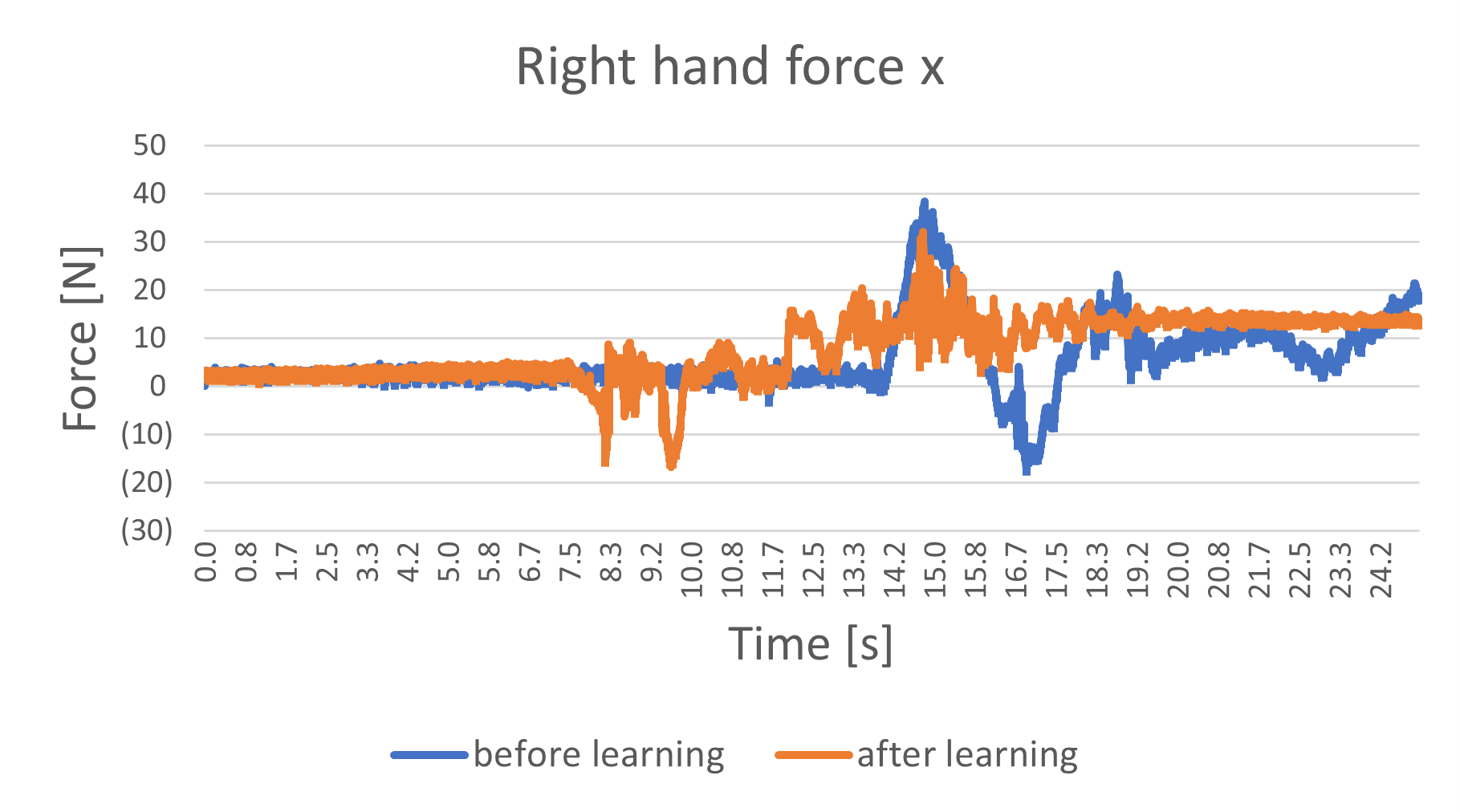}
    \includegraphics[width=0.48\columnwidth]{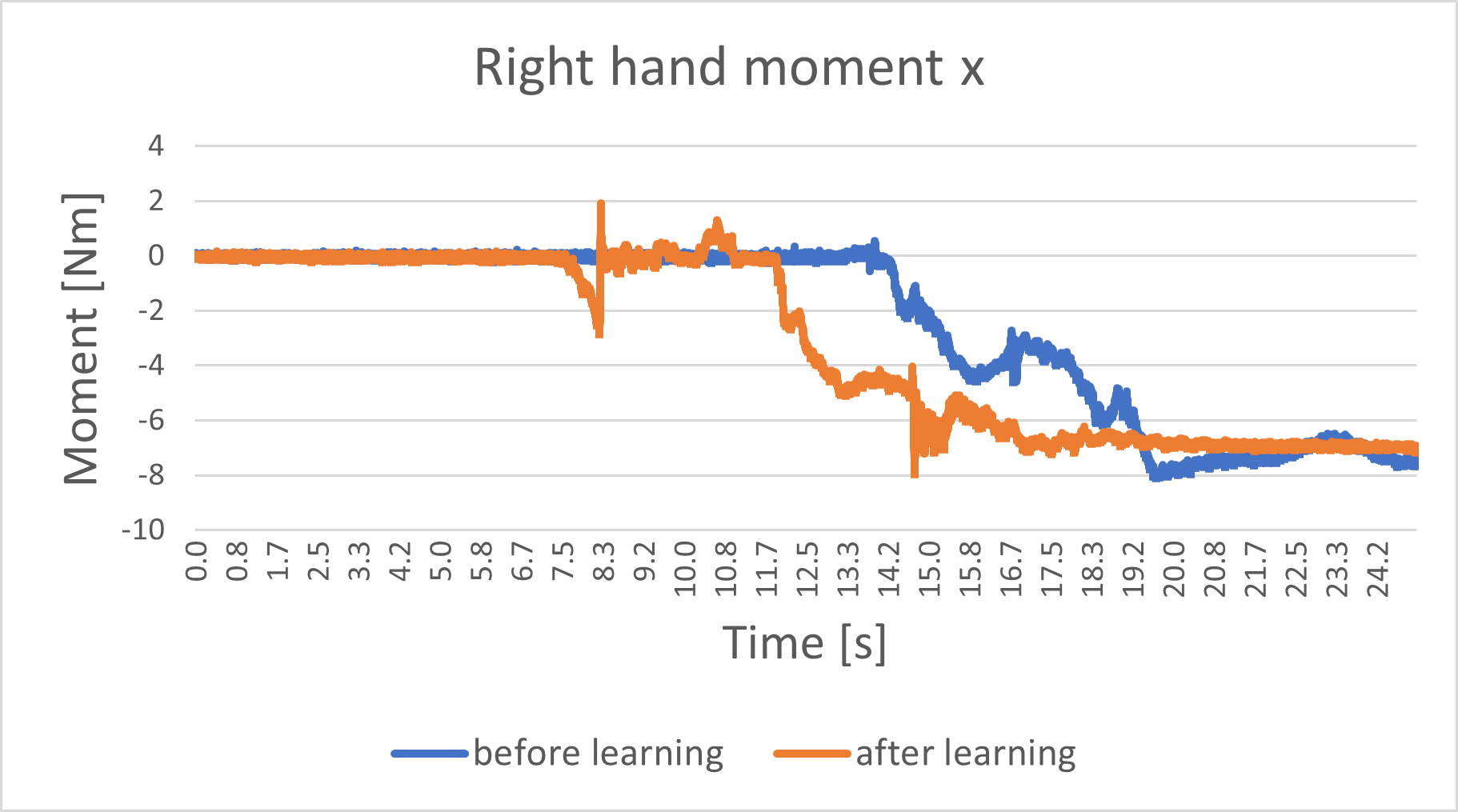}\\
    \includegraphics[width=0.48\columnwidth]{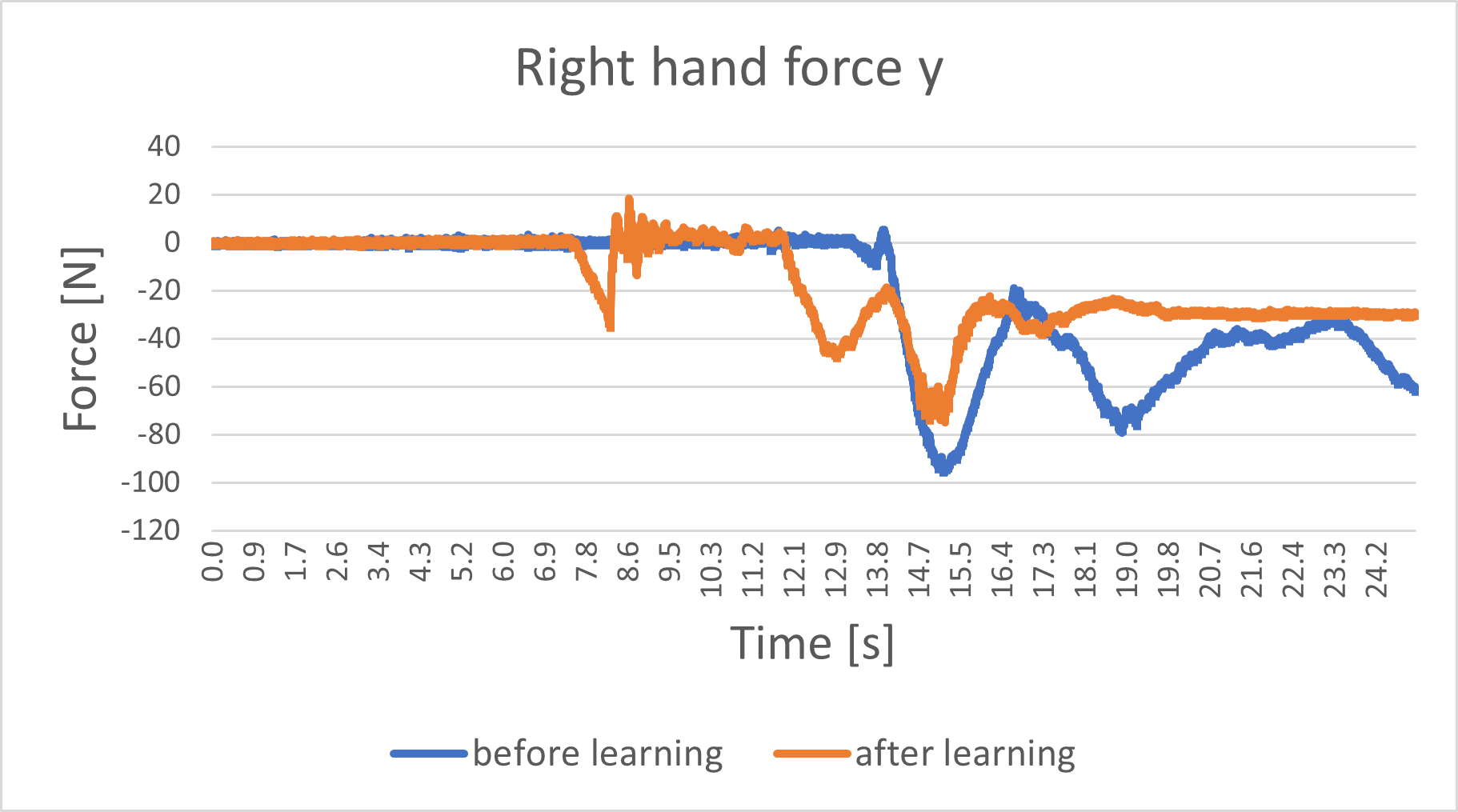}
    \includegraphics[width=0.48\columnwidth]{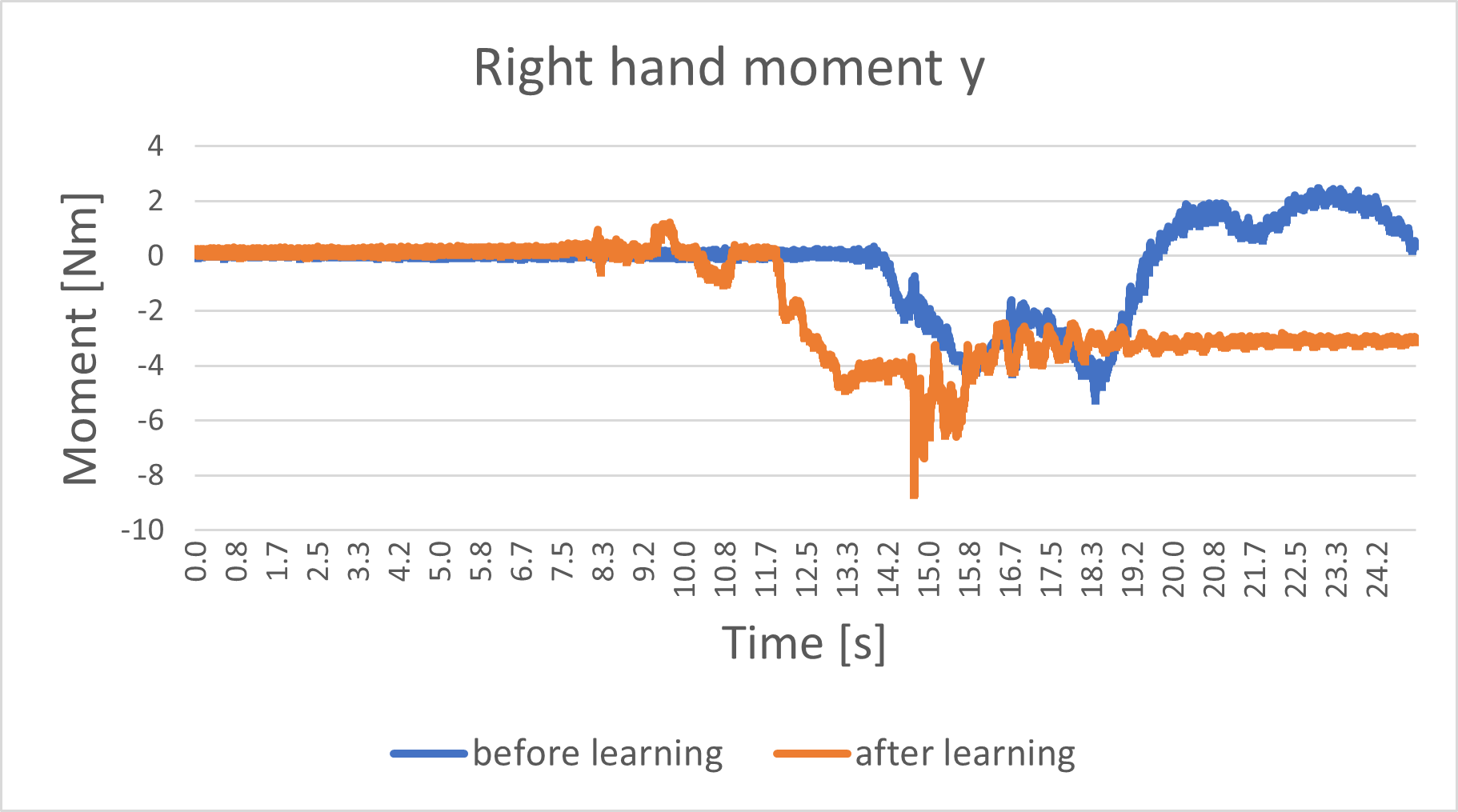}\\
    \includegraphics[width=0.48\columnwidth]{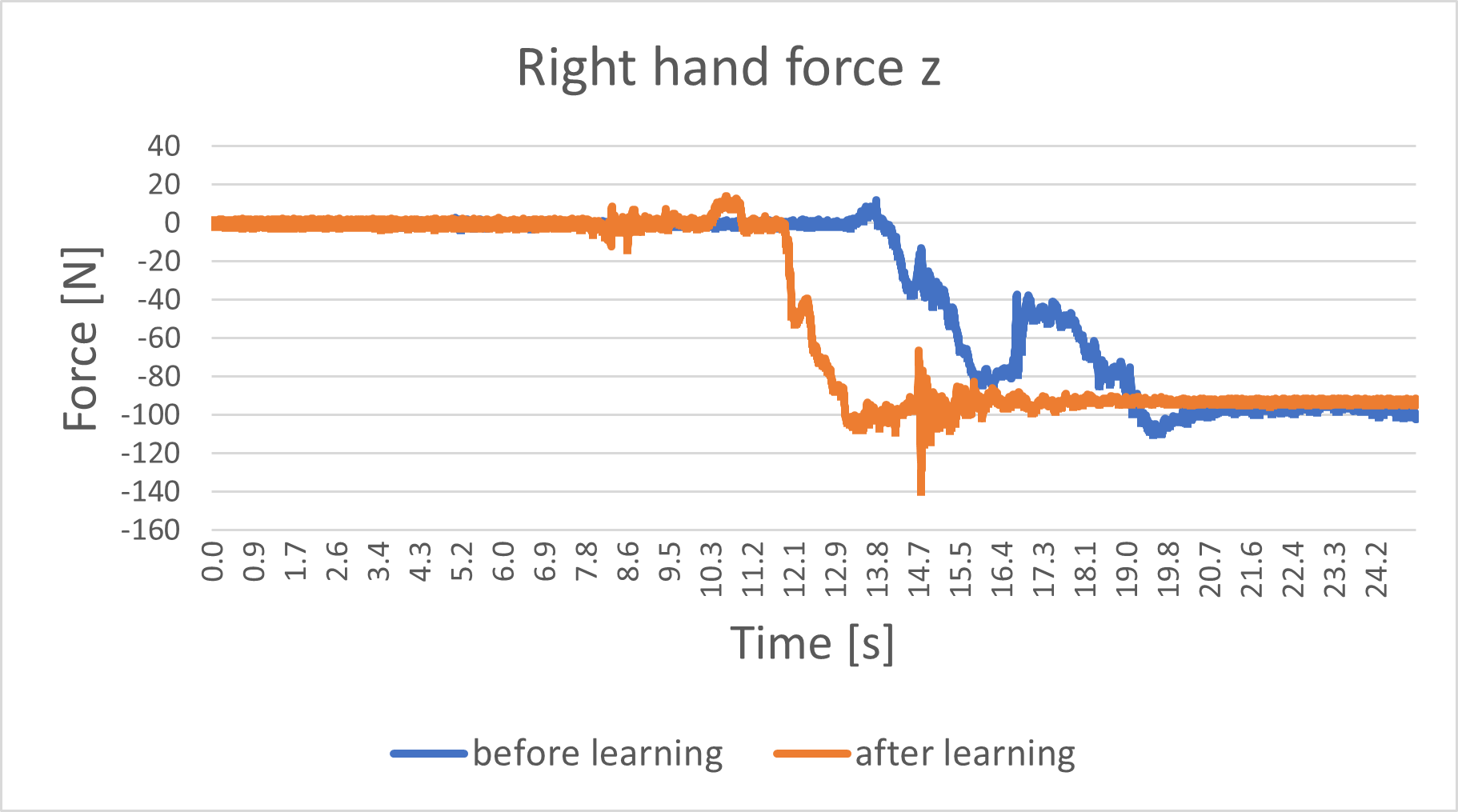}
    \includegraphics[width=0.48\columnwidth]{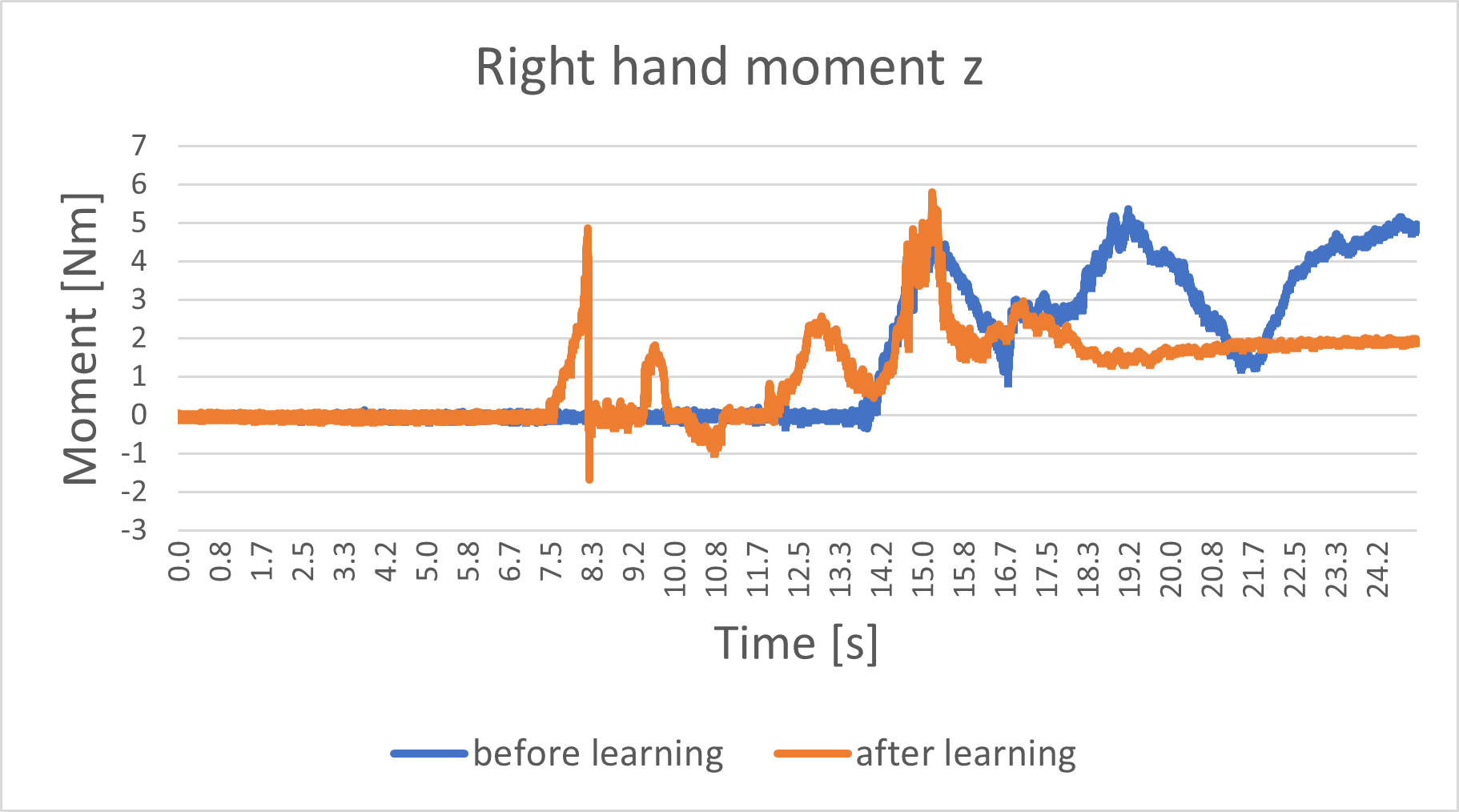}\\
    \caption{Plot of hand wrench sensor value}
    \label{fig:mochiage-compare-wrench}
  \end{figure}

}{ %%%%%% Japanese end: English start %%%%%
  \subsection{Experiments of flexible cloth manipulation}

  %% TABLISによる操縦でJAXONが柔軟布を剥がしている様子を\figref{fig:furoshiki-teleop}に示す。
  %% 左が外部カメラの画像、右がロボット頭部のカメラからのRGB画像となっている。
  %% 操縦者は柔軟布の様子を直接確認しながらそれへ手を伸ばし,ハンドを操縦して把持し、それを剥がした後、脇に捨てるという一連の動作を行う。
  %% この操作の間、RGB画像と操縦システムから出力される全身の関節角度司令を記録しておく。
  JAXON removing the flexible cloth by maneuvering with TABLIS is shown in \figref{fig:furoshiki-teleop}.
  The left image is from an external camera and the right image is from a camera on the robot's head.
  The operator sees the flexible cloth directly, reaches out to grasp it, then removes it and drops it aside.
  During this operation, the system recorded the RGB image and the whole-body joint angle ($\bthetahfpg$) from the operating system.

  %% 学習後の作業の様子を\figref{fig:furoshiki-learned-success}に示す。左側が外部カメラからの画像、右上がロボットの頭部のカメラ画像、右下がAuto Encoderを通して復元された画像となっている。
  %% 布を剥がして捨て、落ちた布を人が置き直し、同じ動作を繰り返すという一連の作業を3度連続で成功させている。ここには一回分のデータを示した。
  %% 現在のRGB画像から次のステップの画像を予測して動作を行うため、一度作業が完了した後に人が再度はじめの状態を作ってやると、再び剥がす動作が誘起される。ここには画像と関節角度を用いた模倣学習の特徴の一つがよく現れているといえる。
  The work after the training is shown in \figref{fig:furoshiki-learned-success}. 
  % The left is the image from the external camera, 
  The upper right is the camera image of the head, and the lower right is the image decoded through Auto Encoder.
  The robot successfully performed a series of tasks three times in a row: removing a cloth, discarding it, a human repositioning the fallen cloth, and repeating the same action. The data for one time is here.
  Since the motion is performed by predicting the next step image from the current RGB image, if a person makes the initial state again after the work is completed once, the removal motion is induced again. This is one of the characteristics of imitation learning using images and joint angles.
  \begin{figure}[tb]
    \centering
    \includegraphics[width=0.99\columnwidth]{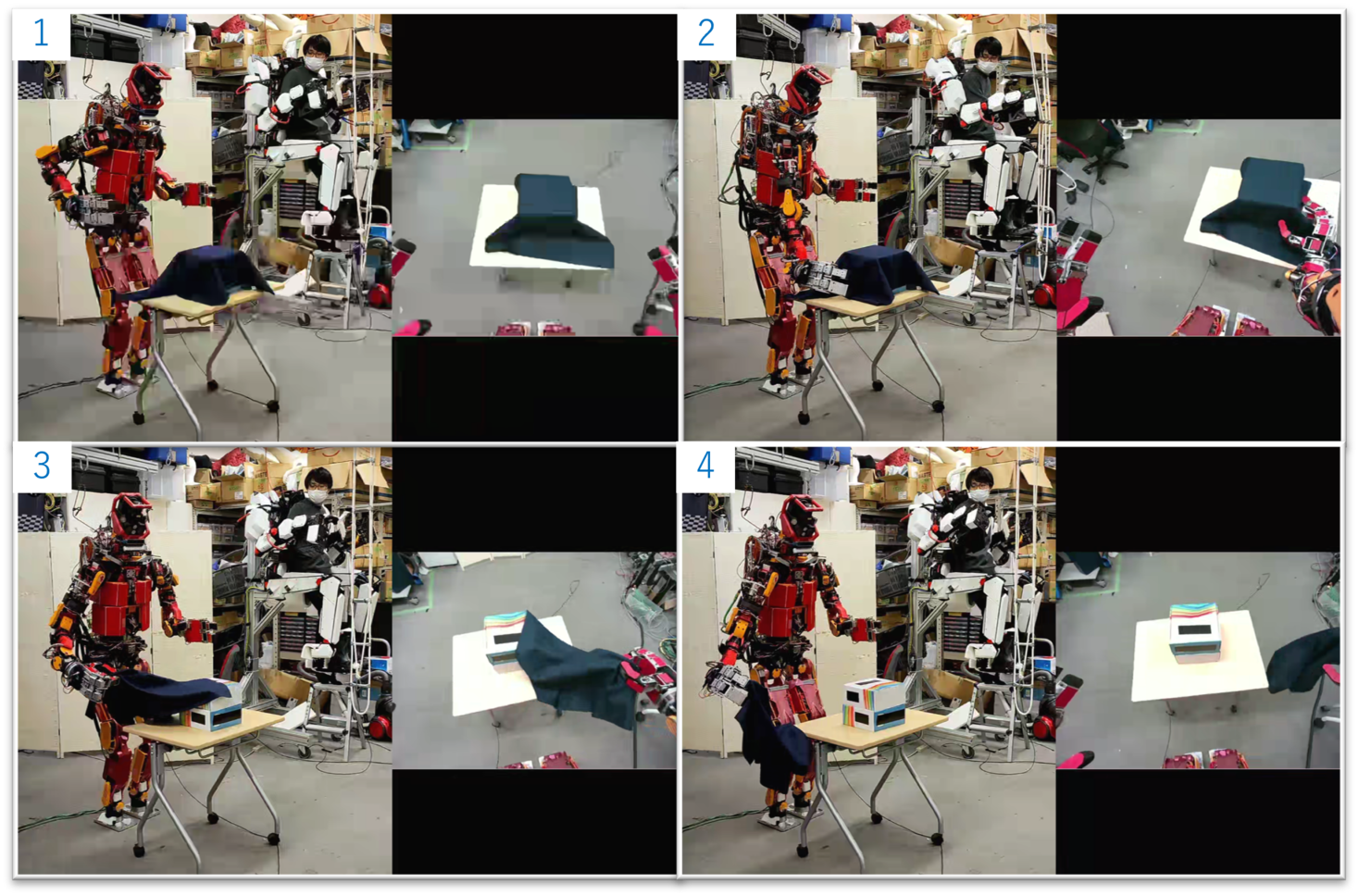}
    \caption{Experiment of removing cloth with proposed teleoperation system}
    \label{fig:furoshiki-teleop}
  \end{figure}
  \begin{figure}[tb]
    \centering
    \includegraphics[width=0.99\columnwidth]{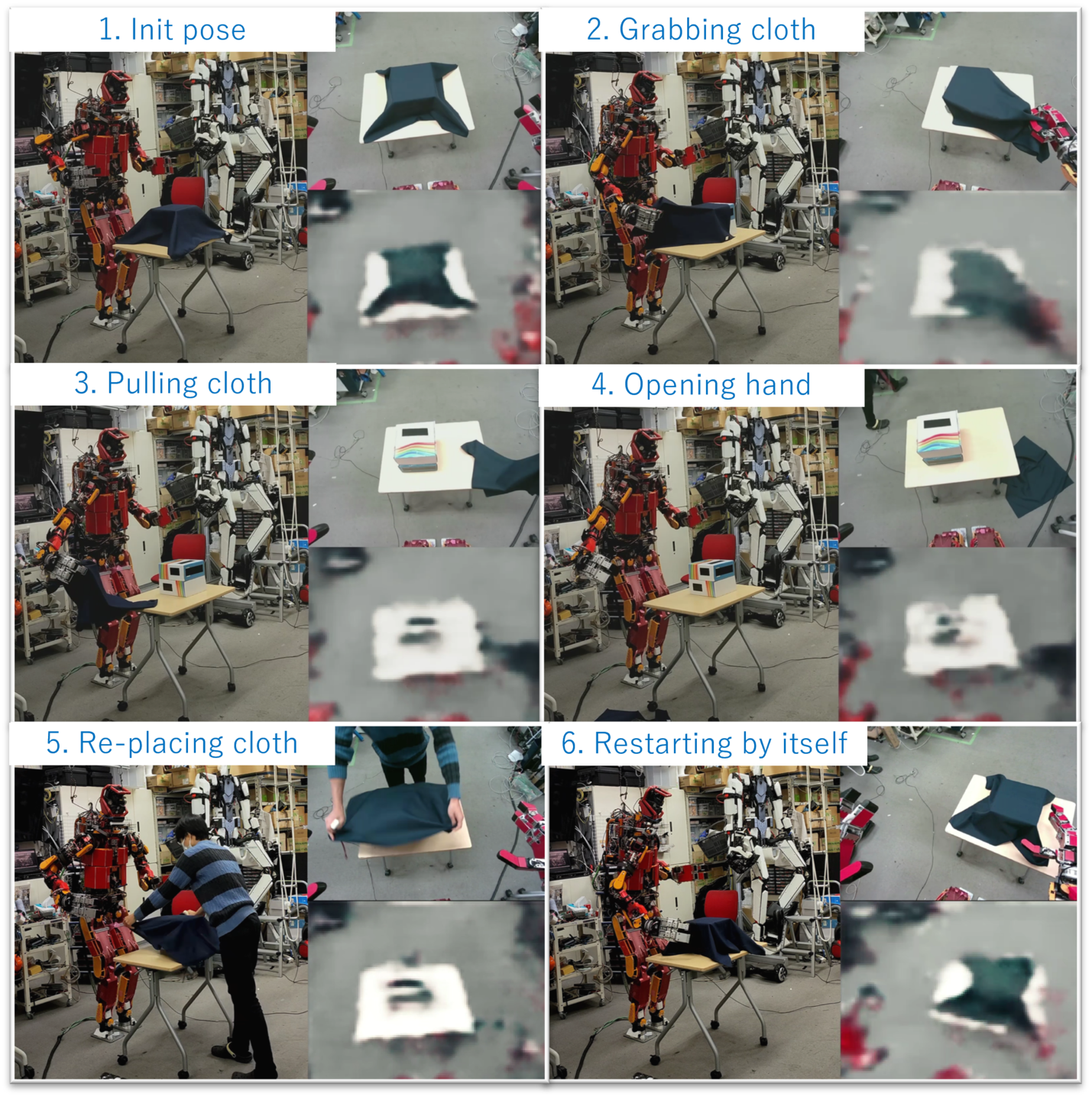}
    \caption{Experiment of succeeded removing cloth with imitation learning}
    \label{fig:furoshiki-learned-success}
  \end{figure}
  %% 模倣ができているかを確認するために、遠隔操縦時と学習後の自律作業時の指の角度の変化を比較したところ、角度変化のタイミングや最大、最小角度が類似していた。
  To confirm whether the robot was able to imitate, we compared the changes in finger angles during teleoperation and during autonomous work after learning, and found that the timing of angle changes and the maximum and minimum angles were similar.

  \subsection{Experiments on manipulating objects with a foot}
%%   ヒューマノイドが足を使ってゴミ箱を開けている様子を\figref{fig:gomibako-teleop}に示す。左が外部動画と右がRGB画像である。
%%   はじめに操縦者の足の位置を取得し提案システムで$\bthetahfpg$を計算する。計算された姿勢の両足の高さの差分が閾値以上(6 cm)となった際に、操縦を止めて\fsc{}からWhole Body Controllerに対して2足歩行モードから片足立ちモードへ切り替えるトリガーを送る。
%%   その際には、重心軌道を計算し遷移するためバランスを考慮した片足立ちへの移行が可能となっている。
%% 遷移後は操縦者による足の操縦を再開する。
%%   片足立ちモードでは力センサの値を低ゲインで操縦者へフィードバックしているため、ゴミ箱のペダルとの接触、操作時の反力を操縦者が認知することができる。
%%   ゴミ箱を開けた後は同様の手順を逆再生し、地面に足を戻す。
%%   この際も両足の高さの差分を検知しモードを切り替えるが、ロボットがペダルを操作している時にモードが切り替わることのないように、両足のx方向の位置が十分近い状態でのみモードを切り替えるようにした。
%%   学習後の作業の様子を\figref{fig:gomibako-success}に示した。
%%   左側が実験時の動画、右上がその時のカメラ画像、右下がAuto Encoderを通して復元された後の画像となっている。
  %%   足の接触状態の切り替えを含む操縦に成功している。
  \begin{figure}[tbh]
    \centering
    \includegraphics[width=0.99\columnwidth]{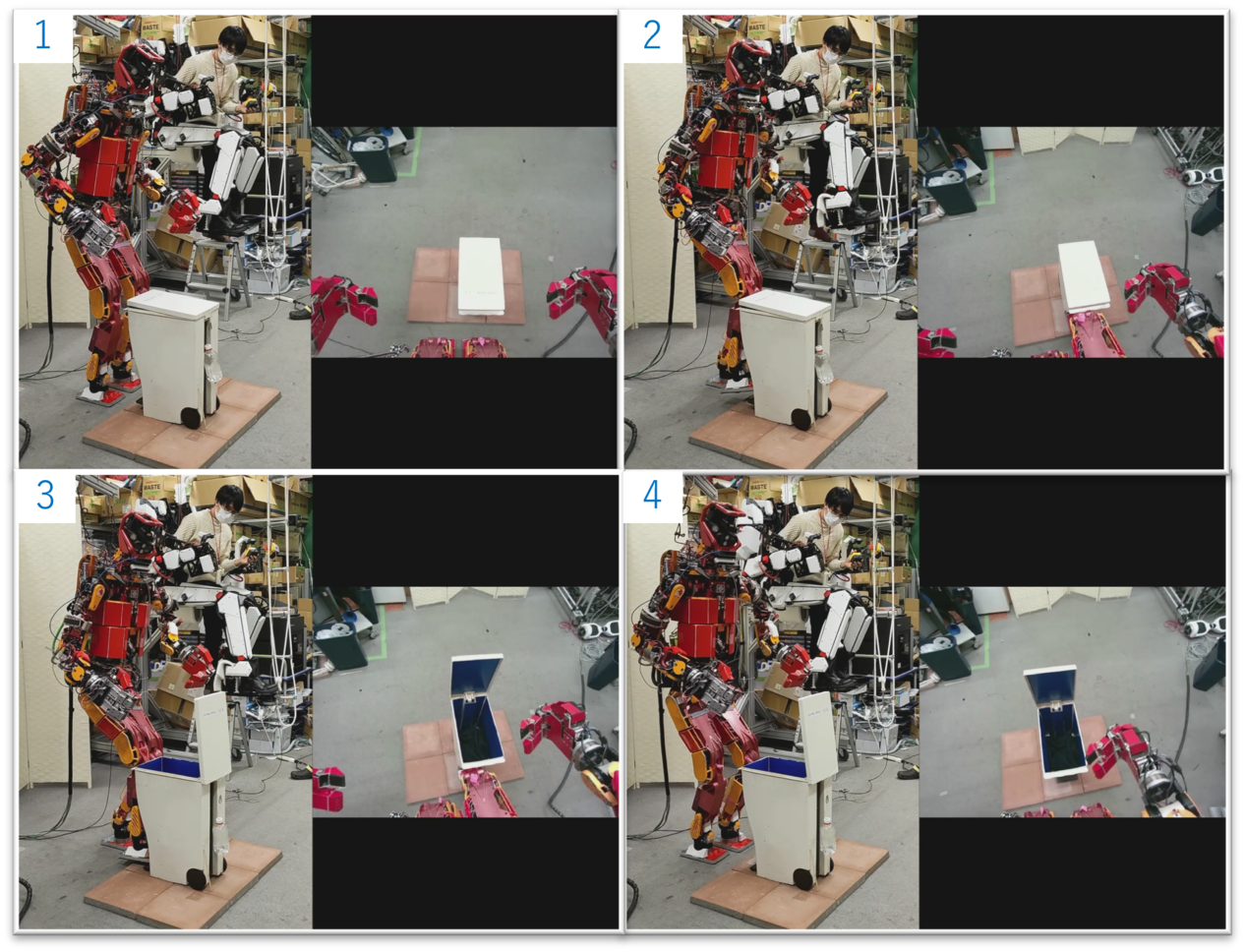}
    \caption{Experiment of operating a foot with proposed teleoperation system}\label{fig:gomibako-teleop}
  \end{figure}
  \begin{figure}[htb]
    \centering
    \includegraphics[width=0.99\columnwidth]{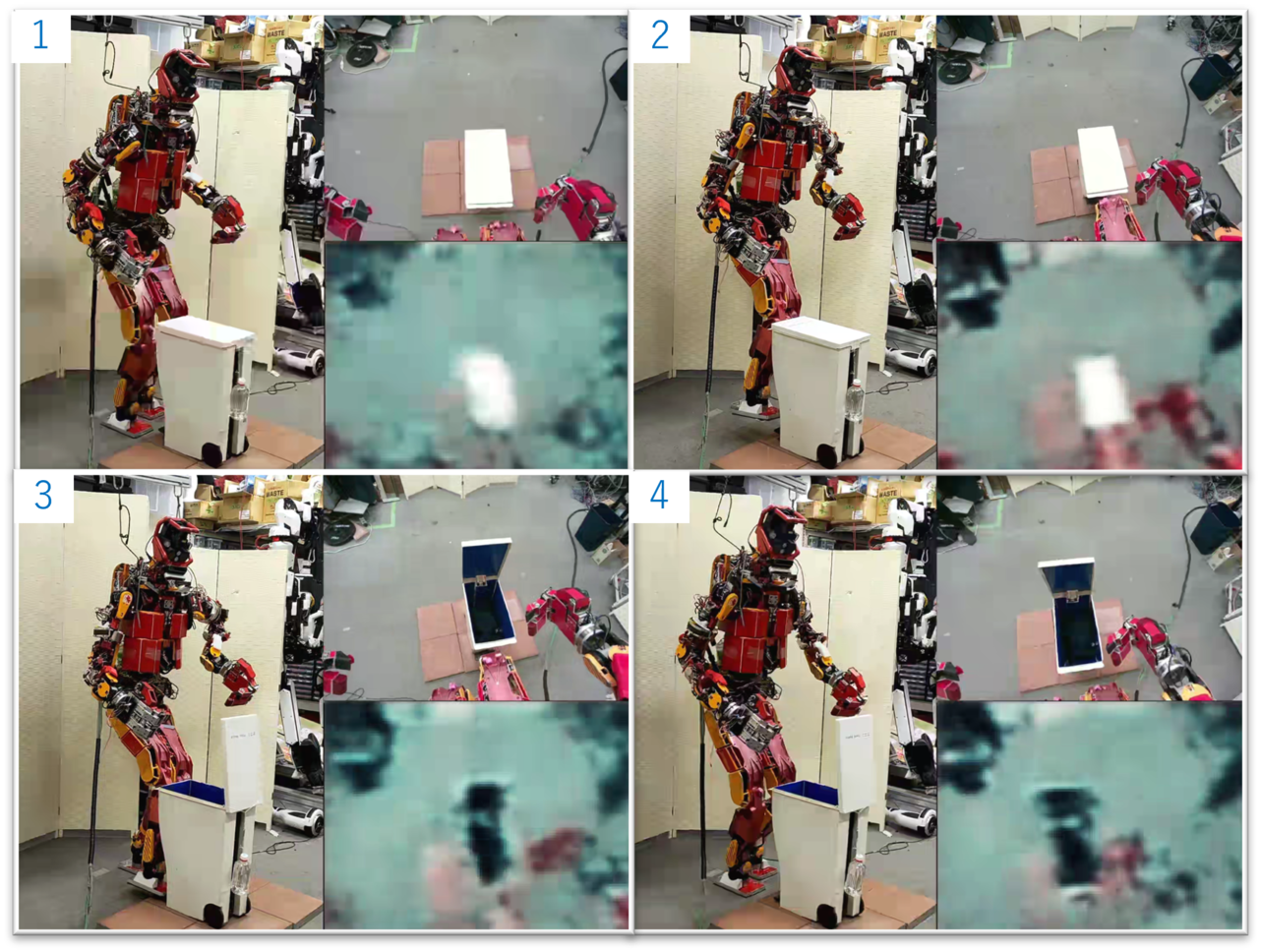}
    \caption{Experiment of succeeded operating a foot with imitation learning}\label{fig:gomibako-success}
  \end{figure}
  A humanoid opening a trash can using its legs is shown in \figref{fig:gomibako-teleop}.
  % The left image is from an external camera and 
  The right image is from a camera on the robot's head.
  First, the position of the operator's feet is obtained and $\bthetahfpg$ is calculated by the proposed system. When the difference between the heights of the two feet in the calculated posture exceeds a threshold value (6 cm), the system stops the maneuver and sends a trigger to the Whole Body Controller from \fsc{} to switch from the biped walking mode to the one-legged standing mode.
  %% In this case, the center of gravity trajectory is calculated and the transition is made so that the transition to standing on one leg can be made in consideration of balance.
  After the transition, the operator resumes control of the robot's legs.
  In the one-leg standing mode, the force sensor value is feedbacked to the operator with a low gain, therefore allowing the operator to recognize the contact with the trash can pedals and the reaction force when operating the trash can.
  After opening the trash can, the same procedure is reproduced in reverse, returning the feet to the ground.
  Again, the difference in the heights of the two feet is detected and the mode is switched.
  % , but only when the x positions of the two feet are close enough so that the mode is not switched when the robot is operating the pedals.
  The work after learning is shown in \figref{fig:gomibako-success}.
  % The left side is the image from the external camera, 
  The upper right is the camera image of the robot's head, and the lower right is the image decoded through Auto Encoder.
  The control including the switching of the contact state of the feet was successful.
  %% \begin{figure}[htb]
  %% \centering
  %% \includegraphics[width=0.48\columnwidth]{figs/gomibako/success/plot/gomibako_suc_force_x.png}
  %% \includegraphics[width=0.48\columnwidth]{figs/gomibako/success/plot/gomibako_suc_moment_x.png}\\
  %% \includegraphics[width=0.48\columnwidth]{figs/gomibako/success/plot/gomibako_suc_force_y.png}
  %% \includegraphics[width=0.48\columnwidth]{figs/gomibako/success/plot/gomibako_suc_moment_y.png}\\
  %% \includegraphics[width=0.48\columnwidth]{figs/gomibako/success/plot/gomibako_suc_force_z.png}
  %% \includegraphics[width=0.48\columnwidth]{figs/gomibako/success/plot/gomibako_suc_moment_z.png}\\
  %% \caption{Plot of foot wrench sensor value}
  %% \label{fig:gomibako-wrench}
  %% \end{figure}

  \subsection{Whole-body heavy object manipulation}
  %% ここでは、約16 kgの重量箱の持ち上げに取り組んだ。
  %% 膝を使った屈み込みと、手の反力を考慮したバランス維持が重要となる。
  %% 遠隔操縦による教示の様子を\figref{fig:mochiage-teleop}に示す。
  %% %%
  %% ロボットが人さし指と中指を90度でロックした初期姿勢から作業を開始する。両手を箱の持ち手に挿入し、反力を感じたら鉛直に持ち上げる作業を遠隔操縦で行った。
  %% 遠隔操縦システムの重心高さに対する制約のおかげで手先の高さが下がると膝を曲げ、ルートリンクをかたむけて重心を下げている。

  %% ヒューマノイドロボットには2脚で立ち続けるという強い制約が存在するため、特に重量物の操作では、手先の外力に合わせて重心位置を変更するといったバランス制御が不可欠である。
  %% 提案システムにおいては、最適化姿勢生成において重心を足平中心に維持する制約を加え、更に手先のセンサ値に合わせて重心位置を移動させるフィードバック制御を用いてバランスを維持している。
  %% 力センサ値に対するローパスフィルタに起因する時間遅れに対応するため、手先にかかる外力が大きく変化する持ち上げ始めでは、十分にゆっくりと手を動かすようにする必要があった。
  %% %%
  %% 学習を利用した作業の様子を\figref{fig:mochiage-teleop}に示す。
  %% 持ち手に対して両手を差し込み、重心を適切に移動させることでバランスを維持しながら持ち上げ作業に成功している。
  %% 実験の中では一度目の差し込み動作に失敗して箱が初期位置からずれしてしまった場合でも、ロボットが自律的に再度、適切な位置に対して手の差し込み直しをする動作が見られた。ロボットが自律的にリカバリー動作を行うという、模倣学習のロバスト性も確認できた。
  %% 操縦時と学習後の右手のセンサ値をプロットしたものを\figref{fig:mochiage-compare-wrench}に示す。
  %% 作業開始時を揃えてプロットしているため、環境の初期状態によって持ち上げはじめのタイミングはずれるのが自然である。しかし特にz軸方向の力や、z軸のモーメントの変化の概形は青色の学習前とオレンジ色の学習後で類似していると言える。

  \begin{figure}[tb]
    \centering
    \includegraphics[width=0.99\columnwidth]{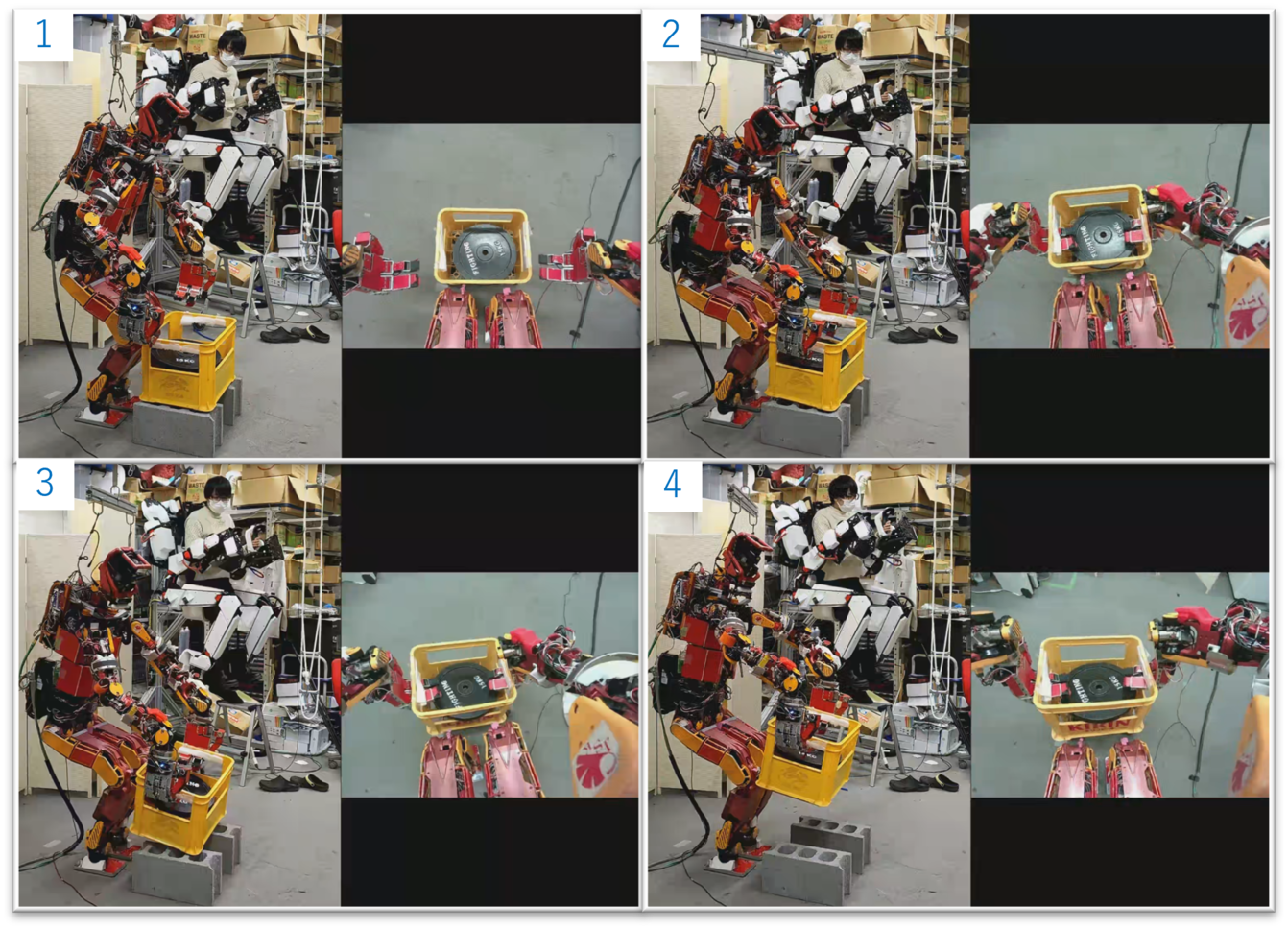}
    \caption{Experiment of lifting a heavy box with proposed teleoperation system}
    \label{fig:mochiage-teleop}
  \end{figure}
  \begin{figure}[tb]
    \centering
    \includegraphics[width=0.99\columnwidth]{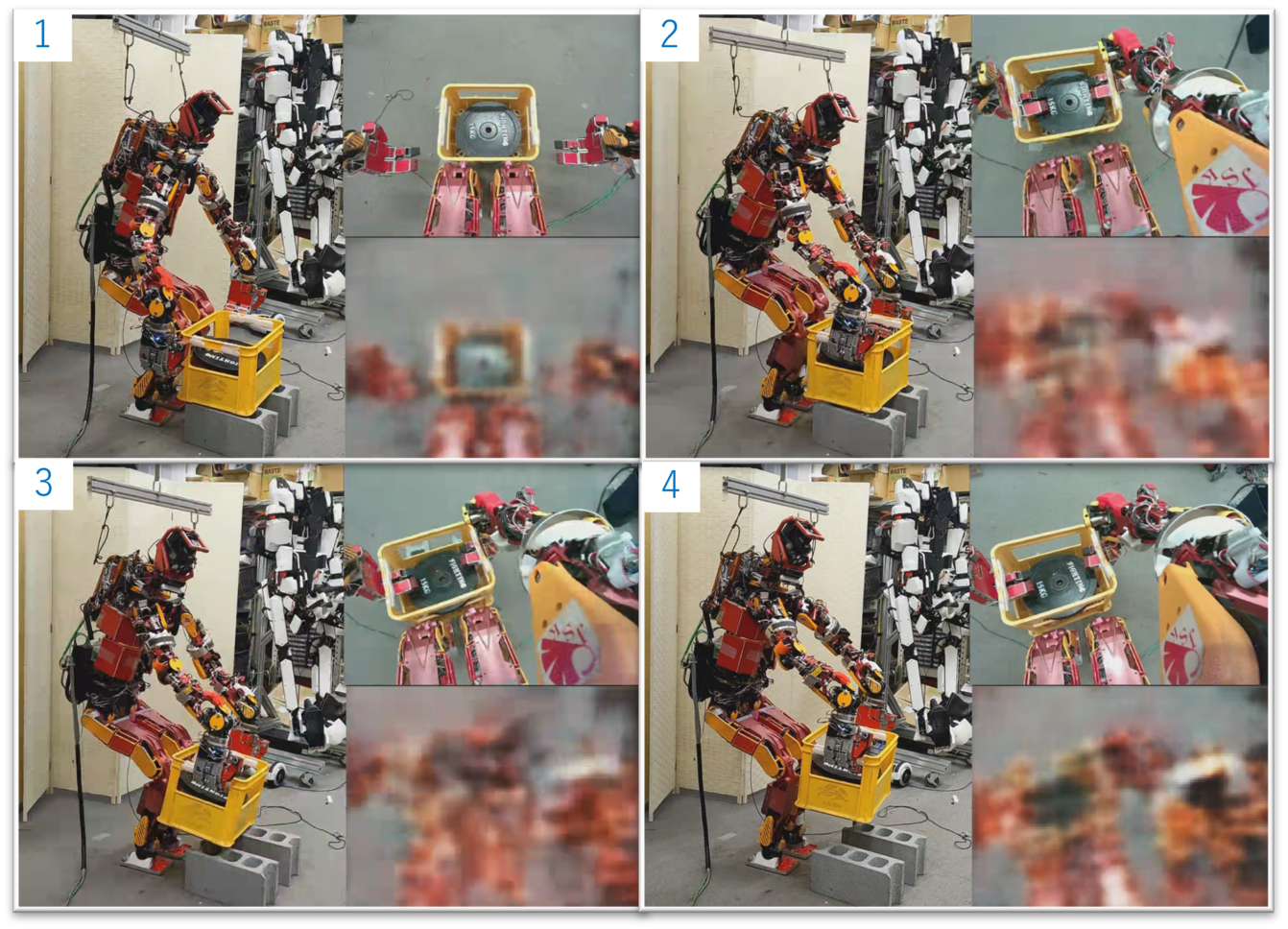}
    \caption{Experiment of succeeded lifting a heavy box with imitation learning}
    \label{fig:mochiage-success}
  \end{figure}

  Here we worked on lifting a heavy box weighing approximately 16 kg.
  Bending with the knees and maintaining balance by taking into account the reaction force of the hands are important.
  Teaching by teleoperation is shown in \figref{fig:mochiage-teleop}.
  %%%%.
  The robot starts work from the initial posture with the index and middle fingers locked at 90 degrees. Both hands were inserted into the box handles and lifted vertically when a reaction force was felt, using teleoperation.
  Thanks to the teleoperation system's constraint on the height of the center of gravity, when the height of the hand tip was lowered, the robot bent its knees and tilted the root link to lower its center of gravity.
  %%%%
  Since humanoid robots have the strong constraint of standing on two legs, balance control, such as changing the position of the center of gravity based on the external force of the hand tip, is indispensable, especially when manipulating heavy objects.
  In the proposed system, a constraint to keep the center of gravity at the center of the foot flat is added in the optimized posture generation, and the balance is maintained by using feedback control to shift the center of gravity position according to the sensor values of the hand tip.
  To cope with the time delay caused by the low-pass filter for the force sensor values, it was necessary to move the hand slowly enough at the beginning of lifting, when the external force on the hand tip changes significantly.
  %%%%.
  The work with the learning is shown in the \figref{fig:mochiage-success}.
  By inserting both hands against the handles and shifting the center of gravity appropriately, the lifting task was successfully performed while maintaining balance.
  In the experiment, even when the first insertion motion failed and the box shifted from the initial position, the robot autonomously reinserted its hands into the appropriate position again. The robustness of the imitation learning was also confirmed by the robot's autonomous recovery behavior.
  %% A plot of the sensor values of the right hand during the maneuver and after learning is shown in \figref{fig:mochiage-compare-wrench}.
  %% Since the plots are aligned with the start of the work, it is natural that the timing of the beginning of lifting is shifted depending on the initial state of the environment. However, in particular, the force in the z-axis direction and the general shape of the change in the moment of the z-axis are similar between the pre-learning (blue) and post-learning (orange).

  %% \begin{figure}[htb]
  %%   \centering
  %%   \includegraphics[width=0.48\columnwidth]{figs/mochiage/success2/plot/mochiage_compare_right_force_x_sec.png}
  %%   \includegraphics[width=0.48\columnwidth]{figs/mochiage/success2/plot/mochiage_compare_right_moment_x_sec.png}\\
  %%   \includegraphics[width=0.48\columnwidth]{figs/mochiage/success2/plot/mochiage_compare_right_force_y_sec.png}
  %%   \includegraphics[width=0.48\columnwidth]{figs/mochiage/success2/plot/mochiage_compare_right_moment_y_sec.png}\\
  %%   \includegraphics[width=0.48\columnwidth]{figs/mochiage/success2/plot/mochiage_compare_right_force_z_sec.png}
  %%   \includegraphics[width=0.48\columnwidth]{figs/mochiage/success2/plot/mochiage_compare_right_moment_z_sec.png}\\
  %%   \caption{Plot of hand wrench sensor value}
  %%   \label{fig:mochiage-compare-wrench}
  %% \end{figure}
}

\section{DISCUSSION} \label{sec:discussion}
\ifthenelse{\boolean{Draft}}{ %%%%%%% Japanese start %%%%%%%
    %% 最後に本研究の限界について, 操縦方法, 最適化計算を含む制御, 学習方法の観点から述べる.
  %% まず操縦方法について,今回は操縦者が対象物を直接見ながら作業を行った。しかし、理想的にはロボットのカメラ画像をVRゴーグルを介して提示するといった手法を採用すべきである。一方で, その場合にはカメラ画角が大きく変化してしまい, 模倣学習の汎化能力が著しく下がる可能性がある.
  %% また、このシステムでは四肢以外のリンクの位置姿勢を自動的に決定するため、操縦者が制御することができない。このため、腰でものを押すといった能動的にリンクを環境と接触させることは難しい。この問題は操縦者の腰や肘の位置姿勢を取得すれば解決できるが、最適化計算や自律的なバランス制御との衝突が起こる。本システムでは、腰の傾き等は基本的に自動的に決めるべきであり、個別の環境接触などは操縦ではなくプランナで対応するという立場で開発を行った。

  %% さらに、3指5自由度のMSLHANDに対して操縦インターフェースには1自由度しか無い。ソフトウェア的に対象物に合わせて指を制御する方法や、操縦インターフェースを改良する方法がある。
  %% 次に最適化計算について,現在は目的関数や制約に対する重みパラメタを作業に応じて人が調整している。これらのパラメタは自動的に調整されるのが理想的である。重みの決定アルゴリズムを人が一般化して与えることが可能だと考えられる。

  %% バランス制御に関しても、モードの切り替えが必要であるという小さな限界が存在する。石黒らの従来のシステムではモードの切り替えをせずに足も腕と同様に動かすことができた。しかし、本システムでは1足になるときにトリガーを送っている。
  %% これは、バランス維持のために両足と地面の接触を固定する機能や、転倒回避のための自律的な足の踏み出し機能をもつ２足モードの状態では、接触状態の変更ができないためである。
  %% これは、操縦者の司令とロボットの自律的なバランス制御の兼ね合いに関する問題であり、本システムでは自律的なバランス制御を優先している。制御理論やインターフェースの工夫により解決すべき課題の一つである。
  %% 最後に学習方法について,今回は人が作業の始点と終点を明示的に与え作業を区切って学習を行ったが、作業の区切りも自動的に判断し学習できるのが理想である。そのためには、例えば対象物の位置姿勢を基準とした作業の分節化を行うことなどが考えられる。
  }{ %%%%%% Japanese end: English start %%%%%
  We discuss the limitations of this study in terms of the teleoperation method, the control including optimization calculations, and the learning method.
  First, regarding the method of manipulation, in the present study, the operator performed the work while looking directly at the object. Ideally, however, a method such as presenting the robot's camera image through VR goggles should be adopted. On the other hand, in such a case, the camera angle of view would change significantly, and the generalization ability of imitation learning might be significantly reduced.
  In addition, since this system automatically determines the position and posture of links other than the limbs, the operator cannot control them. Therefore, it is difficult to bring the links into active contact with the environment, such as pushing an object with the waist. This problem can be solved by acquiring the positional posture of the operator's hips and elbows, but it conflicts with optimization calculations and autonomous balance control. In this system, we developed the system from the standpoint that the tilt of the hips, etc. should be determined automatically, and that individual environmental contacts, etc. should be handled by the planner, not by the pilot.
  Furthermore, while MSLHAND has 3 fingers and 5 DOFs, the maneuvering interface has only 1 DOF. There are ways to control the fingers according to the object from a software perspective or to improve the maneuvering interface.
  Next, regarding the optimization calculation, currently, the weight parameters for the objective function and constraints are adjusted by a person according to the task. Ideally, these parameters should be adjusted automatically. It is thought that a generalized algorithm for determining weights can be given by a person.
  A small limitation also exists concerning balance control, which is the need to switch modes. In the previous system by Ishiguro et al., the legs could be moved in the same way as the arms without switching modes. However, this system sends a trigger when it becomes one leg.
  This is because the contact state cannot be changed in the two-legged mode, which has the function of fixing contact between both feet and the ground to maintain balance and the autonomous foot step-out function to avoid tipping over.
  This is a problem related to the combination of the operator's command and the robot's autonomous balance control, and this system gives priority to autonomous balance control. This is one of the issues that should be solved by devising the control theory and interface.
  Finally, regarding the learning method, in the present study, the robot learned by explicitly giving the start and end points of the work, but ideally, the robot should be able to automatically determine the breakpoints and learn. 
  To achieve this, for example, it is possible to segment the work based on the position and posture of the object.
} %%%%%%% English end %%%%%%%

\section{CONCLUSIONS} \label{sec:conclusion}
\ifthenelse{\boolean{Draft}}{ %%%%%%% Japanese start %%%%%%%
  本研究では、浮遊リンク系の二足歩行ロボットによる模倣学習システムを開発することを目指した。
  そのために、バイラテラルで全身の操縦が可能な操縦デバイスと高耐久ヒューマノイドロボットを繋ぐ操縦システムを開発した。
  ここでは、トルク・接触力最適化姿勢生成手法を導入し、浮遊リンク系のヒューマノイドにおいて長時間のデータ収集にも耐え得るシステムとした。
  %% 本研究では二足歩行型のヒューマノイドにおける模倣学習のための教示システムとして、高負荷作業や長期的な作業継続のためのハードウェアと浮遊リンク系においても長期的なデータ取得が可能な制御を含む操縦模倣システムを構築した.
  このシステムを使用して、柔軟布の操作、足を使った物体操作、ルートリンクの姿勢変化と重心移動が重要な重量物の持ち上げを模倣することに成功した。

  最後に本研究の限界について, 操縦方法, 最適化計算を含む制御, 学習方法の観点から述べる.
  まず操縦方法について,今回の操縦では操縦者が操作対象物を直接目視で確認しながら作業を行った。しかし、理想的にはロボットの頭部のカメラ画像をVRゴーグルを介して提示するといった手法を採用すべきである。一方で, その場合に頭部からのカメラ画像が大きく動いてしまい, 模倣学習の汎化能力が著しく下がる可能性がある.
  また、現状の操縦システムでは四肢以外のリンクの操縦ができない。エンドエフェクタ以外のリンクの位置姿勢は自動的に決定される。このため、例えば腰でものを押す、肘でものをどかすといった能動的にリンクを環境と接触させることは難しい。この問題は操縦者の腰や肘の位置姿勢を取得すれば解決できるが、最適化計算や自律的なバランス制御との衝突が起こる。本システムでは、腰の傾き等は基本的に自動的に決めるべきであり、個別の環境接触などは操縦ではなくプランナで対応するという立場で開発を行った。
  さらに、3指5自由度のMSLHANDに対して操縦インターフェースには1自由度しか無い。ソフトウェア的に対象物に合わせて指の動きを制御する方法や、手先の操縦インターフェースを改良する方法がある。
  次に最適化計算について,現在は目的関数や制約に対する重みのパラメタを作業に応じて人が調整している。これらのパラメタは自動的に調整されるのが理想的である。重みの決定アルゴリズムを人が一般化して与えることが可能だと考えられる。
  バランス制御に関しても、モードの切り替えが必要であるという小さな限界が存在する。石黒らの従来のシステムではモードの切り替えをせずに足も腕と同様に動かすことができた。しかし、本システムでは1足になるときにトリガーを送っている。これは、２足モードの状態ではバランス維持のために両足と地面の接触を固定する機能や、転倒回避のための自律的な足の踏み出し機能が働くため、接触状態の変更ができないためである。これは、操縦者の司令とロボットの自律的なバランス制御の兼ね合いに関する問題であり、本システムでは自律的なバランス制御を優先している。制御理論やインターフェースの工夫により解決すべき課題の一つである。
  最後に学習方法について,今回は人が作業の始点と終点を明示的に与え作業を区切って学習を行ったが、作業の区切りも自動的に判断し学習までができるのが理想である。そのためには、例えば操作対象物の位置姿勢を基準とした作業の分節化を行うことなどが考えられる。
}{ %%%%%% Japanese end: English start %%%%%
  %% 本研究では、浮遊リンクの二足歩行ロボットによる模倣学習システムを開発することを目指した。
  %% そのために、バイラテラルの全身操縦デバイスと高耐久ヒューマノイドロボットを繋ぐ操縦システムを開発した。
  %% ここでは、トルク・接触力最適化姿勢生成手法を導入し、長時間のデータ収集にも耐え得るシステムとした。
  %% このシステムを使用して、柔軟布の操作、足を使った物体操作、ルートリンクの姿勢変化と重心移動が重要な重量物の持ち上げを模倣することに成功した。

  %% 最後に本研究の限界について, 操縦方法, 最適化計算を含む制御, 学習方法の観点から述べる.
  %% まず操縦方法について,今回は操縦者が対象物を直接見ながら作業を行った。しかし、理想的にはロボットのカメラ画像をVRゴーグルを介して提示するといった手法を採用すべきである。一方で, その場合にはカメラ画角が大きく変化してしまい, 模倣学習の汎化能力が著しく下がる可能性がある.
  %% また、このシステムでは四肢以外のリンクの位置姿勢を自動的に決定するため、操縦者が制御することができない。このため、腰でものを押すといった能動的にリンクを環境と接触させることは難しい。この問題は操縦者の腰や肘の位置姿勢を取得すれば解決できるが、最適化計算や自律的なバランス制御との衝突が起こる。本システムでは、腰の傾き等は基本的に自動的に決めるべきであり、個別の環境接触などは操縦ではなくプランナで対応するという立場で開発を行った。

  %% さらに、3指5自由度のMSLHANDに対して操縦インターフェースには1自由度しか無い。ソフトウェア的に対象物に合わせて指を制御する方法や、操縦インターフェースを改良する方法がある。
  %% 次に最適化計算について,現在は目的関数や制約に対する重みパラメタを作業に応じて人が調整している。これらのパラメタは自動的に調整されるのが理想的である。重みの決定アルゴリズムを人が一般化して与えることが可能だと考えられる。

  %% バランス制御に関しても、モードの切り替えが必要であるという小さな限界が存在する。石黒らの従来のシステムではモードの切り替えをせずに足も腕と同様に動かすことができた。しかし、本システムでは1足になるときにトリガーを送っている。
  %% これは、バランス維持のために両足と地面の接触を固定する機能や、転倒回避のための自律的な足の踏み出し機能をもつ２足モードの状態では、接触状態の変更ができないためである。
  %% これは、操縦者の司令とロボットの自律的なバランス制御の兼ね合いに関する問題であり、本システムでは自律的なバランス制御を優先している。制御理論やインターフェースの工夫により解決すべき課題の一つである。
  %% 最後に学習方法について,今回は人が作業の始点と終点を明示的に与え作業を区切って学習を行ったが、作業の区切りも自動的に判断し学習できるのが理想である。そのためには、例えば対象物の位置姿勢を基準とした作業の分節化を行うことなどが考えられる。
  This study aimed to develop an imitation learning system with a bipedal robot with a floating link.
  For this purpose, we developed a teleoperation system that connects a bilateral whole-body maneuvering device and a highly durable humanoid robot.
  Here, we introduced a torque/contact force-optimized posture generation method to make the system capable of withstanding long hours of data collection.
  Using this system, we successfully imitated flexible fabric manipulation,  object manipulation with feet, and lifting of heavy objects where the pose change of the root link and center-of-gravity shift are critical.

} %%%%%%% English end %%%%%%%

\bibliographystyle{ieeetr}
\bibliography{bib}

\end{document}